%% file: main.tex
\documentclass{article}
\usepackage[T1]{fontenc}
\usepackage{iclr2027_conference,times}
\input{math_commands.tex}

\usepackage{hyperref}
\usepackage{url}
\usepackage{booktabs}
\usepackage{graphicx}
\usepackage{amsmath,amssymb}
\usepackage{dsfont}
\usepackage{subcaption}
\usepackage{makecell}
\usepackage{pifont}
\usepackage{xcolor}
\usepackage{array}
\usepackage{multirow}
\newsavebox{\resulttable}
\usepackage{tikz}
\usepackage{xspace}
\usepackage{enumitem}
\usepackage{placeins}
\usepackage{titletoc}
\titlecontents{psection}[1.8em]{}{\contentslabel{1.8em}}{\hspace*{-1.8em}}{\titlerule*[0.6pc]{.}\contentspage}

\newcommand{\editedjanaka}[1]{#1}
\newcommand{\editjanaka}[1]{#1}
\newcommand{\editjana}[1]{#1}
\newcommand{\rev}[1]{#1}
\usetikzlibrary{arrows.meta,positioning,calc}
\input{figures/vector_defs}
\newcommand{\jns}{\textsc{JaxNaviSim}\xspace}
\newcommand{\sns}{ShipNaviSim}
\newcommand{\gcade}{\mathrm{gc\text{-}ADE}}
\newcommand{\ind}{\mathds{1}}
\newcommand{\reachg}{R_G}

\newcommand{\cmark}{\ding{51}}
\newcommand{\xmark}{\ding{55}}
\title{\jns{}: A GPU-Accelerated, Data-Driven\\
Simulator and Learning Benchmark\\
for Maritime Navigation}
\author{
  Quang Anh Pham$^{1}$\thanks{Equal contribution.}, \,
  Janaka Chathuranga Brahmanage$^{1}$\footnotemark[1], \,
  Hue Tse Leong$^{1}$\footnotemark[1], \,
  Akshat Kumar$^{1}$, \\
  \textbf{Tien Anh Mai$^{1}$, \,
  Siu Loon Hoe$^{1}$, \,
  Venky Shankararaman$^{1}$} \\[4pt]
  $^{1}$Singapore Management University \\
  \texttt{qa.pham.2025@phdcs.smu.edu.sg}
}

\iclrfinalcopy %
\begin{document}
\let\origfloatpagefraction\floatpagefraction
\renewcommand{\floatpagefraction}{0.85}
\maketitle
\input{sections/00_abstract}

\input{sections/01_intro}
\input{sections/02_related}

\input{sections/03_jaxnavisim}
\input{sections/06_baselines}

\input{sections/06_results}
\input{sections/08_limitations_conclusion}

\label{page:mainend}
\input{sections/09_statements}
\bibliography{references}
\bibliographystyle{iclr2027_conference}
\clearpage
\appendix
\raggedbottom
\setcounter{topnumber}{4}
\setcounter{bottomnumber}{3}
\setcounter{totalnumber}{6}
\renewcommand{\bottomfraction}{0.8}
\startcontents[appendix]
{\color{blue}
\section*{Appendix contents}
\printcontents[appendix]{p}{1}{\setcounter{tocdepth}{1}\setlength{\parskip}{0pt}}
\par}
\input{sections/10_appendix}
\input{dynamics_details}

\input{sections/dynamics_anchorage}

\input{navigation_results}

\input{sections/12_rebuttal_appendix}
\end{document}

%% file: math_commands.tex
\usepackage{amsmath,amsfonts,bm}

\def\eqref#1{equation~\ref{#1}}

\def\1{\bm{1}}

\newcommand{\test}{\mathcal{D_{\mathrm{test}}}}

\DeclareMathAlphabet{\mathsfit}{\encodingdefault}{\sfdefault}{m}{sl}
\SetMathAlphabet{\mathsfit}{bold}{\encodingdefault}{\sfdefault}{bx}{n}

%% file: figures/vector_defs.tex
\definecolor{vecact}{HTML}{1b6f8a}    %
\definecolor{vecturn}{HTML}{1b7837}   %
\definecolor{vecdrift}{HTML}{b2182b}  %
\definecolor{vecgnd}{HTML}{404040}    %
\definecolor{vecmiss}{HTML}{ef8a62}   %

\newcommand{\hullglyph}[1]{%
  \begin{scope}[rotate=#1]
    \draw[draw=black!65,fill=black!4,line width=0.4pt]
      (-0.42,-0.15) -- (0.20,-0.15) -- (0.44,0) -- (0.20,0.15) -- (-0.42,0.15) -- cycle;
  \end{scope}}
\newcommand{\hullatc}[2]{%
  \begin{scope}[shift={#1},rotate=#2]
    \draw[draw=black!65,fill=black!4,line width=0.4pt]
      (-0.42,-0.15) -- (0.20,-0.15) -- (0.44,0) -- (0.20,0.15) -- (-0.42,0.15) -- cycle;
  \end{scope}}
\newcommand{\ghosthullatc}[2]{%
  \begin{scope}[shift={#1},rotate=#2]
    \draw[draw=black!45,fill=white,dashed,line width=0.4pt]
      (-0.42,-0.15) -- (0.20,-0.15) -- (0.44,0) -- (0.20,0.15) -- (-0.42,0.15) -- cycle;
  \end{scope}}

\tikzset{
  vecpic/.style={
    x=1cm,y=1cm,
    vec/.style={-{Stealth[length=2.2mm,width=1.6mm]},line width=0.9pt},
    bow/.style={draw=black!45,dashed,line width=0.5pt},
    turn/.style={-{Stealth[length=1.6mm]},vecturn,line width=0.7pt},
    lbl/.style={font=\scriptsize},
    ttl/.style={font=\small\bfseries},
  }}

%% file: sections/00_abstract.tex
\begin{abstract}
Coordinating vessels through the shared waterways of busy ports grows more challenging as vessel traffic increases. Port authorities need realistic
simulations to assess how arrival schedules, anchorage assignments, and routes
affect vessel throughput and navigation safety.
\editjanaka{Existing AIS-based trajectory models predict a vessel's track from its past trajectory in an open-loop fashion. They do not simulate step-by-step responses to other ships or test whether a vessel collides or reaches its goal. Autonomous-driving research has addressed similar limits with closed-loop simulators.
However, existing driving simulators do not fill the gaps in maritime navigation because the nature of the challenges is different. Ships operate on a different timescale, with trajectories that can last over an hour. They need to follow different rules, such as traffic separation schemes (TSS), and cannot stop or turn quickly.}
To address this problem, we introduce \jns, a data-driven simulator and learning benchmark coupling Automatic Identification System (AIS) data with fast, controllable traffic simulation\editjanaka{, in which learned models steer vessels through recorded traffic}. \textit{First}, we release a dataset of $\sim${141 million} AIS records from over {30K} vessels across eight high-density regions of the Singapore Strait, among the world's busiest waterways, with $\sim$6.7 million vessel-hours. {Unlike several existing AIS datasets which mostly cover open seas, ours captures close-quarters vessel traffic in the narrow strait.} \textit{Second}, \editjanaka{we define a navigation task for each recorded vessel trajectory. A model steers that vessel every 10\,s from its recorded state to its anchoring point or region exit, while surrounding ships follow their recorded tracks or are steered by the model. We implement a JAX simulator that runs these tasks in large batches on a GPU.} {We also incorporate multiple vessel dynamics models and  sea surface currents.} \textit{Finally}, \editjanaka{we benchmark imitation learning (IL), offline and online RL, and include Predictive Action Refinement (PAR) as a safety baseline that cuts collision rates to 1--3\%. The comparison shows that low tracking error does not ensure arrival, and learned policies can reach most destinations yet collide in 20--45\% of episodes.}

\end{abstract}

%% file: sections/01_intro.tex
\section{Introduction}
\label{sec:intro}

Maritime traffic management in busy ports requires coordinating vessel movements through constrained, shared waters~\citep{corry2019berth,jia2020pilots}. In a traffic separation scheme (TSS), vessels enter, follow, cross, and leave traffic streams; at anchorages, arriving vessels approach assigned anchor positions among anchored ships and passing traffic. Operational decisions such as arrival timing, anchorage assignment, and approach route shape both a vessel's path and the encounters it faces. This is particularly relevant in busy ports such as Singapore's where vessel movements must be coordinated within limited sea space \citep{agussurja2018scheduling}. 

Historical data record only one outcome of a day's traffic decisions. Port planners, however, are often interested in counterfactual questions:~\textit{how would vessel interactions, traffic throughput, navigation safety change if an arrival occurred later, a vessel were assigned to a different destination, or a different approach route were selected?} Answering them requires behavioral models that generate vessel movements under conditions not recorded historically. Automatic Identification System (AIS) data offer a large record from which realistic vessel behavior can be learned. \editjana{Predicting a vessel's track from its recorded history does not show how it responds when its own actions change later encounters.}
We therefore separate decisions from behavior:~experts or traffic-management systems specify feasible operational decisions, and the simulator evaluates the behavioral consequences of these decisions rather than inferring them from AIS alone.

Prior maritime works evaluate learned models mostly in open loop: a trajectory predicted from the logged history is compared with the recorded one using displacement errors~\citep{li2023shiptrajectory,nguyen2024traisformer}. A navigation policy, however, acts sequentially. In closed loop, each action changes the vessel's subsequent position and observations, and small errors can move the policy away from the demonstrated states and compound over time~\citep{ross2011reduction}. Autonomous-driving research has shown that open-loop scores are poorly aligned with closed-loop performance~\citep{dauner2023pdm}, motivating closed-loop benchmarks such as nuPlan and Waymax~\citep{karnchanachari2024nuplan,gulino2023waymax}. The same holds at sea: low displacement error does not prevent collisions or missed destinations, and avoiding encounters by large detours fails the task. \editjana{Ship passages last tens of minutes to over an hour, and large vessels cannot stop or turn quickly.} We therefore evaluate policies in closed loop on goal completion, arrival timing, trajectory fidelity, kinematics, and collisions, against both logged and reactive traffic.

\editjana{A maritime benchmark should test whether a policy reaches its assigned destination safely among surrounding ships.} Large public AIS archives~\citep{dma2026ais,noaa2026accessais} release raw point records from which users must construct trajectories, and task splits~\citep{ma2026envship}.
Processed resources target trajectory forecasting~\citep{nguyen2024traisformer,ma2026envship} rather than closed-loop navigation. Existing maritime simulation and learning platforms~\citep{pham2025shipnavisim,vaidheeswaran2025marinav} do not jointly provide public expert demonstrations, accelerator-native batched execution, and flexible multi-vessel control. We thus introduce \jns{}, a data-driven simulator and benchmark for learning and evaluating maritime navigation across anchorage and TSS settings. Our contributions are:

\begin{itemize}[leftmargin=*,labelindent=0.5em,nosep]
\item \textbf{Public AIS data and reproducible navigation tasks.}
We release over {141 million} AIS records from more than {30K} vessels across eight high-density anchorage and TSS regions in the Singapore Strait. We turn these data into goal-reaching navigation tasks with predefined training and test splits for reproducible learning and evaluation.
\item \textbf{Scalable simulation with configurable vessel dynamics and interactions.}
Our functional JAX~\citep{bradbury2018jax} simulator supports historical replay, selective vessel control against replayed traffic, and joint self-play through a common interface. Batched GPU simulation reaches 3.1 million single-vessel and 53.6 million multi-vessel steps per second on a consumer-grade RTX 3090\rev{ in a transition-only microbenchmark}, supporting the many rollouts that learning and evaluation require. The simulator accounts for hull dimensions and surrounding traffic, which shape vessel behavior and collision risk~\citep{yang2024harnessing}, and offers alternative vessel dynamics models, and sea surface currents induced external effects.

\item \textbf{A common evaluation protocol and comparative baselines.}
We benchmark imitation learning (IL) and offline and online reinforcement learning (RL) on punctuality, trajectory fidelity, kinematics, and dimension-aware collision rates, and ablate key design choices. Learned policies can achieve high goal completion but may incur high collision rates. \rev{Following the predictive safety filter principle in control~\citep{wabersich2021predictive}, we} therefore \rev{introduce} Predictive Action Refinement (PAR)\rev{, inspired by PDM-Closed~\citep{dauner2023pdm}, as our safety layer}, which refines actions with short-horizon interaction forecasts and substantially reduces collision rates. \rev{We also compare PAR with two cost-based test-time layers to show its effectiveness.} %
\end{itemize}
{Beyond maritime applications, the setting offers learning researchers long episodes, local observations, and dense vessel interactions. With expert demonstrations and controllable simulation, it supports studying, through IL and offline and online RL, how policies behave over many decisions and how goal completion relates to trajectory fidelity and collision avoidance.}

%% file: sections/02_related.tex
\section{Related work}
\label{sec:related}

\begin{table}[t]
\centering
\small
\setlength{\tabcolsep}{4pt}
\resizebox{\linewidth}{!}{%
\begin{tabular}{lcccccccc}
\toprule
& Domain & Real data & Open code/dataset & \editedjanaka{GPU simulation} & Collision check & Multi-agent & IL & Offline/online RL \\
\midrule
nuPlan~\citeyearpar{caesar2021nuplan} & road & \cmark & \cmark & \xmark & box overlap & \xmark & \cmark & \xmark \\
\editedjanaka{NAVSIM~\citeyearpar{dauner2024navsim}} & \editedjanaka{road} & \editedjanaka{\cmark} & \editedjanaka{\cmark} & \editedjanaka{\xmark} & \editedjanaka{box overlap} & \editedjanaka{\xmark} & \editedjanaka{\cmark} & \editedjanaka{\xmark} \\
Nocturne~\citeyearpar{vinitsky2022nocturne} & road & \cmark & \cmark & \xmark & box overlap & \cmark & \cmark & online \\
Waymax~\citeyearpar{gulino2023waymax} & road & \cmark & \cmark & \cmark & box overlap & \cmark & \cmark & online \\
GPUDrive~\citeyearpar{kazemkhani2025gpudrive} & road & \cmark & \cmark & \cmark & box overlap & \cmark & \cmark & online \\
MariNav~\citeyearpar{vaidheeswaran2025marinav} & sea & \cmark & code & \xmark & \xmark & \xmark & \xmark & online \\
\sns{}~\citeyearpar{pham2025shipnavisim} & sea & \cmark & code & \xmark & proximity threshold & \xmark & \cmark & \xmark \\
Zhang et al.~\citeyearpar{zhang2026realistic} & sea & \cmark & \xmark & \xmark & proximity threshold & \xmark & \xmark & online \\
\midrule
\jns{} (ours) & sea & \cmark & \cmark & \cmark & capsule overlap & \cmark & \cmark & \cmark \\
\bottomrule
\end{tabular}}
\caption{\small \jns{} v.s. other simulators. ``Open code/dataset'': code and curated dataset are public (``code'': code only). ``IL'': imitation learning from logged trajectories. ``Offline/online RL'': ``online'' denotes online RL only; \cmark{} denotes both. ``Collision check'': ``overlap'' means the vehicle or vessel shapes intersect; ``proximity threshold'' is a distance-based collision-risk proxy, such as the 555\,m radius of \sns{}.}
\label{tab:comparison}
\vskip -12pt
\end{table}

Table~\ref{tab:comparison} compares \jns{} with other data-driven simulators.

\textbf{Data-driven simulator for autonomous driving (AD) }Early works such as nuPlan~\citep{caesar2021nuplan} combine public driving scenarios and approximately 1,500 hours of human-driving data with closed-loop planning evaluation. \editedjanaka{NAVSIM~\citep{dauner2024navsim} scores a proposed driving trajectory over a 4\,s non-reactive rollout using collision and progress metrics.} Nocturne~\citep{vinitsky2022nocturne} uses real-world driving scenes from the Waymo Open Motion Dataset~\citep{ettinger2021large} to benchmark partially observable multi-agent reinforcement learning and imitation learning. More recent works such as Waymax~\citep{gulino2023waymax} and GPUDrive~\citep{kazemkhani2025gpudrive} focus on GPU-accelerated simulation for scalable policy training and evaluation. Waymax provides a JAX-based simulator that supports batched execution on GPUs, whereas GPUDrive uses a CUDA-based engine for high-throughput multi-agent reinforcement learning. Many other AD simulators exist, owing to large public datasets and the need to benchmark planning and control~\citep{li2024choose,zhang2024multiagent}. Our focus, vessel navigation, however, differs from driving. Driving scenarios typically last seconds (9\,s in Waymax), whereas a vessel's passage through one of our regions lasts tens of minutes to over an hour. A vessel cannot stop or turn quickly, and sea-surface currents affect its motion. These characteristics are not represented in driving simulators.

\textbf{Data-driven simulator for maritime navigation }In contrast to AD, far fewer studies have investigated data-driven simulation for maritime navigation. MariNav~\citep{vaidheeswaran2025marinav} uses an AIS-derived traffic graph and wind data for single-vessel routing. \citet{zhang2026realistic} train a single-vessel curriculum RL policy for fuel-aware routing in an Indian Ocean environment built from proprietary AIS data, but no land, traffic lanes, or anchorages, and assess collision risk by proximity criteria.
The closest work, \sns{}~\citep{pham2025shipnavisim}, learns from AIS data, but its released code runs only CPU-based single-vessel control on synthetic examples. It also measures proximity with a fixed 555\,m near-miss threshold. In contrast, our collision metric detects geometric overlap from vessel dimensions and orientations. This matters most in anchorages, where vessels often operate at close quarters. Overall, no existing maritime platform provides both public AIS data and code, or AD-simulator functionality such as GPU execution and geometry-based collision detection. \jns{} combines these with flexible single- and multi-vessel control in one benchmark, supporting IL, offline and online RL in anchorage and TSS settings.

%% file: sections/03_jaxnavisim.tex
\section{\jns{}}
\label{sec:jaxnavisim}

We next describe the problem setting, dataset and task construction, the simulator interface, and the metrics used to evaluate policies.

\subsection{Maritime Problem Setting}
\label{sec:setting}
\begin{figure}[t]
  \centering\small
  \begin{subfigure}[t]{0.617\textwidth}
    \includegraphics[width=\linewidth]{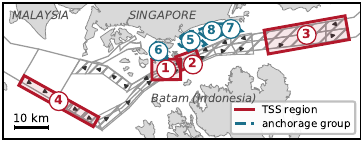}
    \caption{TSS lanes and the eight released regions.}
    \label{fig:regions}
  \end{subfigure}\hfill
  \begin{subfigure}[t]{0.363\textwidth}
    \includegraphics[width=\linewidth]{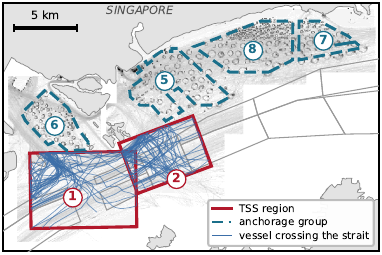}
    \caption{Anchorage density, crossing traffic.}
    \label{fig:cluster}
  \end{subfigure}
  \caption{\small\textbf{The Singapore Strait setting.} Regions 1--4: Intersection~1, Intersection~2, TSS East, TSS West. Regions 5--8: anchorage groups AEW, AWW, AEPBC, AESPA, each a cluster of contiguous anchorages, together holding 20 of Singapore's 32 anchorages and 72\% of their area. (a)~TSS lanes and direction of travel. (b)~AIS position density on 3~June~2024 (grey, log scale). Each ring is an anchored ship swinging around its anchor; no other ship may anchor within it, which tightly constrains space. Median vessels at rest over 2024: 111, 66, 42, and 34 in regions 8, 5, 6, and 7. Blue: vessels crossing the strait through the intersections.}
  \label{fig:released-regions}
  \vskip -10pt
\end{figure}

The Singapore Strait is one of the world's busiest waterways. We release eight regions of it that differ in how vessels navigate (Figure~\ref{fig:released-regions}). The four anchorage groups (regions~5--8) are highly constrained spaces, where pilots maneuver slowly among closely anchored ships to reach an anchoring position. TSS East and TSS West (regions~3 and~4) act as a marine highway: through traffic follows the one-way lanes of the traffic separation scheme (TSS) at cruising speed. At Intersections~1 and~2 (regions~1 and~2), vessels leave or join these lanes to enter/exit port waters among crossing traffic. Together, the regions span slow maneuvering in confined water, lane following, and merging and crossing. Each task asks a vessel to reach a goal among surrounding traffic: in an anchorage region, an anchoring position or, for vessels passing through, the point where they leave the region; in a TSS region, the end of its recorded pass. Although we focus on the Singapore Strait, where our dataset is available, nothing in \jns{} is specific to it. A new setting needs only AIS dataset, which follows a standard format worldwide, and charted region outlines. %

\input{sections/03_data}
\input{sections/04_simulator}

\input{sections/045_dynamics}

\input{sections/05_benchmark}

%% file: sections/03_data.tex
\subsection{Goal-reaching navigation task}

We formulate maritime navigation as a goal-reaching problem. Each recorded vessel passage defines one navigation episode. For episode $m$, let $\tau^m = \left(z^m_0,z^m_1,\ldots,z^m_{T_m}\right)$, and $z^m_t = \left(x^m_t,y^m_t,v^m_t,\psi^m_t\right)$
denote the recorded trajectory, where $(x^m_t,y^m_t)$ is the vessel position, $v^m_t$ is its speed, and $\psi^m_t$ is its heading. The controlled vessel is initialized from $z^m_0$, and its recorded terminal position $g^m = \left(x^m_{T_m},y^m_{T_m}\right)$
defines its navigation goal. In an anchorage, this is the anchoring position of an arriving vessel or the exit point of a transiting one; in a TSS or intersection region,  end of recorded passage.

At each step, the policy observes the goal and recent states of the controlled vessel and nearby traffic, and selects an action (see section~\ref{sec:simulator}, action model). An episode succeeds if the vessel enters a radius (e.g., 100\,m) around its destination within the evaluation window. We also evaluate arrival timing, trajectory fidelity, kinematics, and collisions. Surrounding vessels replay their recorded trajectories, or any subset can instead be controlled by learned policies. %

\subsection{Dataset Construction}
\label{sec:data}

Our released dataset covers the eight high-density regions in Figure~\ref{fig:released-regions} and is built from raw 2024 AIS data provided by the Maritime and Port Authority of Singapore (MPA). It contains $\sim${141 million} AIS records from more than {30,000} vessels at around 3 minute intervals.

\textbf{Static vessel information.}
Static records give vessel identifiers, length, beam, and AIS vessel type. Dimensions define each vessel's geometry for collision checking, and type helps selects the controlled vessels, such as cargo ships and tankers, whose size makes them hard to maneuver.

\textbf{Dynamic trajectory information.}
Vessels broadcast AIS reports at irregular intervals of seconds to minutes. Each report gives a timestamp, lat-long, speed and course over ground, heading, and navigational status (e.g., at anchor). As simulation and learning need a uniform time step, we linearly interpolate each vessel's reports to the simulator's 10\,s step, and build region-specific trajectories linked to static vessel information. Each controlled-vessel trajectory comes with the recorded states of surrounding vessels in the same region and time interval; this traffic, to which the recorded vessel responded, is replayed in simulation and used for collision checks.

\textbf{Episode construction }An episode begins when a vessel enters a region and ends at its first observed terminal event. In an anchorage, this event is either an \emph{arrival}, when the vessel comes to a sustained stop inside the region, or a \emph{transit}, when it leaves the region without one; in TSS and intersection regions, it is always a transit. For anchorage arrivals, we keep only vessels that anchored, removing those that stopped alongside another ship to service it (e.g., bunkering). We control large cargo ships and tankers, which are challenging to maneuver, with IL and RL policies, whereas the traffic context also includes vessels of other types. We train on the 273 days of January--September and test on the 92 days of October--December, which evaluates generalization to later traffic within each region (Table~\ref{tab:pop}). \rev{For model selection, we recommend that benchmark users hold out October as a validation month: train on January--September, select on October, and test on November--December (Appendix~\ref{app:sel}).}
Unlike other AIS datasets focused on open-sea passage, our benchmark covers varied close-quarters tasks, such as arrivals and transits among anchored ships, TSS lane following, merging and crossing traffic (Table~\ref{tab:pop}). Appendix~\ref{app:datasheet} documents the different preprocessing steps.

\begin{table}[t]
\centering\small
\setlength{\tabcolsep}{4pt}
\resizebox{\linewidth}{!}{%
\begin{tabular}{lrrrrrrrrrr}
\toprule
& \multicolumn{5}{c}{Anchorage groups} & \multicolumn{5}{c}{Intersection/TSS regions} \\
\cmidrule(lr){2-6}\cmidrule(lr){7-11}
& AEW & AESPA & AEPBC & AWW & \textit{Total} & Int.~1 & Int.~2 & TSS E & TSS W & \textit{Total} \\
\midrule
Train episodes & 22,258 & 20,396 & 11,985 & 7,404 & 62,043 & 107,243 & 92,763 & 68,906 & 26,152 & 295,064 \\
Train transitions (millions) & 1.93 & 2.58 & 1.20 & 0.68 & 6.39 & 18.53 & 11.25 & 28.20 & 10.13 & 68.11 \\
Test episodes & 7,600 & 7,068 & 4,219 & 2,454 & 21,341 & 34,836 & 30,778 & 18,958 & 6,538 & 91,110 \\
Test arrivals & 22\% & 76\% & 54\% & 53\% & --- & --- & --- & --- & --- & --- \\
\bottomrule
\end{tabular}}
\caption{\small {Test arrivals give the share of test episodes that are arrivals; the rest are transits, and intersection/TSS regions have only transits. Models are trained and evaluated separately in each region.}}
\label{tab:pop}
\vskip -10pt
\end{table}

%% file: sections/04_simulator.tex
\subsection{Simulator design}
\label{sec:simulator}

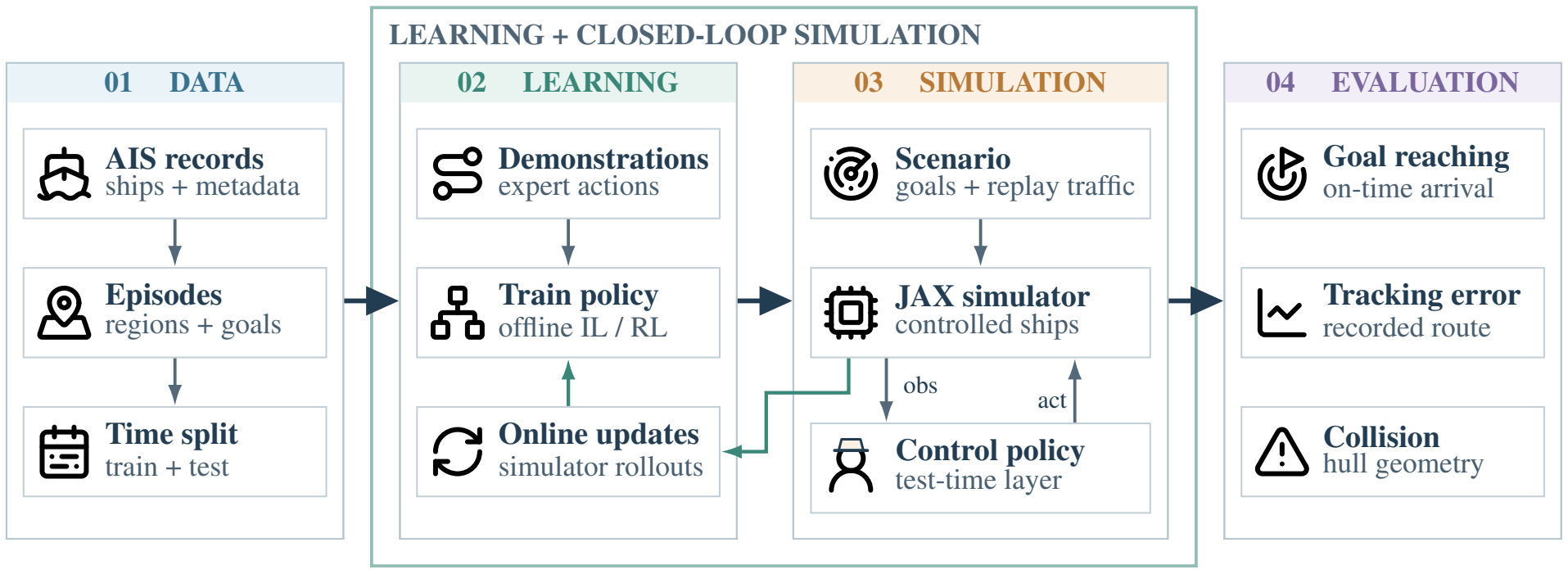
\begin{figure}[t]
\centering
\input{figures/pipeline_architecture}
\vspace{-6pt}
\caption{\small\rev{\textbf{The \jns{} framework.} AIS records are processed into region episodes with goals and split by time. Policies are trained on the expert demonstrations and can be updated online with simulator rollouts. In closed-loop simulation, a policy, optionally with a test-time layer such as PAR (Section~\ref{sec:geo}), controls selected vessels among replayed traffic, and rollouts are scored on goal reaching, tracking error and hull-geometry collisions.}}
\label{fig:framework}
\vskip -10pt
\end{figure}

The simulator is a functional, batched JAX environment that replays recorded traffic around controlled vessels (\rev{Figure~\ref{fig:framework};} Figure~\ref{fig:arch} in Appendix~\ref{app:simulator}).

\textbf{Scenarios and closed-loop control.}
A scenario holds logged trajectories, vessel metadata, control masks, and navigation goals. Let $s_t$ be the simulated scene of $N$ vessels, including observation history, and $\mathcal C\subseteq\{1,\ldots,N\}$ denote the controlled vessels, selected by a per-vessel control mask. Each vessel $i\in\mathcal C$, with local observation $o^i_t=O^i(s_t)$ and goal $g^i$, acts through a policy $\pi_\theta$, possibly shared across vessels, and the scene advances as:
\begin{equation}
    a^i_t\sim\pi_\theta(\cdot\mid o^i_t,g^i)\quad(\forall i\in\mathcal C),\qquad
    s_{t+1}=F\big(s_t,\{a^i_t\}_{i\in\mathcal C};\ell_{t+1}\big),
    \label{eq:sim}
\end{equation}
where $F$ advances the scene by one step, applying the actions under the chosen dynamics model (e.g.,~\eqref{eq:dynamics}) and setting uncontrolled vessels to their recorded states $\ell_{t+1}$. With one controlled vessel we drop $i$. Controlled vessels evolve in closed loop and respond to one another; replayed vessels do not react. A policy sees only local observations and vessel's goal, never future replay samples. The mask selects replay ($\mathcal C=\emptyset$), a single ego, mixed control, or self-play ($\mathcal C=\{1,\ldots,N\}$). \rev{Self-play thus gives reactive traffic, with every vessel acting on its own observations. Rule-based reactive agents \citep[e.g., the Intelligent Sailing Model;][]{krasowski2026ism} on masked vessels are left for future work.} We evaluate a single ego ($|\mathcal C|=1$) in the anchorage regions (Section~\ref{sec:anchorage_results}), where most vessels are stationary, and multiple controlled vessels in the TSS regions (Section~\ref{sec:tss_results}).

\textbf{Observations.}
At step $t$, each controlled vessel observes from its own perspective its current and past $H$ states, the same history of its 10 nearest vessels within 1 nautical mile, its goal, and, when enabled, the local sea current. A state holds position, speed, heading, length, beam, and AIS vessel type (Appendix~\ref{app:vesseltype}). Unobserved vessels stay in the scene and are still checked for collisions.

\textbf{Actions and dynamics model.}
Following \sns{}~\citep{pham2025shipnavisim}, we use a continuous three-dimensional delta action space~\citep{gulino2023waymax}, $a_t=(\Delta x_t,\Delta y_t,\Delta\psi_t)$, the changes in position $p_t=(x_t,y_t)$ and heading $\psi_t$ over one $\Delta t=10$\,s step:
\begin{equation}
 p_{t+1}=p_t+(\Delta x_t,\Delta y_t),\qquad
 \psi_{t+1}=\operatorname{wrap}(\psi_t+\Delta\psi_t),
 \label{eq:dynamics}
\end{equation}
where $\operatorname{wrap}(\theta)=((\theta+\pi)\bmod 2\pi)-\pi$ keeps the heading in $[-\pi,\pi)$; next speed is the distance traveled per step, $v_{t+1}=\sqrt{\Delta x_t^2+\Delta y_t^2}/\Delta t$. A delta action applies to any moving object and needs no vessel-specific dynamics; only loose kinematic bounds (Appendix~\ref{app:actionbound}) constrain it, without enforcing acceleration or other hydrodynamic responses. Other motion models are described later.

\textbf{Rewards and termination.} A shaped reward combines goal progress, a clearance penalty, and goal-arrival, collision, and region-exit terms (Appendix~\ref{app:reward}). Episodes end on goal arrival, region exit, or an exhausted step budget (Appendix~\ref{app:termination}).

\textbf{Accelerator-native execution.}
Trajectories and metadata are structured JAX arrays; reset, step, observation, and reward are pure functions. \texttt{lax.scan} runs rollouts and \texttt{vmap} batches scenarios; simulation and policy inference can stay on the accelerator. Section~\ref{sec:throughput} measures batched stepping with logged actions, a workload distinct from end-to-end policy evaluation (Appendix~\ref{app:compute}).

\textbf{Evaluation metrics.}\label{sec:benchmark} 
We evaluate each controlled vessel over its full rollout; Appendix~\ref{app:metrics} gives the definitions. \textbf{Goal reaching ($\reachg$)} is the fraction of episodes that reach the goal radius within the evaluation window (Appendix~\ref{app:termination}); it measures precise, on-time arrival. \textbf{Collision rate ($C$)} is the fraction of episodes in which the ego's hull, modeled from its length and beam, overlaps another vessel's. \textbf{Closest point of approach (CPA)} is the minimum clearance between the ego's hull and any other hull over the rollout, in metres; it reveals near misses that $C$ does not count. \textbf{Tracking error} is the average (ADE) and final displacement (FDE) error from the recorded track, measured by $\gcade$, following \sns{}, and by FDE$_{\neg R}$, computed \emph{over missed goals only}. \textbf{Kinematics} are per-episode extremes of acceleration, curvature, and course--heading drift (Appendix~\ref{app:kinematics}); they test whether the motion is plausible for a large ship.

%% file: figures/pipeline_architecture.tex
\begingroup
\definecolor{pipeInk}{HTML}{223D52}
\definecolor{pipeLine}{HTML}{B7C7CF}
\definecolor{pipeBlue}{HTML}{EAF3F8}
\definecolor{pipeTeal}{HTML}{E9F4F1}
\definecolor{pipeAmber}{HTML}{FAF1E4}
\definecolor{pipeLilac}{HTML}{F2EEF8}
\definecolor{pipeBlueStrong}{HTML}{39728F}
\definecolor{pipeTealStrong}{HTML}{398778}
\definecolor{pipeAmberStrong}{HTML}{B97936}
\definecolor{pipeLilacStrong}{HTML}{79669C}
\newcommand{\pipeIcon}[3]{%
  \node[inner sep=0pt] at (#2,#3)
    {\includegraphics[page=#1,width=.50cm]{figures/pipeline_icons.pdf}};%
  \ifnum#1=9\relax
    \filldraw[fill=pipeAmber,draw=pipeInk,line width=.6pt,line join=round]
      (#2-.105,#3+.15) -- (#2-.08,#3+.245) --
      (#2+.08,#3+.245) -- (#2+.105,#3+.15) -- cycle;
    \draw[pipeInk,line width=.75pt,line cap=round]
      (#2-.14,#3+.15) -- (#2+.14,#3+.15);
  \fi}
\resizebox{0.9\linewidth}{!}{%
\begin{tikzpicture}[
  x=1cm,y=1cm,
  panel/.style={draw=pipeLine,fill=white,
    minimum height=4.04cm,inner sep=0pt},
  card/.style={draw=pipeLine!85,fill=white,
    minimum height=.76cm,inner sep=0pt},
  title/.style={font=\bfseries\fontsize{7.5}{8.5}\selectfont,
    text=pipeInk,anchor=west,inner sep=0pt},
  sub/.style={font=\scriptsize,text=pipeInk!83,anchor=west,inner sep=0pt},
  arrow/.style={-{Latex[length=1.8mm,width=1.2mm]},
    draw=pipeInk!78,line width=.75pt},
  stagearrow/.style={-{Triangle[length=2.8mm,width=2.4mm]},
    draw=pipeInk,line width=1.15pt},
  feedback/.style={-{Latex[length=1.8mm,width=1.2mm]},
    draw=pipeTealStrong,line width=.85pt}]
  \draw[draw=pipeTealStrong!55,line width=.75pt]
    (2.94,-3.38) rectangle (9.94,1.36);
  \node[font=\scriptsize\bfseries,text=pipeInk!80,anchor=west,inner sep=0pt]
    at (3.08,1.13) {LEARNING + CLOSED-LOOP SIMULATION};

  \node[panel,minimum width=2.86cm] (data) at (1.27,-1.13) {};
  \node[panel,minimum width=2.86cm] (learn) at (4.61,-1.13) {};
  \node[panel,minimum width=3.18cm] (world) at (8.11,-1.13) {};
  \node[panel,minimum width=2.86cm] (eval) at (11.61,-1.13) {};
  \fill[pipeBlue] (-.16,.56) rectangle (2.70,.89);
  \fill[pipeTeal] (3.18,.56) rectangle (6.04,.89);
  \fill[pipeAmber] (6.52,.56) rectangle (9.70,.89);
  \fill[pipeLilac] (10.18,.56) rectangle (13.04,.89);
  \draw[draw=pipeLine] (-.16,-3.15) rectangle (2.70,.89);
  \draw[draw=pipeLine] (3.18,-3.15) rectangle (6.04,.89);
  \draw[draw=pipeLine] (6.52,-3.15) rectangle (9.70,.89);
  \draw[draw=pipeLine] (10.18,-3.15) rectangle (13.04,.89);
  \node[font=\scriptsize\bfseries,text=pipeBlueStrong] at (1.27,.73) {01 \quad DATA};
  \node[font=\scriptsize\bfseries,text=pipeTealStrong] at (4.61,.73) {02 \quad LEARNING};
  \node[font=\scriptsize\bfseries,text=pipeAmberStrong] at (8.11,.73) {03 \quad SIMULATION};
  \node[font=\scriptsize\bfseries,text=pipeLilacStrong] at (11.61,.73) {04 \quad EVALUATION};

  \node[card,minimum width=2.57cm] (ais) at (1.27,-.05) {};
  \pipeIcon{1}{.33}{-.05}
  \node[title] at (.67,.08) {AIS records};
  \node[sub] at (.67,-.17) {ships + metadata};
  \node[card,minimum width=2.57cm] (episodes) at (1.27,-1.23) {};
  \pipeIcon{2}{.33}{-1.23}
  \node[title] at (.67,-1.10) {Episodes};
  \node[sub] at (.67,-1.35) {regions + goals};
  \node[card,minimum width=2.57cm] (split) at (1.27,-2.41) {};
  \pipeIcon{3}{.33}{-2.41}
  \node[title] at (.67,-2.28) {Time split};
  \node[sub] at (.67,-2.53) {train + test};
  \draw[arrow] (ais.south) -- (episodes.north);
  \draw[arrow] (episodes.south) -- (split.north);

  \node[card,minimum width=2.57cm] (demo) at (4.61,-.05) {};
  \pipeIcon{4}{3.67}{-.05}
  \node[title] at (4.01,.08) {Demonstrations};
  \node[sub] at (4.01,-.17) {expert actions};
  \node[card,minimum width=2.57cm] (policy) at (4.61,-1.23) {};
  \pipeIcon{5}{3.67}{-1.23}
  \node[title] at (4.01,-1.10) {Train policy};
  \node[sub] at (4.01,-1.35) {offline IL / RL};
  \node[card,minimum width=2.57cm] (online) at (4.61,-2.41) {};
  \pipeIcon{6}{3.67}{-2.41}
  \node[title] at (4.01,-2.28) {Online updates};
  \node[sub] at (4.01,-2.53) {simulator rollouts};
  \draw[arrow] (demo.south) -- (policy.north);
  \draw[feedback] (online.north) -- (policy.south);

  \node[card,minimum width=2.88cm] (scenario) at (8.11,-.05) {};
  \pipeIcon{7}{7.01}{-.05}
  \node[title] at (7.38,.08) {Scenario};
  \node[sub] at (7.38,-.17) {goals + replay traffic};
  \node[card,minimum width=2.88cm] (sim) at (8.11,-1.23) {};
  \pipeIcon{8}{7.01}{-1.23}
  \node[title] at (7.38,-1.10) {JAX simulator};
  \node[sub] at (7.38,-1.35) {controlled ships};
  \node[card,minimum width=2.88cm] (captain) at (8.11,-2.55) {};
  \pipeIcon{9}{7.01}{-2.55}
  \node[title] at (7.38,-2.42) {Control policy};
  \node[sub] at (7.38,-2.67) {test-time layer};
  \draw[arrow] (scenario.south) -- (sim.north);
  \draw[arrow] ([xshift=.80cm]captain.north) --
    ([xshift=.80cm]sim.south);
  \draw[arrow] ([xshift=-.80cm]sim.south) --
    ([xshift=-.80cm]captain.north);
  \node[font=\tiny,text=pipeInk,anchor=west,inner sep=0pt]
    at (7.45,-1.84) {obs};
  \node[font=\tiny,text=pipeInk,anchor=west,inner sep=0pt]
    at (8.59,-1.98) {act};

  \draw[feedback] ([xshift=-1.12cm]sim.south) --
    (6.99,-1.91) -- (6.29,-1.91) -- (6.29,-2.41) --
    (online.east);

  \node[card,minimum width=2.57cm] (goal) at (11.61,-.05) {};
  \pipeIcon{10}{10.67}{-.05}
  \node[title] at (11.01,.08) {Goal reaching};
  \node[sub] at (11.01,-.17) {on-time arrival};
  \node[card,minimum width=2.57cm] (track) at (11.61,-1.23) {};
  \pipeIcon{11}{10.67}{-1.23}
  \node[title] at (11.01,-1.10) {Tracking error};
  \node[sub] at (11.01,-1.35) {recorded route};
  \node[card,minimum width=2.57cm] (contact) at (11.61,-2.41) {};
  \pipeIcon{12}{10.67}{-2.41}
  \node[title] at (11.01,-2.28) {Collision};
  \node[sub] at (11.01,-2.53) {hull geometry};

  \draw[stagearrow] (data.east) -- (learn.west);
  \draw[stagearrow] (learn.east) -- (world.west);
  \draw[stagearrow] (world.east) -- (eval.west);
\end{tikzpicture}%
}%
\endgroup

%% file: sections/045_dynamics.tex
\newlength{\ladderH}
\subsection{Additional Dynamics and Weather Conditions}

\begin{figure}[tb]
\centering
\begin{minipage}[b]{0.49\textwidth}
  \centering
  \input{figures/delta_vectors}
\end{minipage}
\hfill
\begin{minipage}[b]{0.49\textwidth}
  \centering
  \setlength{\ladderH}{0.2270\linewidth}%
  \begin{minipage}[c]{0.84\linewidth}
    \includegraphics[width=\linewidth]{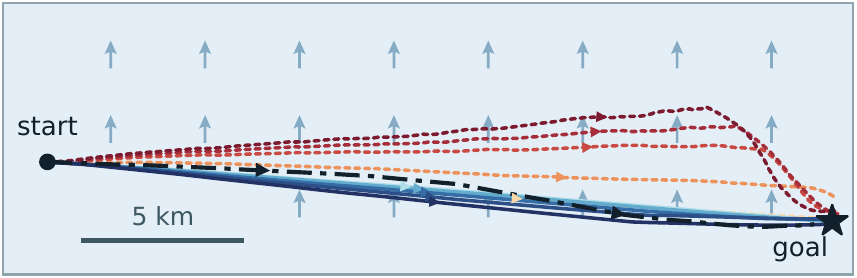}
  \end{minipage}\hfill
  \begin{minipage}[c]{0.14\linewidth}
    \centering
    \includegraphics[height=\ladderH]{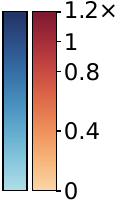}
  \end{minipage}\\[0.4em]
  \includegraphics[width=0.95\linewidth]{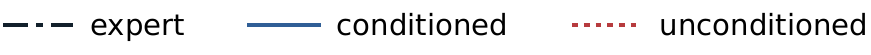}
\end{minipage}
\caption{\small Delta dynamics with the external effect and sea current-conditioned policy behavior.}
\label{fig:drift-current}
\vskip -15pt
\end{figure}

\label{sec:dynamics}
\textbf{External conditions.} In the delta action space (\eqref{eq:dynamics}), each action is simply the change between two consecutive points in a recorded ship track~\citep{pham2025shipnavisim}. This is convenient, but large vessels have no actuator on the sway axis: their propeller and rudder provide surge and yaw, but no commanded sway~\citep{do2010underactuated}. We next model what does move a ship sideways. Surface current is sampled
from the HYCOM reanalysis~\citep{chassignet2007hycom} at each vessel's own
position and time, giving a displacement $\delta_k$ per step. Figure~\ref{fig:hycom} in the appendix shows what the
sea-current field looks like.
The left panel of Figure~\ref{fig:drift-current} illustrates the forward dynamics: the commanded
displacement (blue) and heading change (green) combine with current drift
(red) to produce the vessel's movement over ground (grey).
Expert actions are obtained by inverting these dynamics
(Appendix~\ref{app:dynamics}), solving for the command that, together with the
external effect $\delta_k$, produces the observed step. Taking that action therefore lands the vessel on the next AIS
sample exactly. In the right panel of
Figure~\ref{fig:drift-current}, the blue
spectrum is the current-conditioned policy and the red spectrum the
unconditioned one, each shown under multipliers from zero to $1.2\times$ the
reference sea current field. The conditioned policy stays close to the expert
track, while the unconditioned one is pushed farther upward as the current
strengthens but still turns back toward the goal. Conditioning on the sea current therefore
helps the policy stay close to the expert trajectory.

\textbf{Additional dynamics models.}
The simulator provides four models: \emph{Delta} adds drift to
Equation~\ref{eq:dynamics}; \emph{Port-glide} removes the sway command and uses
the AIS heading; \emph{Water-track} uses the direction of travel through the
water; and \emph{Nomoto} applies a first-order steering lag
\citep{nomoto1957steering}. Of the four, only Port-glide cannot reproduce a
recorded track, because it has no way to produce the sideways part of the
motion, so replaying its extracted actions no longer returns the expert to its
own track (Table~\ref{tab:dynamics}). We therefore use Water-track, which
derives the heading from the vessel's own motion through the water. We refer
the reader to Appendix~\ref{app:dynamics} for the full derivation of each model
and how it changes the demonstrations.

%% file: figures/delta_vectors.tex
\begin{tikzpicture}[vecpic]
  \coordinate (dtend) at (2.75,0.80);
  \hullglyph{20}
  \node[lbl,black!60] at (-0.10,-0.62) {$p_k$};
  \hullatc{(dtend)}{58}
  \node[lbl,black!60] at ($(dtend)+(-0.05,-0.60)$) {$p_{k+1}$};
  \draw[bow] (0,0) -- (20:1.15);
  \draw[vec,vecact]   (0,0) -- (2.0,1.35)
        node[lbl,vecact,above left=-2pt] {$(\Delta x_k,\Delta y_k)$};
  \draw[vec,vecdrift] (2.0,1.35) -- (dtend)
        node[lbl,vecdrift,above=0pt,pos=0.45] {$\delta_k$};
  \draw[vec,vecgnd]   (0,0) -- (dtend);
  \draw[turn] (20:0.72) arc (20:58:0.72);
  \node[lbl,vecturn] at (64:1.15) {$\Delta\psi_k$};

  \begin{scope}[shift={(3.35,1.05)},font=\tiny]
    \draw[vec,vecact]   (0,0) -- (0.34,0);
    \node[right=2pt,font=\tiny] at (0.34,0) {commanded displacement};
    \draw[vec,vecturn] (0,-0.34) -- (0.34,-0.34);
    \node[right=2pt,font=\tiny] at (0.34,-0.34) {commanded turn};
    \draw[vec,vecdrift] (0,-0.68) -- (0.34,-0.68);
    \node[right=2pt,font=\tiny] at (0.34,-0.68) {sea-current drift};
    \draw[vec,vecgnd]   (0,-1.02) -- (0.34,-1.02);
    \node[right=2pt,font=\tiny] at (0.34,-1.02) {step over ground};
  \end{scope}
\end{tikzpicture}

%% file: sections/06_baselines.tex
\section{Baseline methods}
\label{sec:baselines}
All IL and RL baselines share a late-fusion encoder architecture~\citep{nayakanti2023wayformer} over the ego history, local neighbors, and goal, used in both the policy and value networks (Figure~\ref{fig:actorcritic} in Appendix~\ref{app:methods}). For most baselines, we also use action chunking, predicting a sequence of future actions rather than one action per step, following its success in IL~\citep{zhao2023act} and offline RL~\citep{li2025qchunking}. Appendix~\ref{app:methods} gives the details of each method.

\textbf{Trajectory-prediction baselines}
\label{sec:tpbaselines}
We evaluate two published, open-source trajectory-prediction models for AIS data: TrAISformer~\citep{nguyen2024traisformer}, a generative transformer over discretized AIS attributes, and GLRP~\citep{yang2025glrp}, which models ship interactions with a graph. Each model forecasts the vessel's future track once at the episode's first step, and the track is executed open-loop without re-planning.

\textbf{Imitation learning.}
Flow-matching behavior cloning (FM-BC) trains the chunk policy with conditional flow matching~\citep{lipman2023flow}, as in recent flow-matching policies~\citep{black2024pi0}. To act, it turns Gaussian noise into an action chunk with a few Euler integration steps along the learned flow. Normalizing-flow (NF) behavior cloning (NF-BC)~\citep{ghugare2025nf} replaces the flow-matching head with a conditional normalizing flow, RealNVP~\citep{dinh2017realnvp}.

\textbf{Offline RL.}
QC-FQL~\citep{li2025qchunking}  learns an action-chunking critic and a one-step actor from the offline data; the actor maximizes the critic's value while being distilled from our frozen FM-BC policy. To further reduce the cost of training an additional distilled policy as well as sampling from the flow-matching policy during training of QC-FQL, we adopt the DEAS~\citep{kim2026deas} critic training objective and the best-of-$N$ (BoN) sampling strategy~\citep{stiennon2020summarize}, where we sample multiple outputs from FM-BC policy and select the action sequence with the highest Q-value. We refer to this method as DEAS-BoN.

\textbf{Online RL.}
ReinFlow~\citep{zhang2025reinflow} fine-tunes FM-BC in the simulator with PPO~\citep{schulman2017ppo} updates and a flow-matching regularizer. HR-PPO~\citep{cornelisse2024hrppo} trains a per-step actor with a KL penalty that keeps it close to FM-BC policy, and PPO trains the same per-step actor without that penalty.

\textbf{Multi-agent RL.}
To test behavior in the multi-agent setting of Section~\ref{sec:tss_results},
we deploy an independent copy of the trained policy on each controlled vessel.
All copies share the same learned weights but act from their own local
observations and goals, so their actions affect a shared scene and the vessels
respond to one another in closed loop.

\textbf{Predictive Action Refinement (PAR)}
\label{sec:geo}
Learned policies reach their goals at high rates but also collide frequently, and neither offline nor online policy improvement lowers their collision rates (Section~\ref{sec:anchorage_results}). \rev{Following the predictive safety filter principle in control~\citep{wabersich2021predictive}, we} therefore \rev{introduce} Predictive Action Refinement (PAR)\rev{, inspired by PDM-Closed~\citep{dauner2023pdm} for autonomous driving, as} a test-time \rev{safety} layer that makes a frozen policy's actions collision-aware without retraining. PAR forecasts observed neighbours at constant velocity over a short future horizon at each step, checks the policy's proposed action sequence against these forecasts, and, if needed, replaces it with an alternative from a small set of slowdowns, turns, and goal-directed sequences. A sequence counts as feasible if the ego's predicted hull never overlaps a forecast neighbour's hull over the horizon. PAR keeps the policy's sequence when it is feasible, unless it makes little progress near the goal; otherwise, it executes the feasible alternative that makes the most progress toward the goal or, if none is feasible, the one with the largest predicted clearance. Appendix~\ref{app:controller} gives the formal definition and further details. \rev{Although PAR is rule-based, we use it in our main experiments because, on the same policies, it reduces collisions more than two other test-time safety methods, cost-guided sampling~\citep{zheng2025diffusion} and MPPI~\citep{williams2017information}, as shown in Appendix~\ref{app:cost}.} \rev{One limitation of PAR is that it assumes neighbour states without large independent step-to-step error. Under independent position noise of 10\,m or more, it loses its collision reduction (Appendix~\ref{app:rob}).}

%% file: sections/06_results.tex
\section{Experiments}
\label{sec:results}
We first measure simulation speed (Section~\ref{sec:throughput}), then evaluate navigation policies in the single-agent anchorage setting (Section~\ref{sec:anchorage_results}) and the multi-agent TSS setting (Section~\ref{sec:tss_results}). We also discuss dynamics-model comparisons in Section~\ref{sec:tss_results}, with details reported in Appendix~\ref{app:dynamics-anchorage}.
\subsection{Simulation speed}
\label{sec:throughput}

\begin{figure}[!htbp]
\vspace{-7pt}
\centering
\includegraphics[width=0.57\linewidth]{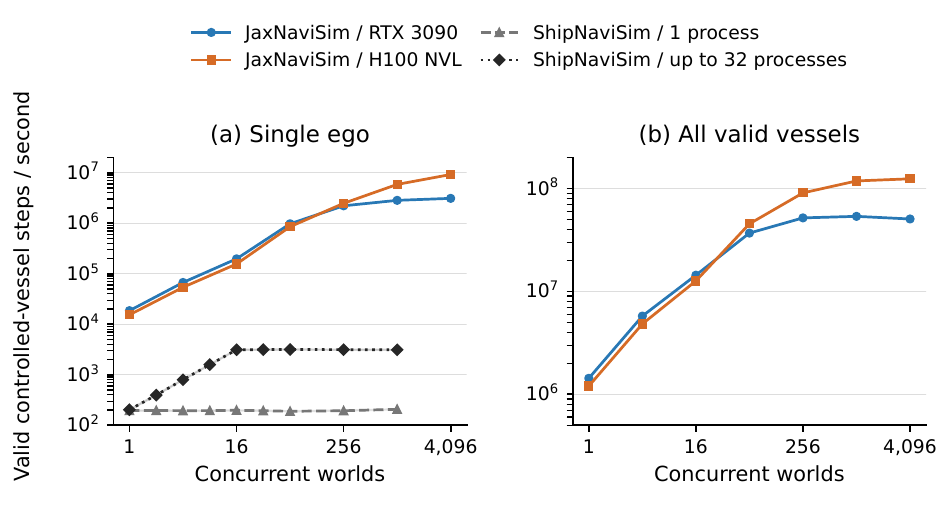}
\vspace{-6pt}
\caption{\small Comparison of simulation throughput with recorded actions\rev{ (transition-only microbenchmark)}: median and minimum--maximum over three repeats at each number of concurrent worlds.}
\label{fig:replaythroughput}
\vskip -10pt
\end{figure}
\rev{As a transition-only microbenchmark,} we measure throughput by stepping all 29,858 AEW training and test scenarios with their recorded actions over up to 4,096 concurrent worlds. Following GPUDrive~\citep{kazemkhani2025gpudrive}, we count valid controlled-vessel steps per second, excluding padded and inactive slots (Appendix~\ref{app:compute}). With one controlled ego per world, \jns{} reaches 3.1 million steps per second on a consumer-grade RTX 3090 and 9.2 million on an H100 NVL, both at 4,096 worlds (Figure~\ref{fig:replaythroughput}a); controlling every valid vessel raises this to 53.6 and 124.2 million, respectively (Figure~\ref{fig:replaythroughput}b). The public CPU implementation of \sns{} reaches 207 ego steps per second in one process and 3,153 with 32 processes, plateauing beyond 16 processes (Figure~\ref{fig:replaythroughput}a, Table~\ref{tab:replaythroughput}). \rev{We also measure a more closely matched replay workload, which consumes the full observation, computes per-step metrics including hull collisions, and applies the CPU reference's termination. It runs at 0.53 million (RTX 3090) and 1.28 million (H100 NVL) steps per second, about 170 and 405 times the 32-process CPU throughput. Closed-loop FM-BC evaluation, with policy inference every step, runs at 22.4 and 56.8 thousand (Appendix~\ref{app:thr}).}

\subsection{Single-agent setting in anchorage regions}
\label{sec:anchorage_results}

\input{tables/anchorage_aew}
\input{tables/anchorage_regions}
\begin{figure}[!htbp]
\vspace{-4pt}
\centering
\includegraphics[width=0.71\linewidth]{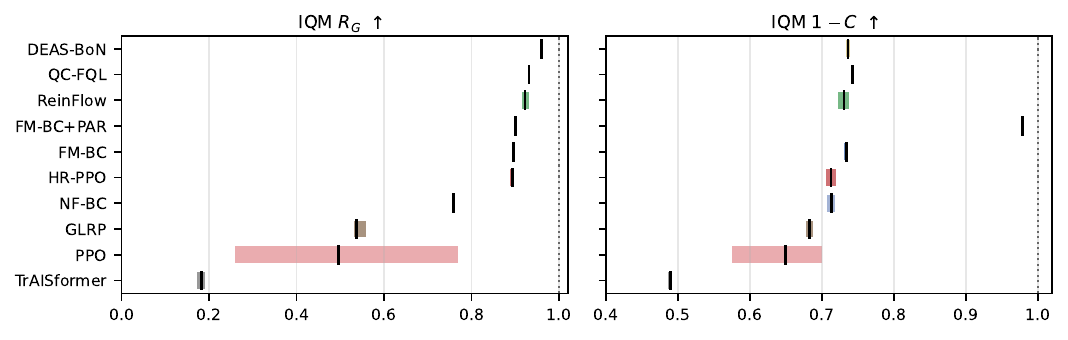}
\vspace{-6pt}
\caption{\small\rev{\textbf{Aggregate over the four anchorage regions} (4 regions $\times$ 3 seeds; test-selected checkpoints). Ticks: interquartile mean (IQM) of $\reachg$ and of the collision-free rate $1-C$; bars: 95\% stratified-bootstrap CIs; dotted line: human replay. Higher is better in both panels, and methods are ordered by IQM $\reachg$.}}
\label{fig:agg}
\vskip -10pt
\end{figure}

We control one vessel among recorded traffic on the full test populations of AEW, AESPA, AEPBC, and AWW, training separate policies with three seeds per region and scoring them with the metrics of Section~\ref{sec:benchmark}. Table~\ref{tab:aew_baselines} compares all baselines on AEW, and Table~\ref{tab:regional_baselines} the main ones on the other three regions; remaining baselines are in Appendix~\ref{app:learningresults}. $\pm$ values are standard deviations over seeds. \rev{All results in this section are \emph{test-selected}: each run's checkpoint was chosen on the test set. Appendix~\ref{app:sel} makes every selection on October instead, retrains the methods that depend on FM-BC, and reports November--December. Under that protocol $\reachg$ is 0.19 points lower on average, and the ranking of two methods changes only where their gap is smaller than the seed standard deviation.}

\textbf{Open-loop trajectory prediction is not enough.}
Executed open-loop, GLRP and TrAISformer reach the goal in only 53.2\% and 22.6\% of AEW episodes, with collision rates of 24.4\% and 50.8\%. GLRP nevertheless has the lowest $\gcade$ in Table~\ref{tab:aew_baselines} (16.84 against 23.18 for FM-BC), and the same holds on the other regions: staying close to the logged track does not imply reaching the goal, which is why closed-loop evaluation matters. Their motion is also implausible for a large ship: the forecasts turn far more sharply than recorded vessels, reaching 5--17 times the maximum curvature of human replay, while all closed-loop policies except unregularized PPO stay below it (Table~\ref{tab:learningkinematics}).

\textbf{Flow-matching imitation is a strong baseline.}
FM-BC reaches the goal in 89.4\% of AEW episodes, against 75.4\% for NF-BC, and has higher $\reachg$ and lower $\gcade$ and collision rate than NF-BC in every region. Its $\reachg$ ranges from 77.7\% on AWW, the region with the smallest training set, to 93.1\% on AESPA. Yet FM-BC collides in 20.4--45.1\% of episodes; on AESPA its median CPA is 8.5\,m, against 76.0\,m for human replay.

\textbf{Policy improvement raises goal reaching, not safety.}
ReinFlow and QC-FQL raise $\reachg$ over FM-BC by 1.7--4.5 and 2.6--8.1 points across the four regions, while their collision rates stay within 1.4 points of FM-BC's. QC-FQL keeps $\gcade$ within 1.1 of FM-BC, whereas ReinFlow increases it (25.48 against 23.18 on AEW). HR-PPO reaches 88.0\% on AEW with higher tracking error, and PPO is unstable across seeds ($33.8\pm47.5\%$). DEAS-BoN attains the highest AEW $\reachg$, 96.7\%, 7.2 points above FM-BC, but its collision rate stays at 20.5\% and its $\gcade$ rises to 31.77. Better task completion therefore does not translate into safer navigation.

\textbf{PAR cuts collisions by an order of magnitude.}
FM-BC + PAR lowers the collision rate to 1.1--3.4\% in all four regions, a 91.1--94.5\% reduction relative to FM-BC. $\reachg$ changes by $+0.1$, $-0.5$, $+0.8$, and $+3.8$ points on AEW, AESPA, AEPBC, and AWW, and $\gcade$ by at most 0.3. Median CPA rises in every region, most on AESPA, from 8.5 to 31.6\,m. On AEW, PAR removes 98\% of FM-BC's collisions with anchored vessels and 76\% of those with moving vessels, which account for most of the residual collisions (Appendix~\ref{app:cases}). 

\rev{\textbf{Aggregate over regions.}
Figure~\ref{fig:agg} summarizes goal reaching and collision avoidance over the four anchorage regions. Following \citet{agarwal2021deep}, we pool the 12 runs of each method (4 regions $\times$ 3 seeds) and report the interquartile mean (IQM) of $\reachg$ and of the collision-free rate $1-C$, with 95\% stratified-bootstrap confidence intervals (Appendix~\ref{app:agg}). The IQM is the mean of the middle half of the runs, so it is less sensitive to outlier runs than the mean. DEAS-BoN, QC-FQL, and ReinFlow have the highest IQM $\reachg$ (0.960, 0.931, and 0.922, against 0.896 for FM-BC), but their IQM $1-C$ stays within 0.01 of FM-BC's (0.734). In contrast, FM-BC + PAR has a similar IQM $\reachg$ (0.901) and raises IQM $1-C$ to 0.979.}

\subsection{Multi-agent setting and additional dynamics-models}
\label{sec:tss_results}

To evaluate multi-agent behavior, we place every qualifying vessel in the four
navigation regions under independent copies of the same trained policy. Each
vessel acts from its own local observations and goal, so all vessels respond to
one another in closed loop. Table~\ref{tab:region_main_fleet} reports the
multi-agent results.
Compared with the same region single-agent navigation results in
Appendix~\ref{app:navresults}, collision rates generally increase when all
vessels are controlled simultaneously. These rates are nevertheless lower than
those in the single-policy anchorage results. These results also suggest future scope for safe multiagent RL and IL methods.
\input{tables/main_fleet}

\textbf{Dynamics models.} We also examine how the dynamics model and sea
current affect policy behavior. Appendix~\ref{app:dynamics-anchorage} compares
the four dynamics models in TSS East and the anchorage regions, including policies
trained without a sea current but tested in a current-enabled simulator and
policies trained with the current. It also separates the effects of including
the sea current information in the simulator, expert action labels, or policy observations,
and reports $\reachg$, collision rate, CPA, and $\gcade$. Across these
comparisons, sea-current conditioning consistently lowers $\gcade$. Models whose
inverse dynamics reproduce the recorded track retain high $\reachg$, whereas
Port-glide's irreducible sideways-motion error causes a substantial drop in
goal reaching.

%% file: tables/anchorage_aew.tex
\begin{table}[!htbp]
 \vspace{-6pt}
\centering \scriptsize
\setlength{\tabcolsep}{4pt}
\sbox{\resulttable}{%
\begin{tabular}{lccccc}
\toprule
Method & $\reachg\uparrow$ & Collision rate $C\downarrow$ & CPA$\uparrow$ (m) & $\gcade\downarrow$ (m) & FDE$_{\neg R}\downarrow$ (m) \\
\midrule
NF-BC & $0.7545\pm0.0003$ & $0.2231\pm0.0028$ & $80.6$ & $26.54\pm0.13$ & $490$ \\
FM-BC & $0.8944\pm0.0057$ & $0.2036\pm0.0036$ & $87.8$ & $23.18\pm0.08$ & $436$ \\
FM-BC + PAR & $0.8955\pm0.0027$ & $\boldsymbol{0.0112\pm0.0007}$ & $\mathbf{90.3}$ & $23.07\pm0.13$ & $470$ \\
ReinFlow & $0.9246\pm0.0059$ & $0.2132\pm0.0095$ & $84.3$ & $25.48\pm0.62$ & $439$ \\
HR-PPO & $0.8796\pm0.0094$ & $0.2200\pm0.0080$ & $80.0$ & $30.25\pm0.86$ & $453$ \\
PPO & $0.3382\pm0.4746$ & $0.4329\pm0.1714$ & $21.1$ & $242.01\pm56.28$ & $3640$ \\
QC-FQL & $0.9199\pm0.0013$ & $0.2025\pm0.0019$ & $87.3$ & $23.57\pm0.15$ & $455$ \\
DEAS-BoN, $N=16$ & $\boldsymbol{0.9665\pm0.0015}$ & $0.2054\pm0.0020$ & $87.2$ & $31.77\pm0.09$ & $375$ \\
\midrule
GLRP (one-shot) & $0.5315\pm0.0048$ & $0.2439\pm0.0024$ & $70.8$ & $\boldsymbol{16.84\pm0.26}$ & $\mathbf{303}$ \\
TrAISformer (one-shot) & $0.2261\pm0.0125$ & $0.5075\pm0.0069$ & $-1.5$ & $38.54\pm1.00$ & $867$ \\
\bottomrule
\end{tabular}}
\ifdim\wd\resulttable>\linewidth\resizebox{\linewidth}{!}{\usebox{\resulttable}}\else\usebox{\resulttable}\fi
\caption{ All baseline families on AEW (7,600 test episodes) under 3 seeds\rev{; checkpoints are test-selected (Appendix~\ref{app:sel}). Bold: best value per metric}.}
\label{tab:aew_baselines}
\vskip -12pt
\end{table}

%% file: tables/anchorage_regions.tex
\begin{table}[!htbp]
\centering \scriptsize
\setlength{\tabcolsep}{4pt}
\renewcommand{\arraystretch}{0.92}
\sbox{\resulttable}{%
\begin{tabular}{llccccc}
\toprule
Region & Method & $\reachg\uparrow$ & Collision rate $C\downarrow$ & CPA$\uparrow$ (m) & $\gcade\downarrow$ (m) & FDE$_{\neg R}\downarrow$ (m) \\
\midrule
AESPA & NF-BC & $0.7961\pm0.0036$ & $0.4897\pm0.0041$ & $1.8$ & $29.52\pm0.02$ & $501$ \\
AESPA & FM-BC & $0.9313\pm0.0004$ & $0.4510\pm0.0016$ & $8.5$ & $26.76\pm0.07$ & $533$ \\
AESPA & FM-BC + PAR & $0.9259\pm0.0043$ & $\boldsymbol{0.0336\pm0.0025}$ & $\mathbf{31.6}$ & $27.05\pm0.01$ & $571$ \\
AESPA & ReinFlow & $0.9482\pm0.0201$ & $0.4638\pm0.0228$ & $6.5$ & $30.00\pm0.34$ & $540$ \\
AESPA & QC-FQL & $\boldsymbol{0.9616\pm0.0014}$ & $0.4461\pm0.0020$ & $9.4$ & $27.85\pm0.09$ & $684$ \\
AESPA & GLRP (one-shot) & $0.5961\pm0.0647$ & $0.5134\pm0.0089$ & $-2.2$ & $\boldsymbol{22.35\pm0.61}$ & $\mathbf{369}$ \\
\midrule
AEPBC & NF-BC & $0.7621\pm0.0031$ & $0.3163\pm0.0106$ & $45.1$ & $26.92\pm0.23$ & $491$ \\
AEPBC & FM-BC & $0.8977\pm0.0064$ & $0.2779\pm0.0018$ & $56.0$ & $24.16\pm0.10$ & $503$ \\
AEPBC & FM-BC + PAR & $0.9058\pm0.0022$ & $\boldsymbol{0.0194\pm0.0019}$ & $\mathbf{63.9}$ & $24.30\pm0.11$ & $565$ \\
AEPBC & ReinFlow & $0.9256\pm0.0182$ & $0.2845\pm0.0078$ & $55.5$ & $26.30\pm0.58$ & $530$ \\
AEPBC & QC-FQL & $\boldsymbol{0.9428\pm0.0021}$ & $0.2746\pm0.0022$ & $57.9$ & $24.62\pm0.05$ & $644$ \\
AEPBC & GLRP (one-shot) & $0.4710\pm0.0278$ & $0.3344\pm0.0017$ & $37.8$ & $\boldsymbol{20.55\pm0.21}$ & $\mathbf{344}$ \\
\midrule
AWW & NF-BC & $0.6989\pm0.0014$ & $0.2569\pm0.0059$ & $76.7$ & $27.56\pm0.06$ & $399$ \\
AWW & FM-BC & $0.7774\pm0.0123$ & $0.2539\pm0.0061$ & $79.9$ & $27.27\pm0.10$ & $417$ \\
AWW & FM-BC + PAR & $0.8150\pm0.0067$ & $\boldsymbol{0.0227\pm0.0039}$ & $84.2$ & $27.52\pm0.09$ & $486$ \\
AWW & ReinFlow & $0.8219\pm0.0145$ & $0.2547\pm0.0133$ & $80.6$ & $28.40\pm0.03$ & $416$ \\
AWW & QC-FQL & $\boldsymbol{0.8583\pm0.0031}$ & $0.2399\pm0.0033$ & $\mathbf{84.3}$ & $26.39\pm0.18$ & $404$ \\
AWW & GLRP (one-shot) & $0.5662\pm0.0306$ & $0.3010\pm0.0103$ & $59.8$ & $\boldsymbol{20.35\pm0.54}$ & $\mathbf{287}$ \\
\bottomrule
\end{tabular}}
\ifdim\wd\resulttable>\linewidth\resizebox{\linewidth}{!}{\usebox{\resulttable}}\else\usebox{\resulttable}\fi
\caption{Baselines on the other anchorage regions, AESPA (7,068 test episodes), AEPBC (4,219), and AWW (2,454), under 3 seeds\rev{; checkpoints are test-selected (Appendix~\ref{app:sel}). Bold: best value per region and metric}.}
\label{tab:regional_baselines}
\vskip -10pt
\end{table}

%% file: tables/main_fleet.tex
\begin{table}[t]
\centering\small
\setlength{\tabcolsep}{4pt}
\sbox{\resulttable}{%
\begin{tabular}{llcccccc}
\toprule
Region & Method & $\reachg\uparrow$ & Collision rate $C\downarrow$ & CPA$\uparrow$ (m) & $\gcade\downarrow$ & FDE$_{\neg R}\downarrow$ (m) \\
\midrule
Intersection 1 & BC-NF & $0.8777\pm0.0145$ & $0.1628\pm0.0189$ & $160.6\pm12.9$ & $65.56\pm0.94$ & $713$ \\
Intersection 1 & BC-FM & $0.9463\pm0.0474$ & $0.1328\pm0.0176$ & $198.0\pm18.0$ & $55.32\pm3.17$ & $221$ \\
Intersection 1 & Independent PPO & $0.9331\pm0.1133$ & $0.6544\pm0.0254$ & $-16.7\pm1.0$ & $266.76\pm11.22$ & $\mathbf{181}$ \\
Intersection 1 & HR-PPO & $\boldsymbol{0.9959\pm0.0012}$ & $\boldsymbol{0.1036\pm0.0022}$ & $\boldsymbol{235.3\pm2.7}$ & $\boldsymbol{53.61\pm0.50}$ & $210$ \\
\midrule
Intersection 2 & BC-NF & $0.7614\pm0.0606$ & $0.1737\pm0.0182$ & $160.6\pm15.5$ & $83.66\pm11.17$ & $1108$ \\
Intersection 2 & BC-FM & $0.9693\pm0.0198$ & $0.1048\pm0.0090$ & $236.4\pm13.2$ & $48.70\pm1.22$ & $187$ \\
Intersection 2 & Independent PPO & $0.6829\pm0.2661$ & $0.5749\pm0.0991$ & $-10.3\pm16.9$ & $213.04\pm27.53$ & $595$ \\
Intersection 2 & HR-PPO & $\boldsymbol{0.9962\pm0.0004}$ & $\boldsymbol{0.0861\pm0.0033}$ & $\boldsymbol{266.4\pm2.0}$ & $\boldsymbol{44.45\pm0.67}$ & $\mathbf{152}$ \\
\midrule
TSS East & BC-NF & $0.8288\pm0.0293$ & $0.1818\pm0.0163$ & $212.0\pm19.6$ & $90.28\pm3.40$ & $1043$ \\
TSS East & BC-FM & $0.9123\pm0.0176$ & $0.1438\pm0.0037$ & $253.0\pm8.1$ & $82.92\pm1.45$ & $196$ \\
TSS East & Independent PPO & $0.9253\pm0.1136$ & $0.6825\pm0.0261$ & $-25.8\pm2.1$ & $477.94\pm4.98$ & $197$ \\
TSS East & HR-PPO & $\boldsymbol{0.9363\pm0.0064}$ & $\boldsymbol{0.1407\pm0.0064}$ & $\boldsymbol{263.0\pm12.9}$ & $\boldsymbol{82.69\pm1.79}$ & $\mathbf{184}$ \\
\midrule
TSS West & BC-NF & $0.7961\pm0.0744$ & $0.1302\pm0.0074$ & $376.2\pm44.3$ & $96.11\pm10.05$ & $1359$ \\
TSS West & BC-FM & $0.9439\pm0.0072$ & $0.1145\pm0.0107$ & $395.4\pm15.1$ & $\boldsymbol{75.14\pm2.61}$ & $\mathbf{205}$ \\
TSS West & Independent PPO & $\boldsymbol{0.9947\pm0.0021}$ & $0.5618\pm0.0075$ & $-15.4\pm2.0$ & $467.38\pm0.34$ & $217$ \\
TSS West & HR-PPO & $0.4320\pm0.1281$ & $\boldsymbol{0.0662\pm0.0174}$ & $\boldsymbol{649.9\pm12.4}$ & $89.87\pm4.76$ & $606$ \\
\bottomrule
\end{tabular}}
\ifdim\wd\resulttable>\linewidth\resizebox{\linewidth}{!}{\usebox{\resulttable}}\else\usebox{\resulttable}\fi
\caption{Four-method comparison under 3 seeds on the four TSS and intersection regions, with every vessel under policy control at the same time. $\reachg$ is scored at a goal radius $r_G=200$\,m. \rev{Bold: best value per region and metric.}}
\label{tab:region_main_fleet}
\vskip -10pt
\end{table}

%% file: sections/08_limitations_conclusion.tex
\section{Conclusion}
\label{sec:limitations}
We presented \jns{}, a data-driven maritime navigation benchmark that combines expert demonstrations, controllable traffic simulation, and evaluation of task completion and behavior across anchorage and TSS settings. Our results show that learned policies reach their goals at high rates but often collide, and that our Predictive Action Refinement (PAR) substantially reduces these collisions. Our work moves toward the realistic simulations port authorities need to assess how scheduling, anchorage, and routing decisions affect throughput and safety.

%% file: sections/09_statements.tex
\subsection*{Reproducibility statement}
Source code to reproduce our experiments is available in the Supplementary Material. The benchmark definition specifies the dataset screens, temporal split, observations, action limits, evaluation windows, and collision geometry. Appendix~\ref{app:methods} gives the learning settings, and Appendices~\ref{app:w40} and~\ref{app:learningresults} report additional results. Per-run configuration records identify the data source, population, and episode protocol used by each evaluation. Reproducing stochastic comparisons requires the same checkpoints, batch plans, random streams, software versions, and hardware configuration. The current measurements use test-set checkpoint selection, as stated in Appendix~\ref{app:methods}. \rev{Appendix~\ref{app:sel} re-selects checkpoints on October and reports November--December.}

\subsection*{Data availability and ethics}
The benchmark's expert AIS demonstrations are publicly available. Section~\ref{sec:data} and Appendix~\ref{app:datasheet} document the population, task construction, and temporal splits. Vessel movement histories and registry attributes can reveal operational information; removing direct identifiers alone does not establish anonymity. \rev{The data are owned by the Maritime and Port Authority of Singapore and are released under its terms of access (non-commercial research and education, no redistribution). These terms permit releasing the vessel identifiers (MMSI and IMO), which the released files keep so that tracks can be linked across days and to registry attributes.} The benchmark is intended for research on simulated navigation and does not validate policies for operational vessel control.

\subsection*{AI use statement}
The research ideas, methodology, and experimental design of this work were led by the authors, who also divided the work among themselves, supervised every stage, and verified the results. We used generative AI tools as assistants: to search for and collect related work, to help implement and debug code under the authors' direction and review, and to revise the manuscript for readability, including checking spelling, the consistency of mathematical notation, LaTeX errors, and the consistency of reported results between the text and tables. AI tools did not generate experimental data; all reported results come from experiments that were actually run and from publicly available published data. The authors take full responsibility for this paper, and no AI tool is an author.

%% file: sections/10_appendix.tex
\section{Limitation}
The anchorage study uses displacement dynamics and replayed neighbours, without reciprocal responses or hydrodynamic validation. The dynamics models of Appendix~\ref{app:dynamics} are kinematic: they assert how a vessel is steered and how the current displaces it, and no coefficient in them is fitted to the recordings. The current gains are set to unit advection rather than estimated, and the surface field is a reanalysis product, not a measurement in these waters. That comparison is a single seed on one region-day, so it separates the dynamics models but does not establish their effect sizes. PAR limits replacements but leaves nominal actions unchanged. Sampled collisions depend on metadata and observation coverage; goal reaching does not establish successful anchoring or regulatory compliance. The four regions belong to one port, with small vessels excluded. Repeated vessel identities, test-set model selection, and method-specific budgets limit broader generalization claims. \rev{Scenes contain only vessels that transmit AIS, so vessels without an AIS transmitter, such as small craft, are absent, and the collision rate $C$ can understate the real-world risk.} \rev{We also do not provide rule-based reactive agents for the replayed vessels, such as the Intelligent Sailing Model~\citep{krasowski2026ism}. They can be added as policies for the vessels marked as controlled, and we leave this to future work.}
\section{Dataset documentation}
\label{app:datasheet}

\paragraph{Purpose and unit of analysis.}
The dataset supports learning and evaluating goal-conditioned vessel navigation within an anchorage group. Each episode corresponds to a trajectory in a daily ego manifest. Its key is the pair consisting of the source-day identifier and trajectory identifier; trajectory identifiers alone are not globally unique. Training and evaluation are separate for AEW, AESPA, AEPBC, and AWW. An episode specifies an initial vessel state, surrounding traffic, and a destination inferred from its observed ending.

\paragraph{Source products.}
The group-wide package contains ego trajectories and manifests, scene trajectories for surrounding vessels, an alongside-analysis table for arrivals, and path-shape statistics. Shared reference files provide vessel particulars and region and destination polygons. Length and beam support capsule geometry, while the AIS ship-type field determines the ego-type screen. Beam resolution prioritizes a quality-controlled vessel table, then the product's vessel breadth, then an AIS-type-group median. Coverage through this procedure does not imply that every dimension is directly measured or that every identifier join is correct.

\paragraph{Scale before benchmark filtering.}
The source collection covers four anchorage groups (AEW, AESPA, AEPBC, and AWW) and four strait regions (INTERSECTION1, INTERSECTION2, TSS\_EAST\_24KM, and TSS\_WEST\_24KM), with 365 daily ego products per region. The released positions have already been extracted from AIS and interpolated at a nominal 10\,s resolution. A regional trajectory segment is an extracted pass through a region; one source trajectory identifier can contribute multiple passes. We take pass counts from the daily extraction metadata. The 2,920 ego files contain 622,758 segments with 447,304,145 samples and 446,681,387 transitions. For segment $i$ with $L_i$ samples, these counts are $N_{\mathrm{sample}}=\sum_i L_i$ and $N_{\mathrm{transition}}=\sum_i\max(L_i-1,0)$. The nominal aggregate trajectory duration is $10N_{\mathrm{transition}}/3600=1{,}240{,}781.6$ vessel-hours, summing $10\max(L_i-1,0)$ seconds per regional segment. This uses the nominal 10\,s interval between successive samples; actual timestamp intervals can differ slightly because of upstream rounding. Counts retain any overlap between regions. The separate surrounding-traffic product contains 2,747,794,743 position records and is excluded from these totals. Transitions exclude boundaries between separate passes, including passes that share a source trajectory identifier.

\paragraph{Vessel identifiers.}
The ego products contain 20,172 distinct International Maritime Organization (IMO) ship identifiers, deduplicated across all days and regions. Every ego row has a seven-digit IMO value that passes the check-digit test; this does not independently verify the identifier against registry records. The released IMO field is joined from a year-wide table keyed by MMSI, so counting IMO values cannot recover vessel distinctions already lost when different ships share an MMSI.

\paragraph{Episode semantics.}
The alongside analysis labels an arrival using the vessel's stopping behaviour and its relation to a nearby companion. The benchmark retains arrivals labelled as anchoring and applies the additional length-ratio rule to serviced arrivals. These are inferred task labels. The source documentation reports 58 manually labelled cases, with agreement on 50 of 54 cases for which a decision was available. This small audit provides limited evidence about the selection heuristic; it is not a representative error estimate for all regions. The transit tortuosity threshold has no corresponding human-labelled validation in the saved evidence.

\paragraph{Episode-selection details.}
We retain episodes lasting 5--60 minutes, with mean speed of at least 1 knot and position-implied step speeds below 40 knots. Position-implied speed is used to identify corrupted position sequences that may coexist with plausible broadcast speed values. For arrivals, we retain records classified as anchoring by the supplied alongside analysis. For arrivals classified as the serviced vessel, we additionally require an ego-to-companion length ratio of at least 1.5 to reduce ambiguous assignments. For transits, we require path tortuosity, defined as the ratio of path length to endpoint displacement, to be at most 1.3. The anchoring and size-ratio rules apply only to arrivals, whereas the tortuosity rule applies only to transits.

\begin{table}[!htbp]
\centering
\small
\setlength{\tabcolsep}{4pt}
\begin{tabular}{lrrr}
\toprule
Region & Episode screen & After vessel-size screen & Final benchmark \\
       & train / test & train / test & train / test \\
\midrule
AEW   & 24,829 / 8,514 & 23,795 / 8,141 & 22,258 / 7,600 \\
AESPA & 22,642 / 7,810 & 21,861 / 7,574 & 20,396 / 7,068 \\
AEPBC & 13,079 / 4,605 & 12,494 / 4,400 & 11,985 / 4,219 \\
AWW   & 11,687 / 3,704 & 9,245 / 3,041  & 7,404 / 2,454 \\
\midrule
Total & 72,237 / 24,633 & 67,395 / 23,156 & 62,043 / 21,341 \\
\bottomrule
\end{tabular}
\caption{Population counts through dataset construction. The first column applies the terminal-event, duration, speed, anchoring, size-ratio, and tortuosity rules. The second removes small and unknown-length vessels from all scenes. The final column gives the benchmark population, restricted to cargo and tanker egos.}
\label{tab:funnel}
\end{table}

\paragraph{Temporal provenance and missing observations.}
The recorded split uses 273 source-day identifiers for training and 92 for testing. The identifiers select files; timestamps specify when observations occurred. The source documentation records that the file named \texttt{20240122} contains observations from the evening of 25 January through the morning of 26 January, and that \texttt{20240125} is absent. This documented case remains within the training period. Scene files also overlap neighbouring files in time, while ego files abut. Consequently, the number of daily products should not be read as a claim of complete calendar-day coverage or of timestamp-level independence between scenes. The package reports gaps on 31 days, with a median blackout duration of 47 minutes. Missing states remain subject to validity masking during evaluation.

\paragraph{Intended scope and access.}
The expert AIS demonstrations are publicly available. The anchorage benchmark represents cargo and tanker navigation among vessels of known length at least 30\,m in four regions of one port. It does not cover the excluded harbour craft or establish transfer to another port. Vessel identities may recur across the temporal split.

\section{Simulator details}

\begin{figure}[!htbp]
\centering\small
\input{figures/pipeline}
\caption{\small The \jns{} pipeline. Recorded AIS trajectories supply offline demonstrations and the replayed traffic of each scenario. Policies are trained offline from demonstrations or online from closed-loop rollouts, in which one or more controlled vessels act among replayed traffic, optionally through a test-time layer. Policies receive only current local observations; only the simulator accesses future replay samples. Section~\ref{sec:benchmark} defines the metrics.}
\label{fig:arch}
\end{figure}
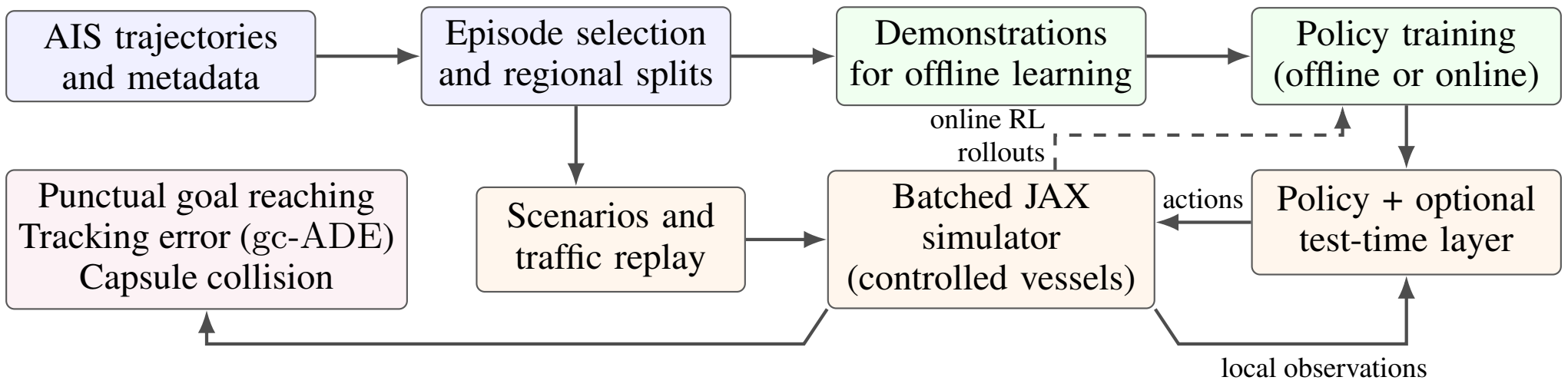

\begin{figure}[!htbp]
    \centering
    \includegraphics[width=0.9\linewidth]{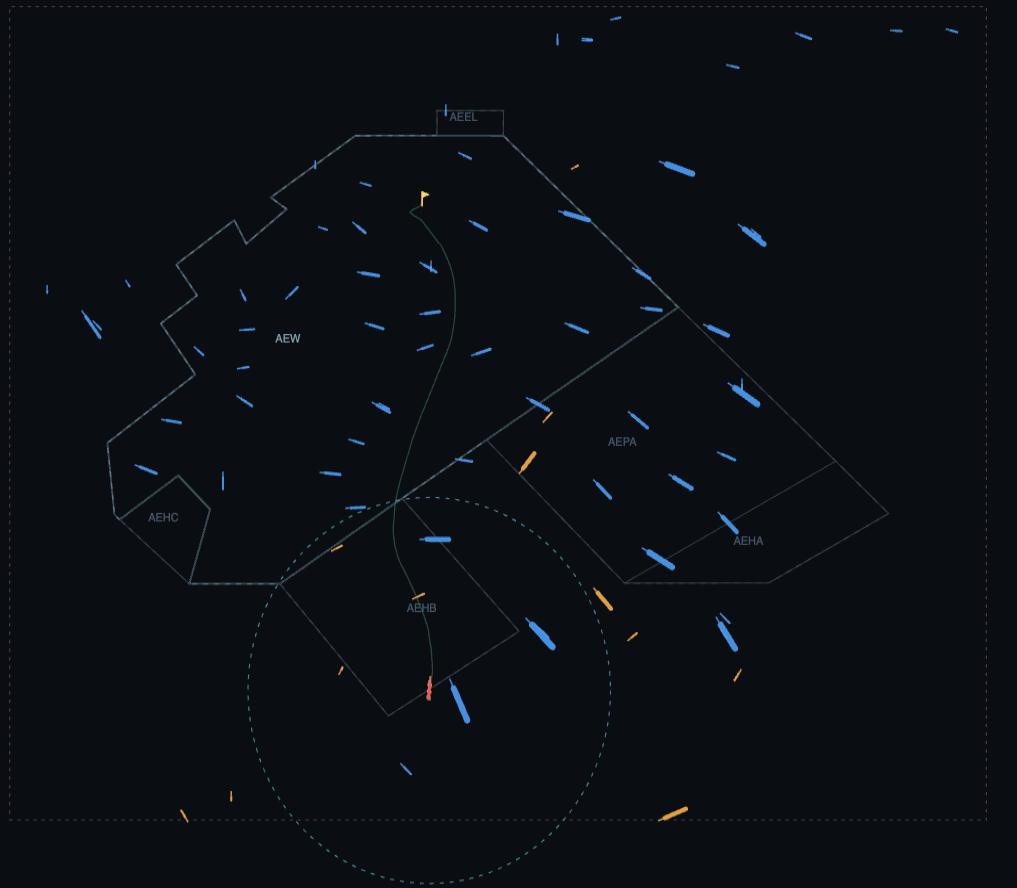}
    \caption{\textbf{Goal-reaching navigation in an anchorage region.}
    The controlled vessel is shown in red and its assigned destination by the yellow flag. The green curve shows the corresponding recorded reference path, while the dashed circle indicates the local traffic-observation range. The surrounding vessel glyphs and anchorage boundaries provide the traffic and geographic context in which the policy must reach the destination.}
    \label{fig:goal-reaching}
\end{figure}
\label{app:simulator}

Figure~\ref{fig:arch} summarizes how recorded data, the simulator, and policy training connect. Figure~\ref{fig:goal-reaching} illustrates an anchorage episode.

\subsection{AIS vessel type}
\label{app:vesseltype}
The AIS ship-and-cargo type is a two-digit code whose tens digit gives the vessel class. The simulator observes the raw code; the policies receive it as a one-hot vector over the seven classes of Table~\ref{tab:vesseltype}, for the controlled vessel and for each observed neighbour. In the anchorage benchmark, the controlled vessels are cargo and tanker vessels (codes 70--89).
\begin{table}[!htbp]
\centering
\small
\begin{tabular}{ll>{\raggedright\arraybackslash}p{6.4cm}}
\toprule
Class & AIS codes & Vessels \\
\midrule
Unknown & 0 & type not available \\
Fishing and operations & 30--39 & fishing, towing, dredging, diving, military, sailing \\
Special & 50--59 & pilot, search and rescue, tug, port tender, law enforcement, medical \\
Passenger & 60--69 & passenger \\
Cargo & 70--79 & cargo \\
Tanker & 80--89 & tanker \\
Other & all other codes & including 1--29, 40--49, 90--99, and out-of-range codes \\
\bottomrule
\end{tabular}
\caption{Coarse vessel classes derived from the AIS ship-and-cargo type.}
\label{tab:vesseltype}
\end{table}

\subsection{Action bounds}
\label{app:actionbound}
The delta action of Equation~\ref{eq:dynamics} is constrained to
\begin{equation}
 \mathcal A=\left\{a:\sqrt{\Delta x^2+\Delta y^2}\le205.78\ \mathrm m,\quad
                         |\Delta\psi|\le\pi/4\right\},
 \label{eq:bound}
\end{equation}
that is, at most 40 knots of displacement and a $45^\circ$ heading change per 10\,s step.

\subsection{Standard shaped reward}
\label{app:reward}
QC-FQL, the critic used for candidate ranking, ReinFlow, PPO, and HR-PPO use one reward weighting. For a transition from $s$ to $s'$, let $I_g$, $I_c$, and $I_o$ indicate goal arrival, a sampled collision under the capsule-based metric, and exit from the region, respectively. The reward is
\begin{equation}
 r(s,s')=0.05\bigl[200I_g+\gamma\Phi(s')-\Phi(s)-20I_c-b(s')-50I_o\bigr],
 \qquad \gamma=0.99,
 \label{eq:standardreward}
\end{equation}
where $\Phi(s)=-\|p^e-g\|/(100\,\mathrm m)$ and terminal potential is zero, including at the step budget. The clearance penalty uses checkable ego--neighbour pairs:
\begin{equation}
 \begin{aligned}
 b(s')&=\max_j\left[\max\left(0,1-\frac{d_{ej}(s')}{m_{ej}(s')}\right)\right]^2,\\
 m_{ej}&=\max\bigl(50\,\mathrm m,\ 2\max(B_e,B_j),\ (60\,\mathrm s)v^{\mathrm{close}}_{ej}\bigr).
 \end{aligned}
 \label{eq:rewardmargin}
\end{equation}
Here $d_{ej}$ is signed capsule clearance, $B$ is beam, and $v^{\mathrm{close}}_{ej}\ge0$ is the closing range rate estimated from simulated history. The penalty is zero if no pair is checkable. The closing-speed margin is not capped; the separate fixed-margin reward option is unused. The clearance, collision, and exit penalties change the objective in addition to the goal-distance potential shaping.

The offline transitions were relabeled with this weighting before training QC-FQL and the regional critics; the recorded observations, actions, and episode boundaries are unchanged. The factor 0.05 is already in the stored offline reward, so QC-FQL's additional reward multiplier is 1. Simulator-trained methods apply the factor during training.

\subsection{Evaluation windows and termination by region}
\label{app:termination}
In the anchorage regions, goal reaching allows a grace period of $G$ steps after the logged demonstration ends: stopping at the last logged sample would treat every delayed arrival as a failure, while an unrestricted horizon can reward excessive delay. The rollout is not stopped at this deadline. It continues for up to $E$ steps past the demonstration, so that a vessel arriving late or not at all is still checked for collisions and its arrival time remains observable:
\begin{equation}
 E=\min(300,T),\qquad G=\min(10,\lfloor0.1T\rfloor).
 \label{eq:windows}
\end{equation}
Let $\tau_g$ be the first step at which the rollout enters the goal radius, with $\tau_g=\infty$ if it never does. An episode counts as a success for $\reachg$ if
\begin{equation}
 \reachg=\ind[\tau_g<T+G].
 \label{eq:reach}
\end{equation}
The grace period is a benchmark tolerance for relative timing, not a port scheduling requirement. Success is position-based: arrivals are not required to stop, align their heading, or keep station, so $\reachg$ does not certify physical anchoring. The rollout stops earlier at goal arrival, region exit, or the global 1{,}000-step budget. After the demonstration ends, surrounding traffic is still replayed from the recorded day, and the collision indicator of Equation~\ref{eq:collision} covers the entire executed rollout. Missing neighbour samples are marked invalid and cannot be collision-checked (Appendix~\ref{app:metrics}).

\paragraph{Goal radius.}
Every run uses a 100\,m goal radius, which defines \emph{reached} for the shaped reward, the termination, and the evaluator alike. Anchorage results are reported at that radius. The navigation regions of Appendix~\ref{app:dynamics} are reported at 200\,m. Those regions cover far more water than the anchorage basins, spanning some 24\,km end to end, and vessels cross them at service speed instead of manoeuvring to a stop at an assigned berth, so a fixed radius is a much tighter target there: a transiting vessel covers an appreciable fraction of 100\,m within a single 10\,s step of Equation~\ref{eq:bound}, which makes arrival at that radius sensitive to sampling rather than to navigation. A 200\,m radius is also the value \sns{}~\citep{pham2025shipnavisim} uses in this TSS setting, so the reported figure remains comparable with prior work. This needs no second run, because reach is radius-free after the fact:
\begin{equation}
 \mathrm{reach}(r_G,W)=\ind\!\left[\min_{t<W} \lVert p_t-g\rVert < r_G\right],
 \label{eq:reachradius}
\end{equation}
so Equation~\ref{eq:reach} is the case $r_G=100$\,m, and the 200\,m figure is an exact re-scoring of the closest approach stored per episode.

\paragraph{The navigation-region horizon.}
The windows above follow a single ego's logged demonstration. The navigation regions instead use a fixed horizon of 1{,}000 steps from each ego's own entry, cut at region exit, and do not stop when that ego's logged demonstration ends. Under joint control the egos enter and leave at different times, so there is no common $T$ to key a window to; using the same fixed horizon for both regimes keeps the two measured on one population. This replaces Equation~\ref{eq:windows} for the navigation regions only.

\section{Metric definitions and implementation conventions}
\label{app:metrics}

\paragraph{Goal-reaching rate ($\reachg$).}
This metric measures the rate at which the ego vessel successfully reaches its goal. We use a goal radius $r_G$ of 100--200\,m, depending on the region; an episode counts as a success if the vessel comes within this radius of its goal within the evaluation window (Equation~\ref{eq:reach}, Appendix~\ref{app:termination}). Compared with the length of tankers and cargo ships (up to around 300\,m), $r_G$ is relatively small.

\paragraph{Goal-reaching index convention.}
For a reference with $T$ samples indexed $0,\ldots,T-1$, the last index counted by $\reachg$ is $T+G-1$, with $G$ from Equation~\ref{eq:windows}. Goal arrival, region exit, and the global step budget can end the rollout sooner. The saved \texttt{reached\_step} is the first goal-hit index, or $-1$ on failure. The saved arrival-delay convention is $\texttt{reached\_step}-T$, so arrival at the last reference sample has value $-1$. This index convention differs by one step from elapsed delay relative to that last sample.

\paragraph{Collision rate.}
We approximate a vessel of length $L_i$ and beam $B_i$ by a capsule: a segment $S_i$ of length $\max(L_i-B_i,0)$ along its heading, expanded by radius $B_i/2$. The signed clearance between vessels $i$ and $j$ is
\begin{equation}
 d_{ij}(t)=\operatorname{dist}(S_i(t),S_j(t))-\frac{B_i+B_j}{2},
 \label{eq:clearance}
\end{equation}
and an episode has a collision if
\begin{equation}
 C=\ind\!\left[\min_{(t,j)\in\mathcal V}d_{ej}(t)\le0\right],
 \label{eq:collision}
\end{equation}
where $e$ is the ego and $\mathcal V$ contains the valid ego--neighbour pairs over the entire executed rollout. Collisions between two replayed neighbours are not attributed to the policy. The metric checks overlap of approximate footprints at 10\,s samples, so it can miss contacts between samples and does not measure regulatory responsibility.

\paragraph{Closest point of approach (CPA).}
The CPA of an episode is its minimum signed clearance $\min_{(t,j)\in\mathcal V}d_{ej}(t)$, measured along the executed rollout rather than predicted from initial velocities. We report its median over episodes in metres.

\paragraph{Collision-evaluation window and observation access.}
The headline collision rate covers the full executed rollout, including the initial sample and any extension beyond the demonstration. It uses valid ego--neighbour pairs in the simulated scene, which can contain vessels outside the policy's ten-neighbour observation. This distinction matters for the controller: its forecast uses only locally observed neighbours, whereas the evaluator can detect collisions with other valid scene vessels. No interpolation is applied between native 10\,s samples. A missing neighbour sample cannot contribute a checked pair at that step. These conventions limit the interpretation of a zero collision indicator.

\paragraph{Tracking error ($\gcade$, FDE$_{\neg R}$).}
Following \sns{}, $\gcade$ compares the simulated and logged ego positions $\hat p_t$ and $p_t$ over the $n=|\mathcal I|$ valid steps $\mathcal I$ of the logged demonstration, evaluated only while a valid reference comparison exists:
\begin{equation}
 \gcade=\frac{\sqrt{\sum_{t\in\mathcal I}\|\hat p_t-p_t\|_2^2}}{n}
       =\frac{\mathrm{RMSE}}{\sqrt n}.
\end{equation}
Thus, equal RMSE can produce different $\gcade$ values when the comparison length changes. The main tables retain this legacy quantity for continuity with the updated benchmark. FDE is measured at the final compared sample; FDE$_{\neg R}$ averages it over episodes that miss the goal deadline ($\reachg=0$). If a rollout terminates early, its tracking score does not by itself penalize the missing remainder; goal reaching and termination must be considered alongside tracking.

\paragraph{Kinematics.}
We report per-episode extrema of acceleration, heading-based curvature, and course--heading drift; Appendix~\ref{app:kinematics} gives their conventions.

\paragraph{Aggregation.}
Each per-seed rate is the mean of binary episode indicators over the complete regional test set. Tables report the arithmetic mean across three seeds, with sample standard deviation where shown. The same episodes occur under each seed, so this standard deviation is not uncertainty over independent traffic populations. We do not pool the four regions into one leaderboard score. \rev{Appendix~\ref{app:agg} reports per-method aggregates over regions and seeds, keeping $\reachg$ and $1-C$ separate.}

All FM-BC rows use the checkpoint-sweep evaluation, and FM-BC + PAR runs the same selected FM-BC checkpoints on the same test episodes. Deterministic XLA settings improve repeatability on a fixed configuration; exact identity across GPU models is not assumed. Standard deviations across three policy seeds describe variation between the trained policies and their stochastic evaluations; they are not confidence intervals over independent real-world deployments. DEAS-BoN shares one regional critic across these seeds.

\section{Learning settings and data access}
\label{app:methods}

\begin{figure}[!htbp]
\centering
\resizebox{0.9\linewidth}{!}{%
  \input{figures/actor_critic_arch}%
}
\caption{Actor (FM-BC, left) and critic (DEAS-BoN, right) networks. Both use the same late-fusion encoder architecture, with separate parameters. Dashed lines expand a block into its layers, and $\times n$ marks a group of layers repeated $n$ times. The symbols are defined in the paragraph ``Network architecture''.}
\label{fig:actorcritic}
\end{figure}
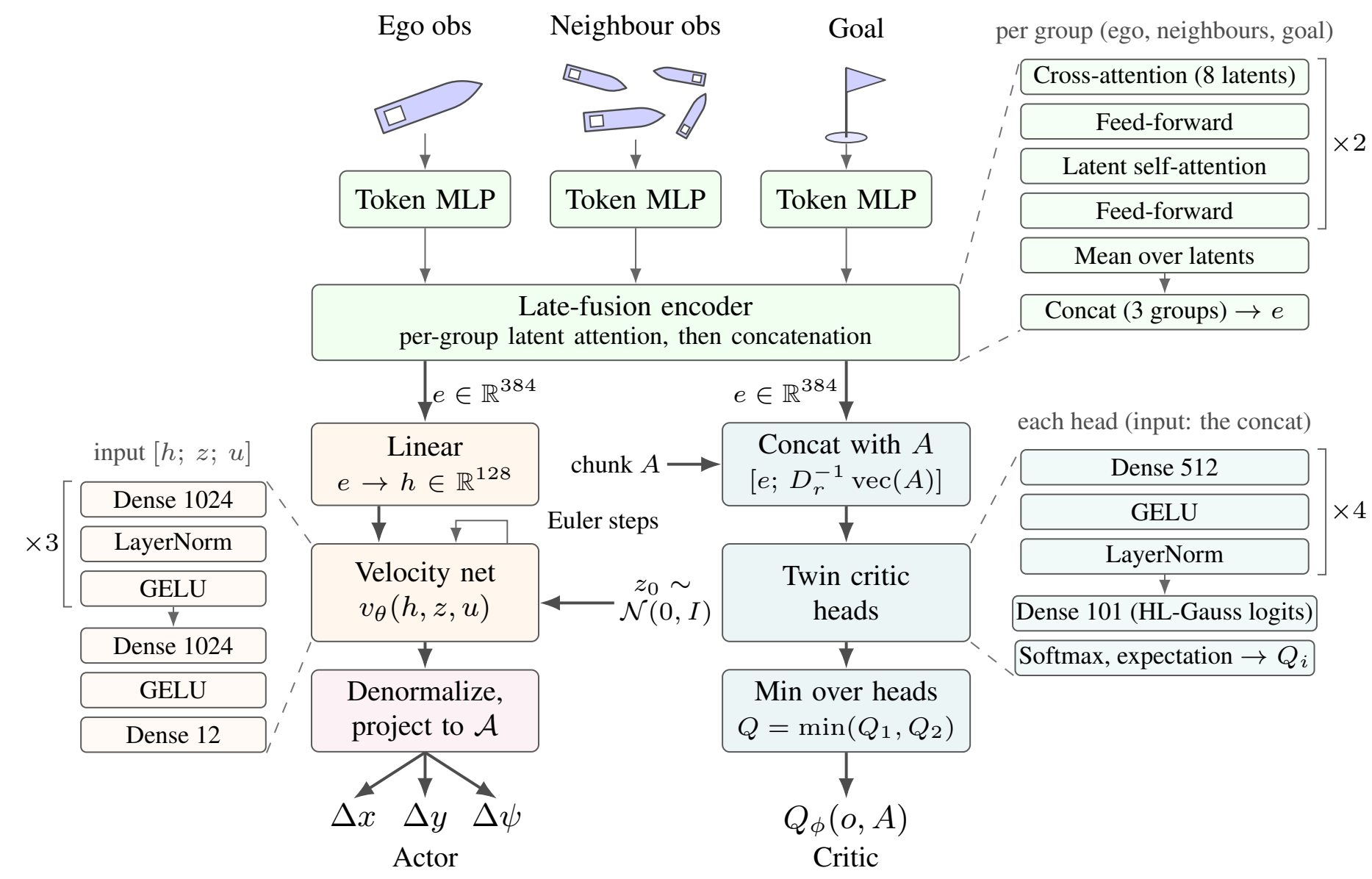

\paragraph{Network architecture.}
Figure~\ref{fig:actorcritic} shows the FM-BC actor and the DEAS-BoN critic; each has its own encoder of the same architecture~\citep{nayakanti2023wayformer}.
\textit{Encoder.} The observation $o$ is split into three token groups: the ego vessel, its ten nearest neighbours (each with its current and previous state), and the goal. A token MLP embeds the tokens of each group. Each group then passes separately through two blocks, each consisting of cross-attention from 8 learned latent vectors to the group's tokens, a feed-forward layer, self-attention among the latents, and a second feed-forward layer. The latents of each group are averaged into a 128-dimensional vector, and the three group vectors are concatenated into the representation $e\in\mathbb R^{384}$.
\textit{Actor.} A linear layer maps $e$ to $h\in\mathbb R^{128}$. The actor generates a chunk $A$ of $k=4$ actions $(\Delta x,\Delta y,\Delta\psi)$ (Equation~\ref{eq:dynamics}). Let $\operatorname{vec}(A)\in\mathbb R^{12}$ flatten the chunk, and let $D_r$ be the diagonal matrix of the componentwise action normalizers of region $r$ (Table~\ref{tab:normalizers}), repeated over the four actions; the flow then operates on the normalized chunk $z=D_r^{-1}\operatorname{vec}(A)$. The velocity network $v_\theta(h,z,u)$, with parameters $\theta$, is an MLP on the concatenation $[h;z;u]$, where $u\in[0,1]$ is the flow time. It has four hidden layers of width 1,024, the first three with layer normalization before the GELU activation, and a 12-dimensional output layer (Dense~12). Because $h$ is computed from $o$, Equation~\ref{eq:fm} writes the network as $v_\theta(o,z,u)$. Sampling starts from Gaussian noise $z_0\sim\mathcal N(0,I)$ and applies $K=3$ Euler steps, $z\leftarrow z+v_\theta(h,z,u)/K$ at $u=0,1/K,2/K$. The result is denormalized to $D_r z$ and reshaped into four actions, each of which is projected into the bounded action set $\mathcal A$ (Equation~\ref{eq:bound}).
\textit{Critic.} The critic, with parameters $\phi$, estimates the value $Q_\phi(o,A)$ of executing chunk $A$ after observation $o$. Its input is the concatenation $[e;\,D_r^{-1}\operatorname{vec}(A)]$ of its encoder output and the normalized chunk. Two heads process this input separately. Each applies four blocks of a 512-unit dense layer, GELU activation, and layer normalization, followed by a dense layer with 101 outputs: the logits of a distribution over 101 fixed value bins (the HL-Gauss head described under DEAS-BoN below). The expectation of this distribution is the head's value $Q_i$, $i\in\{1,2\}$, and the critic returns the smaller of the two, $Q_\phi(o,A)=\min(Q_1,Q_2)$, which limits value overestimation.

\paragraph{FM-BC objective and sampling.}
The FM-BC policy predicts a chunk $A=(a_t,\ldots,a_{t+k-1})$ with $k=4$. With $D_r$ and $\operatorname{vec}$ as above, let $z_1=D_r^{-1}\operatorname{vec}(A)$ be a normalized expert chunk. Conditional flow matching~\citep{lipman2023flow} trains a velocity network with
\begin{equation}
 \mathcal L_{\mathrm{FM}}(\theta)=
 \mathbb E_{\substack{(o,A)\sim\mathcal D_r,\ z_0\sim\mathcal N(0,I)\\u\sim\mathcal U[0,1]}}
 \left[\left\|v_\theta(o,(1-u)z_0+uz_1,u)-(z_1-z_0)\right\|_2^2\right].
 \label{eq:fm}
\end{equation}
Sampling uses three Euler updates from Gaussian noise, followed by denormalization and projection into $\mathcal A^k$. Four actions execute before resampling.

\begin{table}[!htbp]
\centering
\small
\begin{tabular}{ll}
\toprule
FM-BC setting & Value \\
\midrule
Encoder & Wayformer-style late fusion \\
Representation width & 128 \\
Observed history & Current and one preceding state \\
Observed neighbours & Ten nearest valid vessels within 1 nautical mile \\
Additional vessel features & AIS type and beam \\
Action chunk / execution & Four actions / four steps \\
Velocity network & Four hidden layers of width 1,024 \\
Hidden-layer normalization & Layer normalization before activation \\
Training flow time & Uniform on $[0,1]$ \\
Sampling & Three Euler updates from Gaussian noise \\
Gradient updates / batch size & 2,000,000 / 512 \\
Learning rate & $10^{-4}$, cosine decay to the training horizon \\
$L_2$ regularization & $10^{-6}$ \\
Exponential moving average & 0.999 \\
Input normalization & Running statistics \\
Action normalization & Regional training-action absolute p99.9 \\
Action projection & 40-knot displacement disc; $45^\circ$ heading increment \\
Seeds & 0, 1, 2 \\
Retained checkpoints & 1.91M--2.00M updates, spaced by 10,000 \\
Selection & Highest test $\reachg$; lower $\gcade$ breaks ties \\
\bottomrule
\end{tabular}
\caption{Common FM-BC settings. The same recipe is applied separately to each regional dataset. A shared recipe does not imply equal numbers of passes through datasets of different sizes.}
\label{tab:fmsettings}
\end{table}

\begin{table}[!htbp]
\centering
\small
\begin{tabular}{lrrr}
\toprule
Region & $|\Delta x|$ (m) & $|\Delta y|$ (m) & $|\Delta\psi|$ (rad) \\
\midrule
AEW & 105.17 & 93.093 & 0.21136 \\
AESPA & 106.02 & 76.033 & 0.24734 \\
AEPBC & 108.56 & 57.422 & 0.22246 \\
AWW & 94.994 & 81.945 & 0.16107 \\
\bottomrule
\end{tabular}
\caption{The componentwise 99.9th-percentile action normalizers. These quantities scale learning targets; the physical action bound is applied after denormalization.}
\label{tab:normalizers}
\end{table}

The three Euler steps in Table~\ref{tab:fmsettings} are the evaluation setting used for the reported FM-BC sweeps. They override the generic sampler default stored in the training configuration; reproducing a result requires the evaluation arguments as well as the training configuration.

\paragraph{DEAS-BoN.}
A critic for each region is trained on reward-relabeled expert transitions, with the policy frozen, using expectile 0.9, discount 0.99, batch size 512, and one million gradient updates.
The critic uses an HL-Gauss value head~\citep{farebrother2024stop}: each scalar TD target is smoothed into a Gaussian-weighted histogram over 101 fixed bins and fitted by cross-entropy, and the predicted value is the expectation over bin centres.
At test time, best-of-$N$ selection draws $N$ candidate chunks from the policy and executes the one with highest predicted value:
\begin{equation}
 A^\star=\arg\max_{A^{(i)},\ i=1,\ldots,N} Q_\phi(o,A^{(i)}).
 \label{eq:qselect}
\end{equation}

\paragraph{NF-BC and QC-FQL.}
NF-BC uses a conditional RealNVP head with six coupling layers, the common late-fusion encoder, the same action chunks, and two million gradient updates at batch size 512. QC-FQL uses one million updates at batch size 512; actor and critic learning rates are $10^{-4}$ and $3\times10^{-4}$, respectively. The frozen teacher is the selected FM-BC model of the same region and seed \rev{(test-selected for the results of Section~\ref{sec:anchorage_results}; Appendix~\ref{app:sel} reports runs with the October-selected checkpoint)}. The one-step actor is trained against the teacher's three-step Euler output at a shared Gaussian latent, with distillation weight 100 and an inverse-mean-absolute-$Q$ normalization on the value term. Two critics are trained jointly with the actor, using their minimum for the actor objective and target. The target uses the learned actor's next chunk, discounted by $0.99^4$, and a Polyak update coefficient of 0.005. Our implementation is an adaptation rather than the published QC-FQL or FQL recipe~\citep{li2025qchunking,park2025fql}: the FM-BC teacher stays frozen rather than being trained jointly, training is purely offline, the value head is HL-Gauss, and the distillation weight and reward relabeling are our choices. Evaluation samples fresh latent noise; it is not zero-noise evaluation. The one-seed $\alpha=1000$ and frozen-critic pilots are not additional seeds of the reported method.

\paragraph{Simulator-trained methods.}
Our ReinFlow adaptation is identified as R0 in the saved run configurations. It updates the flow policy using a stochastic sampling chain and an expert flow-matching regularizer; its reported evaluation uses the deterministic Euler integrator with a sampled initial latent. HR-PPO trains a per-step actor with a teacher regularizer of weight 0.1: a KL penalty toward FM-BC, estimated as the actor's negative log-likelihood of actions sampled from FM-BC. It uses zero additional action noise at evaluation. Each method uses the corresponding seed's FM-BC teacher. The teacher remains a four-action model even when the HR-PPO adaptation queries its first action at each step. PPO uses the same per-step actor recipe with the teacher penalty removed and zero additional evaluation noise. These methods differ in policy parameterization and execution frequency as well as optimization, and are not a single-factor algorithm comparison. We evaluate all three on every region, with method-specific checkpoint schedules.

\paragraph{Interaction budget and population.}
The saved AEW runs complete 1,525 ReinFlow iterations and 762 HR-PPO iterations, with their selected iterations shown in Appendix~\ref{app:learningresults}. Simulator training draws from training-period scenario pools with a reference-length eligibility cap of 256 samples and an extended storage horizon of 512 samples. This population restriction differs from offline BC's full expert dataset. Table~\ref{tab:rlsettings} records their different update schedules; equal checkpoint counts do not imply equal computation or interaction budgets.

\paragraph{Trajectory-prediction baselines.}
Each forecaster predicts a 360-step (60\,min) track, which is converted to displacement actions and projected into $\mathcal A$ before execution. Both forecasters are run from the authors' code and read the same 180-step (30\,min) history, repeat-padded because a full history exists for only 12.5\% of AEW episodes. The 360-step horizon is longer than either model was published at. TrAISformer uses our vocabulary rather than the upstream latitude--longitude box: region-local metres on a 25\,m grid, 0.25\,m/s of speed, and $5^\circ$ of heading. Its goal enters as two extra token positions at the head of the causal sequence, one per coordinate bin. At 58.5M parameters it is 11.7$\times$ our policy and is run as published and unshrunk; the executed plan is its greedy decode. GLRP's scene is the ego and its ten exported neighbours, and only the ego is scored. Its standardization statistics are computed over the observation window only, rather than over past and future as upstream, and its checkpoint is selected on our held-out validation days. The goal enters GLRP as two extra channels on the ego node's input features and as a learned projection added to the decoder's initial hidden state; the executed plan is the deterministic latent-mean decode.

\begin{table}[!htbp]
\centering
\small
\setlength{\tabcolsep}{5pt}
\begin{tabular}{lll}
\toprule
AEW setting & ReinFlow & HR-PPO \\
\midrule
Parallel environments & 256 & 1,024 \\
Rollout per environment & 64 four-action chunks & 128 individual steps \\
Training iterations & 1,525 & 762 \\
PPO epochs per iteration & 4 & 4 \\
Minibatches per epoch & 16 & 32 \\
Actor learning rate & $3\times10^{-6}$ & $3\times10^{-4}$ \\
Critic learning rate & $3\times10^{-4}$ & $3\times10^{-4}$ \\
Discount / GAE parameter & 0.99 / 0.95 & 0.99 / 0.95 \\
PPO clipping parameter & 0.2 & 0.2 \\
Expert regularization & FM loss, weight 1.0 & Teacher penalty, weight 0.1 \\
Goal bonus / collision penalty & 200 / 20 & 200 / 20 \\
Overall reward scale & 0.05 & 0.05 \\
Retained checkpoint interval & 152 iterations & 76 iterations \\
\bottomrule
\end{tabular}
\caption{Recorded AEW RL settings. The final training iteration is also saved. Rollout budgets describe simulator sampling, not independent new AIS demonstrations. The two regularization weights multiply different losses.}
\label{tab:rlsettings}
\end{table}

\section{Predictive Action Refinement details}
\label{app:controller}

The reported PAR configuration, labeled C3 in the saved parameter study (Table~\ref{tab:envelopes}), uses capsule geometry, a 16-step (160\,s) forecast horizon, zero clearance margin, and per-step monitoring of the remaining four-action queue. All candidate generation and traffic forecasting use the policy's current observation. The zero-margin feasibility predicate is a prediction under this model.

\paragraph{Feasibility test and selection.}
At step $t$, let $A^{\mathrm{nom}}_t$ be the four-action sequence proposed by the policy and $\mathcal K_t$ the candidate set: $A^{\mathrm{nom}}_t$ together with the slowdowns, turns, and goal-directed sequences constructed below. For each candidate $A\in\mathcal K_t$, PAR rolls the ego vessel $e$ forward under $A$ and forecasts each neighbour $j$ observed at step $t$ at constant velocity. Let $\hat d_{ej}(t+h;A)$ be the signed clearance of Equation~\ref{eq:clearance} between the predicted capsules of $e$ and $j$ at $h$ steps ahead, $h=1,\dots,H$ with $H=16$. The predicted clearance of $A$ and the feasible set are
\begin{equation}
 c_t(A)=\min_{(h,j)\in\widehat{\mathcal V}_t(A)}\hat d_{ej}(t+h;A),
 \qquad
 \mathcal F_t=\{A\in\mathcal K_t:c_t(A)\ge 0\},
 \label{eq:feasible}
\end{equation}
where $\widehat{\mathcal V}_t(A)$ contains the checked pairs of a horizon step $h$ and an observed neighbour $j$; near the goal, it ends at $A$'s predicted goal arrival (Goal handling below). A candidate is therefore feasible if its predicted hull overlaps no forecast neighbour over the checked horizon. PAR then executes
\begin{equation}
 A^\star_t=
 \begin{cases}
  A^{\mathrm{nom}}_t, & \text{if } A^{\mathrm{nom}}_t\in\mathcal F_t \text{ and it is not stalled near the goal},\\
  \arg\max_{A\in\mathcal F_t} P_t(A), & \text{else if } \mathcal F_t\neq\emptyset,\\
  \arg\max_{A\in\mathcal K_t} c_t(A), & \text{otherwise},
 \end{cases}
 \label{eq:parselect}
\end{equation}
where $P_t(A)$ is the predicted goal progress and the stalled condition is defined under Goal handling. The last case is a fallback, not a certified safe action.

\paragraph{Traffic forecast and ego rollout.}
Neighbour velocity is estimated by finite differences from its two observed positions. Positions are extrapolated at constant velocity; heading, length, and beam are held fixed. Candidate ego displacements and heading increments are rolled using Equation~\ref{eq:dynamics}. Beyond the fourth action, the final displacement is repeated and the final heading is held. Predicted capsule separation uses the evaluator's geometric representation. The controller observes only ten neighbours, while the evaluator can check all valid scene vessels.

\paragraph{Candidate construction and the rate envelope.}
The 15 candidates are the unchanged nominal chunk, four slowdowns with target speed factors $0.75,0.5,0.25,0$, six turns, and four goal-directed candidates. Following the path-reparameterization idea of PACS~\citep{romer2026pacs}, slowdowns begin by resampling the nominal polyline at cumulative arc lengths. PAR sets a symmetric speed-rate limit $a_{\max}=0.010$\,m/s$^2$. Starting from the observed ego speed $u_{-1}$, each desired speed $\bar u_j$ is limited as
\begin{equation}
 u_j=\max\{0,\operatorname{clip}(\bar u_j,u_{j-1}-a_{\max}\Delta t,
                                      u_{j-1}+a_{\max}\Delta t)\}.
\end{equation}
A second projection applies this limit to the displacement magnitude divided by $\Delta t$, because a chord across a polyline corner can be shorter than the arc used for resampling. A degenerate segment inherits the last available direction, or the observed heading. This projection can change the sampled path, so exact path preservation and the reachability guarantees of PACS are not claimed.

Turn targets are $\pm8^\circ,\pm16^\circ,\pm24^\circ$, with offset $\phi$ applied through the schedule $\phi\min((j+1)/8,1)$. Only the four generated actions are committed; an eight-step schedule does not mean that eight actions are executed before replanning. Replacement heading increments are clipped to $\pm3^\circ$. Goal-directed candidates also use the speed ramp and spread heading changes, approaching the goal without an instantaneous stop. The PAR setting disables the wide-turn and combination expansion pool. All candidates retain the simulator's displacement bound.

These limits apply to replacement actions. A geometrically feasible nominal chunk can still be kept when it violates the rate envelope. The legacy E0 ablation retains that nominal rule, uses instantaneous speed scaling and turns, and enables an 18-candidate expansion pool; its ordinary action bounds remain active. Neither PAR's heading limit nor its speed ramp establishes course-change limits or vessel-specific hydrodynamic feasibility under displacement control.

\paragraph{Selection and queue monitoring.}
The progress in Equation~\ref{eq:parselect} is $P_t(A)=\|g-p_t\|_2-\|g-\hat p_{t+4}(A)\|_2$, where $\hat p_{t+4}(A)$ is the predicted ego position after the four actions of $A$. Among feasible candidates, PAR prefers those on the passing side of its previous replacement when any is feasible, to avoid oscillating between sides. The nominal-deviation penalty is zero in the reported configuration. Before the next physical step, a new forecast checks the remaining queue and can trigger replanning.

\paragraph{Goal handling.}
Within 400\,m of the goal\rev{, a distance we call the \emph{goal-rule radius},} a candidate's clearance horizon ends at its own predicted first entry into the 100\,m goal radius, including that arrival sample. At least one checked forecast sample remains. In this region, a feasible nominal chunk with less than 10.4\,m predicted progress over the four committed actions \rev{(the \emph{progress threshold})} no longer automatically bypasses selection. Chunk substitutions therefore include low-progress handling as well as collision avoidance; a substitution count is not a count of distinct avoidance maneuvers.

\paragraph{Interpreting failure.}
An empty feasible candidate set does not establish that no feasible action exists. Passing the finite-horizon test also does not preclude a later collision, an unobserved neighbour, a forecast error, or overlap between samples. The measured outcome is the full-rollout sampled collision rate under the specified population and dynamics model. Exact tangency can pass $c_t(A)\ge0$ in Equation~\ref{eq:feasible} but counts as a collision under $d_{ej}\le0$.

\input{sections/11_appendix_results}

\FloatBarrier
\section{Replay throughput and computational scope}
\label{app:compute}

\paragraph{Workload and counting.}
The replay microbenchmark uses the AEW population of 22,258 training and 7,600 test scenarios. Every JAX repeat consumes all 29,858 scenarios in the same longest-first order. Each scenario is padded to 128 vessel slots, a configurable setting of this benchmark, with a 10\,s timestep. The single-control mode supplies recorded actions to the ego and replays its neighbours; all-vessel control supplies recorded actions to every valid vessel. Neither mode samples a learned policy or measures reactive multi-agent behavior.

Let $n_{bt}$ be the number of valid controlled vessels in live world $b$ at simulated step $t$. We report
\begin{equation}
 \mathrm{CASPS}=\frac{\sum_{b,t}n_{bt}}{\Delta T},
 \label{eq:casps}
\end{equation}
where $\Delta T$ is measured elapsed stepping time. The single-control numerator is 2,874,927 live ego transitions per full pass; the all-vessel numerator is 253,423,912 valid vessel transitions, including stationary vessels. Invalid slots and worlds that have finished their final episode do not contribute. Multiplying allocated worlds by scan length would overcount this final drain; at 4,096 worlds the saved padding fraction is 3.72\%.

\input{tables/replay_throughput}

\paragraph{GPU runner and timing.}
A ragged device-resident pool stores logged trajectories without padding every episode to the longest duration. The runner maintains a rolling 12-state window and replaces a finished scenario at the end of its logged window. It sweeps 1, 4, 16, 64, 256, 1,024, and 4,096 concurrent worlds on an RTX 3090 (24\,GB) and H100 NVL (approximately 94\,GB). Data loading, assembly, upload, and initial compilation/warm-up occur before the three timed repeats. In-loop history updates and gathering and reseeding completed worlds are inside the timed program. Both GPUs complete the largest tested batch; the 3090's all-vessel rate peaks earlier.

The program calls the logged-action actor, dynamics, observation function, and ego goal-distance check. It computes neither the learned policy nor rewards, the PAR controller, or the full collision and tracking metrics. The observation history configuration also differs from the behavioral experiments. Only the last-time $x$-coordinate slices of the ego and neighbour observation blocks feed an accumulated checksum. Unconsumed observation features may therefore be pruned by compilation; these timings do not establish the cost of materializing every policy input. The timer waits on the returned live-step counter rather than explicitly on the full returned output tuple. \rev{The effect of these choices on a full observation-consuming workload is measured in Appendix~\ref{app:thr}.} We consequently report a stepping microbenchmark rather than end-to-end policy throughput.

\paragraph{CPU reference and comparison limits.}
The public \sns{} implementation runs on an AMD EPYC 7702P, either in one process with round-robin environments or in up to 32 processes over 32 allocated logical CPUs. Each worker steps a shuffled shard for a 120-second budget. Numerators sum actual ego steps; the multiprocess denominator is the largest accumulated worker step-loop time. Loading and resets outside those loops are excluded. Each repeat therefore covers a time-budgeted prefix of the population, unlike the GPU's complete pass. A reported 91 scenarios (0.30\%) fail the CPU action-bound check; JAX instead projects actions into its bounds.

The CPU step also computes built-in kinematic, tracking, and proximity metrics. It terminates on reaching a 200\,m goal radius, whereas the JAX runner records a 100\,m goal flag but continues through the logged window. Thus the methods differ in metric work, termination, action handling, and population coverage. 

\paragraph{Relation to behavioral experiments.}
The timing runner leaves deterministic XLA restrictions disabled. The behavioral evaluation campaign uses deterministic options, including disabled GPU autotuning, across RTX 3090, RTX A5000, L40/L40S, and H100 NVL devices. Exact numerical agreement across these devices is not assumed. The candidate-filtering arms share one batch plan because stochastic draws can change with batching. \rev{Appendix~\ref{app:thr} reports a more closely matched replay workload with full observations, metrics, and the CPU termination rule, and the FM-BC policy-evaluation throughput. PAR throughput and a total training GPU-hour count remain unmeasured.}

%% file: figures/pipeline.tex
\begin{tikzpicture}[
  box/.style={draw=black!65, rounded corners=2pt, align=center,
              text width=2.45cm, minimum height=0.7cm, font=\footnotesize,
              inner sep=3pt},
  row/.style={minimum height=0.9cm},
  line/.style={-{Latex[length=2mm]}, thick, draw=black!70},
  node distance=0.5cm and 0.9cm]
\node[box,fill=blue!6] (logs) {AIS trajectories\\and metadata};
\node[box,fill=blue!6,right=of logs] (select) {Episode selection\\and regional splits};
\node[box,fill=green!6,right=of select] (data) {Demonstrations\\for offline learning};
\node[box,fill=green!6,right=of data] (policy) {Policy training\\(offline or online)};
\draw[line] (logs) -- (select);
\draw[line] (select) -- (data);
\draw[line] (data) -- (policy);
\node[box,row,fill=orange!7,below=0.55cm of data,text width=2.6cm] (sim) {Batched JAX\\simulator\\(controlled vessels)};
\node[box,row,fill=orange!7,left=0.7cm of sim,text width=2.1cm] (test) {Scenarios and\\traffic replay};
\node[box,row,fill=orange!7,below=0.55cm of policy] (actor) {Policy + optional\\test-time layer};
\node[draw=black!65,rounded corners=2pt,align=center,font=\footnotesize,inner sep=3pt,fill=purple!5,anchor=west] (metrics) at (logs.west |- sim) {Punctual goal reaching\\Tracking error ($\gcade$)\\Capsule collision};
\draw[line] (select.south) -- (select.south |- test.north);
\draw[line] (test.east) -- (test.east -| sim.west);
\draw[line] (policy) -- (actor);
\draw[line,dashed] ([xshift=0.55cm]sim.north) -- ++(0,0.3)
  node[left,font=\scriptsize,align=right,inner sep=2pt]{online RL\\rollouts}
  -| ([xshift=-0.55cm]policy.south);
\draw[line] (actor.west) -- node[above,font=\scriptsize]{actions} (actor.west -| sim.east);
\draw[line] (sim.south east) -- ++(0.2,-0.3) -| node[pos=0.32,below,font=\scriptsize]{local observations} (actor.south);
\draw[line] (sim.south west) -- ++(-0.2,-0.3) -| (metrics.south);
\end{tikzpicture}

%% file: figures/actor_critic_arch.tex
\renewcommand\footnotesize{\fontsize{8}{9.5}\selectfont}
\newcommand{\ship}[4]{%
  \begin{scope}[shift={(#1,#2)}, rotate=#4, scale=#3]
    \draw[black!70, fill=blue!18, line join=round, line width=0.5pt]
      (-0.5,0.13) -- (0.25,0.13) .. controls (0.42,0.13) and (0.53,0.06) .. (0.62,0)
      .. controls (0.53,-0.06) and (0.42,-0.13) .. (0.25,-0.13) -- (-0.5,-0.13) -- cycle;
    \draw[black!70, fill=white, line width=0.4pt] (-0.42,-0.075) rectangle (-0.24,0.075);
  \end{scope}}
\newcommand{\rbracket}[3]{%
  \draw[black!60] ($(#1.north east)+(0.08,0)$) -- ++(0.08,0) coordinate (bt)
    -- ($(#2.south east)+(0.16,0)$) coordinate (bb) -- ++(-0.08,0);
  \node[font=\scriptsize, anchor=west, inner sep=1pt] at ($(bt)!0.5!(bb)$) {#3};}
\newcommand{\lbracket}[3]{%
  \draw[black!60] ($(#1.north west)+(-0.08,0)$) -- ++(-0.08,0) coordinate (bt)
    -- ($(#2.south west)+(-0.16,0)$) coordinate (bb) -- ++(0.08,0);
  \node[font=\scriptsize, anchor=east, inner sep=1pt] at ($(bt)!0.5!(bb)$) {#3};}
\begin{tikzpicture}[
  box/.style={draw=black!65, rounded corners=2pt, align=center,
              font=\footnotesize, inner sep=3pt},
  tok/.style={box, fill=green!6, text width=1.45cm, minimum height=0.55cm},
  act/.style={box, fill=orange!7, text width=2.0cm, minimum height=0.8cm},
  cri/.style={box, fill=teal!8, text width=2.2cm, minimum height=0.8cm},
  sbox/.style={draw=black!65, rounded corners=1.5pt, font=\scriptsize,
               minimum height=0.34cm, inner sep=1pt, align=center},
  line/.style={-{Latex[length=2mm]}, thick, draw=black!70},
  thin/.style={-{Latex[length=1.5mm]}, draw=black!60},
  zoom/.style={dashed, draw=black!55},
  lab/.style={font=\scriptsize, inner sep=1.5pt},
  hdr/.style={font=\scriptsize, text=black!70, inner sep=1pt},
]
\def\xA{3.9}    %
\def\xB{5.95}   %
\def\xQ{8.0}    %

\node[font=\footnotesize] at (\xA,1.25) {Ego obs};
\node[font=\footnotesize] at (\xB,1.25) {Neighbour obs};
\node[font=\footnotesize] at (\xQ+0.1,1.25) {Goal};
\ship{\xA}{0.5}{0.95}{20}
\ship{\xB-0.42}{0.75}{0.55}{-15}
\ship{\xB+0.45}{0.78}{0.45}{170}
\ship{\xB-0.15}{0.36}{0.7}{5}
\ship{\xB+0.55}{0.38}{0.42}{60}
\begin{scope}[shift={(\xQ,0.18)}]
  \draw[black!70, fill=blue!10] (0,0) ellipse (0.2 and 0.05);
  \draw[black!70, line width=0.7pt] (0,0) -- (0,0.68);
  \draw[black!70, fill=blue!18, line join=round] (0,0.68) -- (0.38,0.56) -- (0,0.44) -- cycle;
\end{scope}

\node[tok] (tE) at (\xA,-0.42) {Token MLP};
\node[tok] (tN) at (\xB,-0.42) {Token MLP};
\node[tok] (tG) at (\xQ,-0.42) {Token MLP};
\draw[thin] (\xA,0.22) -- (tE.north);
\draw[thin] (\xB,0.16) -- (tN.north);
\draw[thin] (\xQ,0.12) -- (tG.north);

\node[box, fill=green!6, minimum width=6.3cm, text width=6.0cm] (enc) at (\xB,-1.62)
  {Late-fusion encoder\\
   \scriptsize per-group latent attention, then concatenation\\};
\foreach \t in {tE,tN,tG} \draw[thin] (\t.south) -- (\t.south |- enc.north);

\node[act] (lin) at (\xA,-3.0) {Linear\\ \scriptsize $e\to h\in\mathbb R^{128}$};
\node[act, minimum height=0.95cm] (v) at (\xA,-4.25) {Velocity net\\ $v_\theta(h,z,u)$};
\node[box, fill=purple!5, text width=2.0cm] (dp) at (\xA,-5.4) {Denormalize,\\ project to $\mathcal A$};
\draw[line] (\xA,0 |- enc.south) -- node[lab, right]{$e\in\mathbb R^{384}$} (lin.north);
\draw[line] ([xshift=-0.45cm]lin.south) -- ([xshift=-0.45cm]lin.south |- v.north);
\draw[thin] ([xshift=0.8cm]v.north) -- ++(0,0.22) -| ([xshift=0.3cm]v.north);
\node[lab, anchor=west] at ($(v.north east)+(0.02,0.2)$) {Euler steps};
\node[lab, anchor=east, align=center] (z0) at ($(v.east)+(1.75,-0.1)$) {$z_0\sim{}$\\ $\mathcal N(0,I)$};
\draw[line] (z0.west) -- (z0.west -| v.east);
\draw[line] (v.south) -- (dp.north);
\foreach \dx/\l in {-0.7/{\Delta x}, 0/{\Delta y}, 0.7/{\Delta\psi}}
  \draw[line] (dp.south) -- ++(\dx,-0.45) node[below, inner sep=1pt, font=\small] {$\l$};
\node[font=\footnotesize, anchor=north] at (\xA,-6.6) {Actor};

\node[cri] (cat) at (\xQ,-3.0) {Concat with $A$\\ \scriptsize $[e;\,D_r^{-1}\operatorname{vec}(A)]$};
\node[cri, minimum height=0.95cm] (qh) at (\xQ,-4.25) {Twin critic\\ heads};
\node[cri] (qmin) at (\xQ,-5.4) {Min over heads\\ \scriptsize \mbox{$Q=\min(Q_1,Q_2)$}};
\draw[line] (\xQ,0 |- enc.south) -- node[lab, left]{$e\in\mathbb R^{384}$} (cat.north);
\node[lab, anchor=west] (Ain) at ($(lin.east)+(0.25,0)$) {chunk $A$};
\draw[line] (Ain.east) -- (Ain.east -| cat.west);
\draw[line] (cat.south) -- (qh.north);
\draw[line] (qh.south) -- (qmin.north);
\draw[line] (qmin.south) -- ++(0,-0.45) node[below, inner sep=1pt, font=\small] {$Q_\phi(o,A)$};
\node[font=\footnotesize, align=center, anchor=north] at (\xQ,-6.6)
  {Critic};

\def\xZ{11.1}
\def\wZ{2.8cm}
\node[hdr] (eh) at (\xZ,1.2) {per group (ego, neighbours, goal)};
\node[sbox, fill=green!4, minimum width=\wZ, below=0.1cm of eh] (e1) {Cross-attention (8 latents)};
\node[sbox, fill=green!4, minimum width=\wZ, below=0.08cm of e1] (e2) {Feed-forward};
\node[sbox, fill=green!4, minimum width=\wZ, below=0.08cm of e2] (e3) {Latent self-attention};
\node[sbox, fill=green!4, minimum width=\wZ, below=0.08cm of e3] (e4) {Feed-forward};
\node[sbox, fill=green!4, minimum width=\wZ, below=0.08cm of e4] (e5) {Mean over latents};
\node[sbox, fill=green!4, minimum width=\wZ, below=0.2cm of e5] (e6) {Concat (3 groups) $\to e$};
\rbracket{e1}{e4}{$\times2$}
\draw[thin] (e5.south) -- (e6.north);
\draw[zoom] (enc.north east) -- (e1.north west);
\draw[zoom] (enc.south east) -- (e6.south west);

\node[hdr] (ch) at (\xZ,-2.6) {each head (input: the concat)};
\node[sbox, fill=teal!5, minimum width=\wZ, below=0.1cm of ch] (c1) {Dense 512};
\node[sbox, fill=teal!5, minimum width=\wZ, below=0.08cm of c1] (c2) {GELU};
\node[sbox, fill=teal!5, minimum width=\wZ, below=0.08cm of c2] (c3) {LayerNorm};
\node[sbox, fill=teal!5, minimum width=\wZ, below=0.2cm of c3] (c4) {Dense 101 (HL-Gauss logits)};
\node[sbox, fill=teal!5, minimum width=\wZ, below=0.08cm of c4] (c5) {Softmax, expectation $\to Q_i$};
\rbracket{c1}{c3}{$\times4$}
\draw[thin] (c3.south) -- (c4.north);
\draw[zoom] (qh.north east) -- (c1.north west);
\draw[zoom] (qh.south east) -- (c5.south west);

\def\xL{1.45}
\node[hdr] (ah) at (\xL,-2.9) {input $[h;\,z;\,u]$};
\node[sbox, fill=orange!4, minimum width=1.8cm, below=0.1cm of ah] (a1) {Dense 1024};
\node[sbox, fill=orange!4, minimum width=1.8cm, below=0.08cm of a1] (a2) {LayerNorm};
\node[sbox, fill=orange!4, minimum width=1.8cm, below=0.08cm of a2] (a3) {GELU};
\node[sbox, fill=orange!4, minimum width=1.8cm, below=0.2cm of a3] (a4) {Dense 1024};
\node[sbox, fill=orange!4, minimum width=1.8cm, below=0.08cm of a4] (a5) {GELU};
\node[sbox, fill=orange!4, minimum width=1.8cm, below=0.08cm of a5] (a6) {Dense 12};
\lbracket{a1}{a3}{$\times3$}
\draw[thin] (a3.south) -- (a4.north);
\draw[zoom] (v.north west) -- (a1.north east);
\draw[zoom] (v.south west) -- (a6.south east);
\end{tikzpicture}

%% file: sections/11_appendix_results.tex
\section{Four-region behavior-cloning and controller results}
\label{app:w40}
The four regions contain 7,600, 7,068, 4,219, and 2,454 test episodes, respectively. Tables~\ref{tab:fmseeds} and~\ref{tab:geoseeds} give individual-seed FM-BC and PAR results. The FM-BC seed table uses the checkpoint-sweep evaluation of Tables~\ref{tab:aew_baselines} and~\ref{tab:regional_baselines}.
\input{tables/fm_seeds}
\input{tables/geo_seeds}

\section{Four-region learning comparisons}
\label{app:learningresults}
Tables~\ref{tab:headline} and~\ref{tab:learning} report every learned baseline in all four regions, including HR-PPO and PPO beyond AEW. Their FM-BC rows use the checkpoint-sweep evaluation, as in the main anchorage tables. Table~\ref{tab:learning} also reports DEAS-BoN in every region. Table~\ref{tab:selectedsteps} specifies selected checkpoints. All methods have three policy seeds per region; critic selection shares one critic across those seeds.

PPO's instability is visible within regions as well as between them. On AEW its three seeds have punctuality approximately 0.883, 0.120, and 0.012. On AWW, high punctuality coexists with a $\gcade$ of 293.18. Including these runs avoids concealing the behavioral cost and seed variation of the unregularized objective. ReinFlow and HR-PPO also differ from the offline methods in interaction budgets, scenario eligibility, and execution frequency, as described in Appendix~\ref{app:methods}.

\paragraph{Trajectory-prediction baselines in all regions.}
Table~\ref{tab:tpregions} reports both one-shot forecasters in every anchorage region. GLRP reaches the goal punctually in 47.1--59.6\% of episodes and TrAISformer in 14.0--22.6\%, with collision rates of 24.4--51.3\% and 42.1--76.4\%, respectively, below the punctuality and above the collision rate of FM-BC in every region.
\input{tables/headline}
\input{tables/other_learning}
\input{tables/selected_steps}
\input{tables/tp_regions}

\FloatBarrier
\section{Controller envelope and candidate-filtering studies}
\label{app:envelope}
\paragraph{PAR versus the earlier candidate family.}
Table~\ref{tab:e0c3} compares the reported PAR controller with the earlier E0 family on all four regions. PAR increases residual collision rates relative to E0 but lowers the mean per-episode maximum of the legacy curvature statistic. PAR substitutes 3.8--6.8 chunks per episode, compared with E0's 0.6--1.3. These substitutions include the low-progress rule near the goal. In the seed-0 AEW diagnostic, approximately 84\% of PAR substitutions arise from that rule, so the increase cannot be interpreted as an increase in distinct avoidance maneuvers.
\input{tables/e0_c3}

\paragraph{Sensitivity to rate limits.}
The seed-0 study in Table~\ref{tab:envelopes} varies heading limits, speed ramps, turn-spread schedules, and forecast horizon. Stricter settings generally leave more residual collisions, although the ordering is not monotonic and several factors change together. For example, AEW collision rates are 0.0072 at C2, 0.0114 at C3, and 0.0246 at C5. Configuration C3, used for PAR, was chosen from this test-set study before the remaining seeds and regions were evaluated. This selection is disclosed rather than treating the subsequent table as evaluation of an independently fixed hyperparameter.

\paragraph{\rev{Selection history of PAR's settings.}}
\rev{PAR's settings were fixed using test-set runs. We adopted the 160\,s horizon and the goal rule after evaluating FM-BC on the AEW test split. We chose C3 from the 11 settings of Table~\ref{tab:envelopes}, evaluated on the AEW and AESPA test populations with FM-BC seed~0 only. We picked it because it is stricter than the lowest-collision settings, which we judged too permissive for large ships. Every one of the 11 settings removes at least 86.8\% of FM-BC's collisions, so the conclusion does not depend on this choice. No PAR setting was compared on AEPBC or AWW. Appendix~\ref{app:sel} re-selects the setting among these 11 on October data alone, and reports both that setting and C3 on November--December at October-selected FM-BC checkpoints. The October rule selects E0, and C3 ranks last of the nine settings that pass the $\reachg$ condition. Appendix~\ref{app:cost} re-selects PAR's goal-rule radius on October data alone.}

The stored executed-action counters for the AEW/AESPA seed-0 PAR study report no violations beyond numerical tolerance on replacement steps. Whole-rollout counters also include unmodified nominal actions and can exceed the PAR limits. The envelope settings are configurable rate constraints; assigning them merchant-vessel class names would require vessel-specific validation that is absent here. All envelope results used here come from the corrected speed-ramp implementation.
\input{tables/envelopes}

\paragraph{Filtering policy samples.}
A complementary seed-0 study first removes policy samples that fail the capsule forecast, then ranks feasible samples by the standard-reward critic; if none passes, it uses the highest-value sample. Tables~\ref{tab:filtered} and~\ref{tab:filtered_other} compare 16, 64, and 256 samples on the same batch-32 evaluation plan, alongside FM-BC (seed 0) in each region. The samples are drawn from the frozen policy and do not receive PAR's replacement-action limits.

At matched sample counts, filtering reduces collision rates by 10.6--27.2\%, while punctuality changes by at most 0.12 percentage points. The relative reduction grows from 10.6--13.6\% at 16 samples to 23.4--27.2\% at 256 samples. Using 256 instead of 16 unfiltered samples increases both punctuality and $\gcade$ in each region. Filtering slightly lowers that tracking error at every matched count. These results separate the effect of geometric screening from the effects of drawing more policy proposals.

Two controls examine selection and fallback choices. At 64 samples, choosing the greatest predicted minimum clearance when no sample passes gives lower collision rates than the default value-based fallback in every region. A second control samples uniformly from feasible proposals without a critic, or uniformly from all 64 when none passes. It reduces collision rates by 16.3--21.5\% relative to FM-BC (seed 0), with punctuality changes between $-0.16$ and $+0.34$ percentage points. Geometric screening thus yields a collision reduction even without value ranking in this study. Substantial residual collision rates remain; these findings concern the evaluated candidate pools and forecasts rather than the capacity of all policies or sampling schemes.
\input{tables/filtered}
\input{tables/filtered_other_regions}

\FloatBarrier
\section{Arrival time and conditional final displacement}
\label{app:extra}
\paragraph{Aligning the arrival event.}
Let $\tau_g^{\mathrm{expert}}$ be the human replay's first entry into the same 100\,m goal region used for the policy, on the same episode. We report
\begin{equation}
 \Delta_{\mathrm{arrival}}=100\frac{\tau_g-\tau_g^{\mathrm{expert}}}
                                    {\tau_g^{\mathrm{expert}}}.
\end{equation}
Negative values indicate earlier policy arrival and zero denotes matching human arrival. Comparing $\tau_g$ with the reference length $T$ instead measures a different event: the human can enter the goal before its final logged sample. The definition-matched quantity removes that systematic offset.

\begin{figure}[!htbp]
\centering
\includegraphics[width=\linewidth]{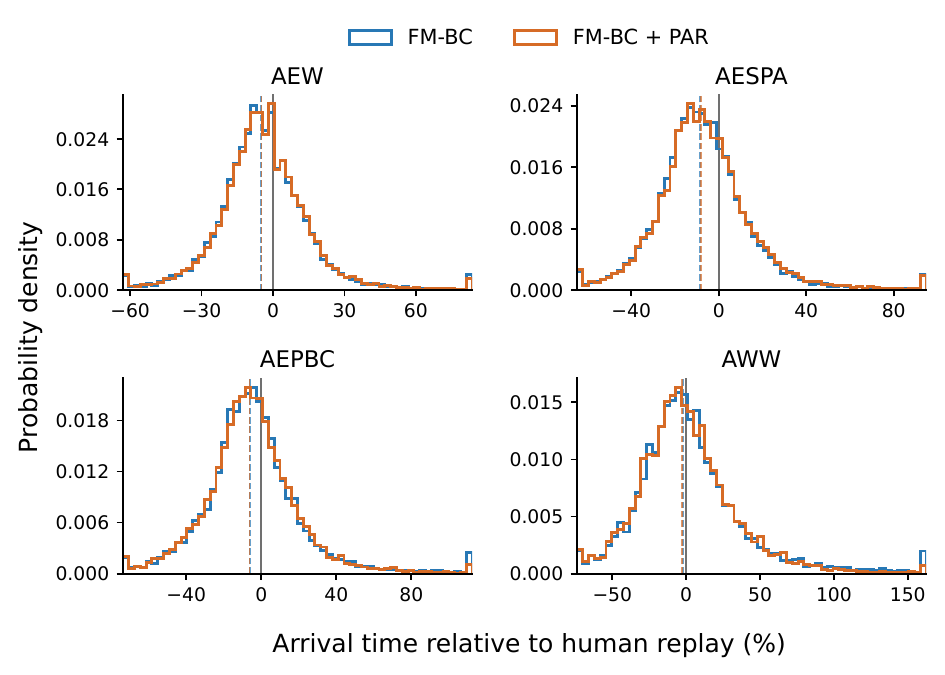}
\caption{Arrival-time distributions drawn from saved timing records, pooled across three policy seeds. Each arm includes its own episodes reaching the goal within $T+E$; negative values mean earlier arrival than human replay. The grey line marks the human reference and dashed lines mark policy medians. Display tails are clipped at pooled 0.5th/99.5th-percentile bounds; Table~\ref{tab:arrivaltime} summarizes the untrimmed values.}
\label{fig:timediff}
\end{figure}

Table~\ref{tab:arrivaltime} averages each seed's summaries, while Figure~\ref{fig:timediff} displays the saved distributions pooled across seeds. Each method includes its own episodes that reach the goal within the extended window $T+E$ (Appendix~\ref{app:termination}). Non-reaching episodes have unobserved arrival times and are excluded, rather than assigned their stopping time as an arrival. These are human-matched comparisons, not policy-versus-controller comparisons restricted to their common reaching subset. The reported median changes of at most 0.7 percentage points therefore describe the reaching populations and are not an individual-level causal timing effect.
\input{tables/arrival_time}

\paragraph{What the conditional FDE means.}
The main-table FDE$_{\neg R}$ is computed from the stored reference-window FDE on episodes with $\reachg=0$. It differs from conditioning on failure within $T+E$, and from measuring goal distance at the final executed sample. From FM-BC to FM-BC + PAR (Tables~\ref{tab:aew_baselines} and~\ref{tab:regional_baselines}) it rises from 436 to 470\,m on AEW, 533 to 571\,m on AESPA, 503 to 565\,m on AEPBC, and 417 to 486\,m on AWW. Because the methods change which episodes miss the deadline, this increase alone does not establish deterioration on the same failures.

For DEAS-BoN with 16 candidates, conditional FDE is 375, 465, 450, and 380\,m on AEW, AESPA, AEPBC, and AWW, respectively (Tables~\ref{tab:aew_baselines} and~\ref{tab:learning}). These values are below those of FM-BC (436, 533, 503, and 417\,m; Table~\ref{tab:headline}), but likewise average each method's own missed-deadline episodes. The filtering tables use the same conditional definition. Comparing the error on a shared set of failures would answer a different question and is not available from these aggregate summaries.

\FloatBarrier
\section{Kinematic metrics and their interpretation}
\label{app:kinematics}
The saved kinematic statistics follow the \sns{} implementation~\citep{pham2025shipnavisim}. For displacement $d_t=p_t-p_{t-1}$, speed is $v_t=\|d_t\|/\Delta t$, acceleration is $(v_t-v_{t-1})/\Delta t$, and drift is the wrapped angle between $\operatorname{atan2}(d_{t,y},d_{t,x})$ and the recorded heading $\psi_t$. The first speed difference uses the observed initial speed. The implementation computes these extrema over the executed portion of the reference window; the extended collision-evaluation window uses a different mask.

The legacy parametric expression uses a per-step heading increment where a heading rate would normally occur. Above its $v_t^2\ge10^{-3}$ threshold it simplifies to
\begin{equation}
 \tilde\kappa_t=\frac{\operatorname{wrap}(\psi_t-\psi_{t-1})}{v_t},
 \qquad
 \kappa_{\psi,t}=\frac{\tilde\kappa_t}{\Delta t}
               =\frac{\operatorname{wrap}(\psi_t-\psi_{t-1})}{\|d_t\|},
\end{equation}
where $\operatorname{wrap}(\theta)=((\theta+\pi)\bmod 2\pi)-\pi$ maps an angle difference to $[-\pi,\pi)$. Both are zeroed below that speed threshold. Tables retain $\tilde\kappa$ for continuity with the saved results and do not label it rad/m. Dividing by 10\,s gives heading change per travelled metre, which is still distinct from curvature derived from successive courses over ground when heading and displacement direction differ.

Tables~\ref{tab:kinematics} and~\ref{tab:learningkinematics} average per-episode extrema, then average seeds. PAR raises the mean maximum $\tilde\kappa$ relative to FM-BC but remains below human replay in all four regions. Acceleration extrema remain close to FM-BC's because most actions are unchanged; this is compatible with limiting only replacements. AIS jitter, heading discontinuities, and low speeds affect the extrema, so proximity to human extrema alone does not validate maneuverability or establish a cause of collisions.
\input{tables/kinematics}
\input{tables/learning_kinematics}

Table~\ref{tab:selectionkinematics} adds the kinematics of DEAS-BoN with 16 candidates and filtered DEAS-BoN with 256 candidates, alongside FM-BC. Both selection schemes increase the magnitudes of mean acceleration and deceleration extrema in every region. Their higher punctuality therefore accompanies changes in executed motion, even though they select proposals from the frozen policy. These summaries describe the resulting rollouts; they do not establish vessel-specific maneuverability.
\input{tables/selection_kinematics}

\FloatBarrier
\section{Descriptive collision-case analysis}
\label{app:cases}
An AEW seed-0 scan attributes collisions to partner class and motion on all 7,600 test episodes, using PAR rollouts that reproduce the seed-0 PAR evaluation exactly and a separate FM-BC rollout of the same checkpoint. The FM-BC scan contains 2,069 episode--partner collision pairs, compared with 94 under PAR. Of the FM-BC pairs, 1,837 (88.8\%) involve partners classified as anchored from their recorded speed. Under PAR, 38 residual pairs involve anchored partners and 56 involve moving partners (Table~\ref{tab:caseclasses}). Exposure is dominated by anchored vessels, so pair shares should be read alongside exposure counts.

The reduction in pair counts is larger for anchored partners than for moving partners. This is consistent with the relevance of stationary obstacles in the anchorage task, but it does not isolate forecast error, candidate limitations, observation truncation, or collision responsibility. A speed threshold also does not imply perfectly constant position or heading throughout a forecast horizon. The main rates are those of Tables~\ref{tab:aew_baselines} and~\ref{tab:regional_baselines}.
\input{tables/contact_cases}

%% file: tables/fm_seeds.tex
\begin{table}[!htbp]
\centering\small
\setlength{\tabcolsep}{4pt}
\sbox{\resulttable}{%
\begin{tabular}{lrrccc}
\toprule
Region & Seed & Step & $\reachg$ & $C$ & $\gcade$ \\
\midrule
AEW & 0 & $1920000$ & $0.8880$ & $0.2007$ & $23.19$ \\
AEW & 1 & $1930000$ & $0.8962$ & $0.2076$ & $23.25$ \\
AEW & 2 & $2000000$ & $0.8991$ & $0.2025$ & $23.09$ \\
\midrule
AESPA & 0 & $1940000$ & $0.9317$ & $0.4502$ & $26.83$ \\
AESPA & 1 & $1920000$ & $0.9310$ & $0.4529$ & $26.77$ \\
AESPA & 2 & $2000000$ & $0.9314$ & $0.4501$ & $26.69$ \\
\midrule
AEPBC & 0 & $1990000$ & $0.8978$ & $0.2764$ & $24.21$ \\
AEPBC & 1 & $2000000$ & $0.8912$ & $0.2773$ & $24.05$ \\
AEPBC & 2 & $1930000$ & $0.9040$ & $0.2799$ & $24.22$ \\
\midrule
AWW & 0 & $1930000$ & $0.7632$ & $0.2514$ & $27.16$ \\
AWW & 1 & $1970000$ & $0.7861$ & $0.2608$ & $27.30$ \\
AWW & 2 & $1910000$ & $0.7828$ & $0.2494$ & $27.35$ \\
\bottomrule
\end{tabular}}
\ifdim\wd\resulttable>\linewidth\resizebox{\linewidth}{!}{\usebox{\resulttable}}\else\usebox{\resulttable}\fi
\caption{FM-BC results at the selected checkpoint for each seed. These are the FM-BC evaluations of Tables~\ref{tab:aew_baselines} and~\ref{tab:regional_baselines}.}
\label{tab:fmseeds}
\end{table}

%% file: tables/geo_seeds.tex
\begin{table}[!htbp]
\centering\small
\setlength{\tabcolsep}{4pt}
\sbox{\resulttable}{%
\begin{tabular}{lrccr}
\toprule
Region & Seed & $\reachg$ & $C$ & Subst./ep \\
\midrule
AEW & 0 & $0.8925$ & $0.0114$ & $4.32$ \\
AEW & 1 & $0.8962$ & $0.0117$ & $3.57$ \\
AEW & 2 & $0.8978$ & $0.0104$ & $3.57$ \\
\midrule
AESPA & 0 & $0.9250$ & $0.0352$ & $4.93$ \\
AESPA & 1 & $0.9220$ & $0.0349$ & $4.50$ \\
AESPA & 2 & $0.9305$ & $0.0307$ & $4.42$ \\
\midrule
AEPBC & 0 & $0.9068$ & $0.0194$ & $6.18$ \\
AEPBC & 1 & $0.9033$ & $0.0175$ & $6.17$ \\
AEPBC & 2 & $0.9073$ & $0.0213$ & $6.90$ \\
\midrule
AWW & 0 & $0.8073$ & $0.0257$ & $7.49$ \\
AWW & 1 & $0.8187$ & $0.0183$ & $6.81$ \\
AWW & 2 & $0.8191$ & $0.0240$ & $6.16$ \\
\bottomrule
\end{tabular}}
\ifdim\wd\resulttable>\linewidth\resizebox{\linewidth}{!}{\usebox{\resulttable}}\else\usebox{\resulttable}\fi
\caption{PAR results by seed. Substitutions count replaced action chunks, including near-goal low-progress handling, rather than distinct avoidance maneuvers.}
\label{tab:geoseeds}
\end{table}

%% file: tables/headline.tex
\begin{table}[!htbp]
\centering\small
\setlength{\tabcolsep}{4pt}
\sbox{\resulttable}{%
\begin{tabular}{llccccc}
\toprule
Region & Method & $\reachg\uparrow$ & Collision rate $C\downarrow$ & CPA (m) & $\gcade\downarrow$ & FDE$_{\neg R}$ (m) \\
\midrule
AEW & FM-BC & $0.8944\pm0.0057$ & $0.2036\pm0.0036$ & $87.8$ & $\boldsymbol{23.18\pm0.08}$ & $436$ \\
AEW & HR-PPO & $0.8796\pm0.0094$ & $0.2200\pm0.0080$ & $80.0$ & $30.25\pm0.86$ & $453$ \\
AEW & QC-FQL & $\boldsymbol{0.9199\pm0.0013}$ & $\boldsymbol{0.2025\pm0.0019}$ & $87.3$ & $23.57\pm0.15$ & $455$ \\
\midrule
AESPA & FM-BC & $0.9313\pm0.0004$ & $0.4510\pm0.0016$ & $8.5$ & $\boldsymbol{26.76\pm0.07}$ & $533$ \\
AESPA & HR-PPO & $0.9466\pm0.0015$ & $0.4576\pm0.0041$ & $7.8$ & $32.78\pm0.12$ & $752$ \\
AESPA & QC-FQL & $\boldsymbol{0.9616\pm0.0014}$ & $\boldsymbol{0.4461\pm0.0020}$ & $9.4$ & $27.85\pm0.09$ & $684$ \\
\midrule
AEPBC & FM-BC & $0.8977\pm0.0064$ & $0.2779\pm0.0018$ & $56.0$ & $\boldsymbol{24.16\pm0.10}$ & $503$ \\
AEPBC & HR-PPO & $0.9049\pm0.0049$ & $0.3070\pm0.0097$ & $46.4$ & $30.56\pm0.50$ & $590$ \\
AEPBC & QC-FQL & $\boldsymbol{0.9428\pm0.0021}$ & $\boldsymbol{0.2746\pm0.0022}$ & $57.9$ & $24.62\pm0.05$ & $644$ \\
\midrule
AWW & FM-BC & $0.7774\pm0.0123$ & $0.2539\pm0.0061$ & $79.9$ & $27.27\pm0.10$ & $417$ \\
AWW & HR-PPO & $0.8370\pm0.0371$ & $0.2680\pm0.0120$ & $73.1$ & $32.04\pm0.51$ & $381$ \\
AWW & QC-FQL & $\boldsymbol{0.8583\pm0.0031}$ & $\boldsymbol{0.2399\pm0.0033}$ & $84.3$ & $\boldsymbol{26.39\pm0.18}$ & $404$ \\
\bottomrule
\end{tabular}}
\ifdim\wd\resulttable>\linewidth\resizebox{\linewidth}{!}{\usebox{\resulttable}}\else\usebox{\resulttable}\fi
\caption{Learning baselines on the four-region test population. Rates and $\gcade$ are mean $\pm$ sample standard deviation across three seeds; CPA and FDE$_{\neg R}$ are means of per-seed summaries. FDE$_{\neg R}$ conditions on missing the punctual deadline. \rev{Checkpoints are test-selected (Appendix~\ref{app:sel}). Bold: best value per region for $\reachg$, $C$, and $\gcade$.} HR-PPO uses KL weight 0.1. Expert replay has $\reachg=1$ by construction; its regional CPA is 119.8, 76.0, 106.0, and 119.0\,m. Further baselines appear in Appendix~\ref{app:learningresults}.}
\label{tab:headline}
\end{table}

%% file: tables/other_learning.tex
\begin{table}[!htbp]
\centering\small
\setlength{\tabcolsep}{4pt}
\sbox{\resulttable}{%
\begin{tabular}{llccccc}
\toprule
Region & Method & $\reachg\uparrow$ & Collision rate $C\downarrow$ & CPA (m) & $\gcade\downarrow$ & FDE$_{\neg R}$ (m) \\
\midrule
AEW & NF-BC & $0.7545\pm0.0003$ & $0.2231\pm0.0028$ & $80.6$ & $26.54\pm0.13$ & $490$ \\
AEW & ReinFlow & $0.9246\pm0.0059$ & $0.2132\pm0.0095$ & $84.3$ & $25.48\pm0.62$ & $439$ \\
AEW & PPO & $0.3382\pm0.4746$ & $0.4329\pm0.1714$ & $21.1$ & $242.01\pm56.28$ & $3640$ \\
AEW & DEAS-BoN & $0.9665\pm0.0015$ & $0.2054\pm0.0020$ & $87.2$ & $31.77\pm0.09$ & $375$ \\
\midrule
AESPA & NF-BC & $0.7961\pm0.0036$ & $0.4897\pm0.0041$ & $1.8$ & $29.52\pm0.02$ & $501$ \\
AESPA & ReinFlow & $0.9482\pm0.0201$ & $0.4638\pm0.0228$ & $6.5$ & $30.00\pm0.34$ & $540$ \\
AESPA & PPO & $0.0075\pm0.0034$ & $0.4883\pm0.1432$ & $10.4$ & $185.71\pm13.65$ & $3358$ \\
AESPA & DEAS-BoN & $0.9768\pm0.0016$ & $0.4505\pm0.0041$ & $8.4$ & $36.60\pm0.21$ & $465$ \\
\midrule
AEPBC & NF-BC & $0.7621\pm0.0031$ & $0.3163\pm0.0106$ & $45.1$ & $26.92\pm0.23$ & $491$ \\
AEPBC & ReinFlow & $0.9256\pm0.0182$ & $0.2845\pm0.0078$ & $55.5$ & $26.30\pm0.58$ & $530$ \\
AEPBC & PPO & $0.6672\pm0.3032$ & $0.3362\pm0.0546$ & $26.8$ & $246.84\pm59.39$ & $2076$ \\
AEPBC & DEAS-BoN & $0.9534\pm0.0021$ & $0.2780\pm0.0047$ & $56.2$ & $28.90\pm0.06$ & $450$ \\
\midrule
AWW & NF-BC & $0.6989\pm0.0014$ & $0.2569\pm0.0059$ & $76.7$ & $27.56\pm0.06$ & $399$ \\
AWW & ReinFlow & $0.8219\pm0.0145$ & $0.2547\pm0.0133$ & $80.6$ & $28.40\pm0.03$ & $416$ \\
AWW & PPO & $0.9849\pm0.0195$ & $0.2691\pm0.0349$ & $45.0$ & $293.18\pm26.13$ & $2993$ \\
AWW & DEAS-BoN & $0.8350\pm0.0070$ & $0.2498\pm0.0011$ & $79.2$ & $28.47\pm0.21$ & $380$ \\
\bottomrule
\end{tabular}}
\ifdim\wd\resulttable>\linewidth\resizebox{\linewidth}{!}{\usebox{\resulttable}}\else\usebox{\resulttable}\fi
\caption{Additional four-region learning baselines, using the same aggregation and metric definitions as Table~\ref{tab:headline}. PPO has no teacher penalty. DEAS-BoN uses $N=16$ candidates, a batch-64 plan, and one regional critic shared across policy seeds. Each row includes all three seeds, including poorly performing runs.}
\label{tab:learning}
\end{table}

%% file: tables/selected_steps.tex
\begin{table}[!htbp]
\centering\small
\setlength{\tabcolsep}{4pt}
\sbox{\resulttable}{%
\begin{tabular}{llrrr}
\toprule
Region & Method & Seed 0 & Seed 1 & Seed 2 \\
\midrule
AEW & FM-BC & $1920000$ & $1930000$ & $2000000$ \\
AEW & NF-BC & $1940000$ & $1980000$ & $1950000$ \\
AEW & ReinFlow & $1520$ & $304$ & $1520$ \\
AEW & HR-PPO & $760$ & $762$ & $762$ \\
AEW & PPO & $762$ & $760$ & $760$ \\
AEW & QC-FQL & $100000$ & $200000$ & $300000$ \\
\midrule
AESPA & FM-BC & $1940000$ & $1920000$ & $2000000$ \\
AESPA & NF-BC & $1910000$ & $1910000$ & $1930000$ \\
AESPA & ReinFlow & $1520$ & $1525$ & $304$ \\
AESPA & HR-PPO & $762$ & $762$ & $760$ \\
AESPA & PPO & $152$ & $532$ & $380$ \\
AESPA & QC-FQL & $800000$ & $1000000$ & $300000$ \\
\midrule
AEPBC & FM-BC & $1990000$ & $2000000$ & $1930000$ \\
AEPBC & NF-BC & $1970000$ & $1980000$ & $1930000$ \\
AEPBC & ReinFlow & $304$ & $1520$ & $1520$ \\
AEPBC & HR-PPO & $760$ & $760$ & $762$ \\
AEPBC & PPO & $760$ & $760$ & $762$ \\
AEPBC & QC-FQL & $100000$ & $200000$ & $400000$ \\
\midrule
AWW & FM-BC & $1930000$ & $1970000$ & $1910000$ \\
AWW & NF-BC & $1920000$ & $1930000$ & $1930000$ \\
AWW & ReinFlow & $1525$ & $304$ & $1368$ \\
AWW & HR-PPO & $760$ & $762$ & $762$ \\
AWW & PPO & $762$ & $684$ & $762$ \\
AWW & QC-FQL & $100000$ & $100000$ & $200000$ \\
\bottomrule
\end{tabular}}
\ifdim\wd\resulttable>\linewidth\resizebox{\linewidth}{!}{\usebox{\resulttable}}\else\usebox{\resulttable}\fi
\caption{Selected checkpoints. FM-BC, NF-BC, and QC-FQL values are gradient updates; ReinFlow, HR-PPO, and PPO values are rollout/update iterations. Each checkpoint is selected using test $\reachg$, with $\gcade$ breaking ties. Controller and critic evaluations use the selected FM-BC checkpoint of the same seed.}
\label{tab:selectedsteps}
\end{table}

%% file: tables/tp_regions.tex
\begin{table}[!htbp]
\centering\small
\setlength{\tabcolsep}{4pt}
\sbox{\resulttable}{%
\begin{tabular}{llccccc}
\toprule
Region & Method & $\reachg\uparrow$ & Collision rate $C\downarrow$ & CPA (m) & $\gcade\downarrow$ & FDE$_{\neg R}$ (m) \\
\midrule
AEW & GLRP (one-shot) & $0.5315\pm0.0048$ & $0.2439\pm0.0024$ & $70.8$ & $16.84\pm0.26$ & $303$ \\
AEW & TrAISformer (one-shot) & $0.2261\pm0.0125$ & $0.5075\pm0.0069$ & $-1.5$ & $38.54\pm1.00$ & $867$ \\
\midrule
AESPA & GLRP (one-shot) & $0.5961\pm0.0647$ & $0.5134\pm0.0089$ & $-2.2$ & $22.35\pm0.61$ & $369$ \\
AESPA & TrAISformer (one-shot) & $0.1401\pm0.0070$ & $0.7644\pm0.0091$ & $-15.7$ & $58.92\pm1.20$ & $1430$ \\
\midrule
AEPBC & GLRP (one-shot) & $0.4710\pm0.0278$ & $0.3344\pm0.0017$ & $37.8$ & $20.55\pm0.21$ & $344$ \\
AEPBC & TrAISformer (one-shot) & $0.1655\pm0.0110$ & $0.5134\pm0.0002$ & $-1.9$ & $47.78\pm0.70$ & $1046$ \\
\midrule
AWW & GLRP (one-shot) & $0.5662\pm0.0306$ & $0.3010\pm0.0103$ & $59.8$ & $20.35\pm0.54$ & $287$ \\
AWW & TrAISformer (one-shot) & $0.2009\pm0.0170$ & $0.4211\pm0.0049$ & $18.4$ & $52.09\pm0.57$ & $1175$ \\
\bottomrule
\end{tabular}}
\ifdim\wd\resulttable>\linewidth\resizebox{\linewidth}{!}{\usebox{\resulttable}}\else\usebox{\resulttable}\fi
\caption{One-shot trajectory-prediction baselines in all four anchorage regions, each queried once at the first step and executed open-loop; three seeds each. Aggregation and metrics follow Table~\ref{tab:headline}.}
\label{tab:tpregions}
\end{table}

%% file: tables/e0_c3.tex
\begin{table}[!htbp]
\centering\small
\setlength{\tabcolsep}{4pt}
\sbox{\resulttable}{%
\begin{tabular}{llccc}
\toprule
Region & Method & $\reachg$ & $C$ & max $\tilde\kappa$ \\
\midrule
AEW & FM-BC & $0.8944$ & $0.2036$ & $0.0055$ \\
AEW & $+$ E0 & $0.9054$ & $0.0026$ & $0.0502$ \\
AEW & $+$ PAR & $0.8955$ & $0.0112$ & $0.0205$ \\
\midrule
AESPA & FM-BC & $0.9313$ & $0.4510$ & $0.0150$ \\
AESPA & $+$ E0 & $0.9439$ & $0.0086$ & $0.0854$ \\
AESPA & $+$ PAR & $0.9259$ & $0.0336$ & $0.0342$ \\
\midrule
AEPBC & FM-BC & $0.8977$ & $0.2779$ & $0.0135$ \\
AEPBC & $+$ E0 & $0.9223$ & $0.0061$ & $0.0913$ \\
AEPBC & $+$ PAR & $0.9058$ & $0.0194$ & $0.0316$ \\
\midrule
AWW & FM-BC & $0.7774$ & $0.2539$ & $0.0229$ \\
AWW & $+$ E0 & $0.8499$ & $0.0042$ & $0.1639$ \\
AWW & $+$ PAR & $0.8150$ & $0.0227$ & $0.0365$ \\
\bottomrule
\end{tabular}}
\ifdim\wd\resulttable>\linewidth\resizebox{\linewidth}{!}{\usebox{\resulttable}}\else\usebox{\resulttable}\fi
\caption{The earlier E0 candidate family and PAR (configuration C3), three-seed means. E0 retains the simulator's displacement and heading bounds but has no additional rate envelope. The curvature column is the legacy heading-based statistic defined in Appendix~\ref{app:kinematics}; it is not curvature in rad/m.}
\label{tab:e0c3}
\end{table}

%% file: tables/envelopes.tex
\begin{table}[!htbp]
\centering\small
\setlength{\tabcolsep}{3pt}
\sbox{\resulttable}{%
\begin{tabular}{lrrrrcccc}
\toprule
Setting & $|\Delta\psi|$ ($^\circ$) & $|a|$ (m/s$^2$) & Spread & Horizon & AEW $C$ & AEW $\reachg$ & AESPA $C$ & AESPA $\reachg$ \\
\midrule
E0 & --- & --- & 1 & 16 & $0.0028$ & $0.9014$ & $0.0083$ & $0.9427$ \\
E1 & 17 & 0.150 & 1 & 16 & $0.0047$ & $0.9016$ & $0.0144$ & $0.9403$ \\
E2 & 10 & 0.050 & 4 & 16 & $0.0034$ & $0.8979$ & $0.0170$ & $0.9288$ \\
E3 & 10 & 0.050 & 4 & 32 & $0.0041$ & $0.8959$ & $0.0151$ & $0.9290$ \\
E4 & 17 & 0.150 & 1 & 24 & $0.0054$ & $0.9029$ & $0.0117$ & $0.9393$ \\
E5 & 20 & 0.100 & 2 & 16 & $0.0037$ & $0.9001$ & $0.0112$ & $0.9393$ \\
C1 & 5 & 0.030 & 4 & 16 & $0.0082$ & $0.8976$ & $0.0235$ & $0.9336$ \\
C2 & 5 & 0.015 & 6 & 16 & $0.0072$ & $0.8989$ & $0.0235$ & $0.9329$ \\
C3 & 3 & 0.010 & 8 & 16 & $0.0114$ & $0.8925$ & $0.0352$ & $0.9250$ \\
C4 & 2 & 0.005 & 8 & 16 & $0.0229$ & $0.8891$ & $0.0565$ & $0.9138$ \\
C5 & 2 & 0.000 & 8 & 16 & $0.0246$ & $0.8875$ & $0.0603$ & $0.9085$ \\
\bottomrule
\end{tabular}}
\ifdim\wd\resulttable>\linewidth\resizebox{\linewidth}{!}{\usebox{\resulttable}}\else\usebox{\resulttable}\fi
\caption{Envelope study on FM-BC seed 0, with all settings for a region evaluated on one GPU after the speed-ramp correction. PAR uses configuration C3. Spread is the schedule length used to construct a turn; only four actions are committed before replanning. E5 retains combinations of bounded turns and slowdowns; other rate-limited settings disable expansion. Settings change multiple candidate-construction choices and are not a single-factor sweep. Their names are configuration identifiers, not validated vessel classes.}
\label{tab:envelopes}
\end{table}

%% file: tables/filtered.tex
\begin{table}[!htbp]
\centering\small
\setlength{\tabcolsep}{3pt}
\sbox{\resulttable}{%
\begin{tabular}{llccccc}
\toprule
Region & Method & $\reachg\uparrow$ & Collision rate $C\downarrow$ & CPA (m) & $\gcade\downarrow$ & FDE$_{\neg R}$ (m) \\
\midrule
AEW & FM-BC & $0.8880$ & $0.2007$ & $88.3$ & $23.19$ & $429$ \\
AEW & DEAS-BoN, $N=16$ & $0.9670$ & $0.2078$ & $86.7$ & $31.76$ & $377$ \\
AEW & \quad $+$ filter & $0.9663$ & $0.1795$ & $86.9$ & $31.50$ & $378$ \\
AEW & DEAS-BoN, $N=64$ & $0.9741$ & $0.2067$ & $84.5$ & $38.46$ & $312$ \\
AEW & \quad $+$ filter & $0.9746$ & $0.1687$ & $84.7$ & $37.98$ & $319$ \\
AEW & DEAS-BoN, $N=256$ & $0.9746$ & $0.2126$ & $83.6$ & $47.02$ & $268$ \\
AEW & \quad $+$ filter & $0.9738$ & $0.1549$ & $83.8$ & $46.15$ & $277$ \\
AEW & Filter 64, least violation & $0.9742$ & $0.1537$ & $84.7$ & $37.41$ & $320$ \\
AEW & Random feasible, 64 & $0.8914$ & $0.1576$ & $87.6$ & $23.10$ & $434$ \\
\midrule
AESPA & FM-BC & $0.9317$ & $0.4502$ & $8.7$ & $26.83$ & $552$ \\
AESPA & DEAS-BoN, $N=16$ & $0.9775$ & $0.4498$ & $8.1$ & $37.00$ & $457$ \\
AESPA & \quad $+$ filter & $0.9779$ & $0.4012$ & $10.6$ & $36.44$ & $478$ \\
AESPA & DEAS-BoN, $N=64$ & $0.9803$ & $0.4537$ & $8.1$ & $44.29$ & $369$ \\
AESPA & \quad $+$ filter & $0.9803$ & $0.3779$ & $11.3$ & $43.31$ & $381$ \\
AESPA & DEAS-BoN, $N=256$ & $0.9781$ & $0.4614$ & $6.8$ & $52.70$ & $297$ \\
AESPA & \quad $+$ filter & $0.9779$ & $0.3441$ & $11.0$ & $51.09$ & $304$ \\
AESPA & Filter 64, least violation & $0.9803$ & $0.3476$ & $11.7$ & $41.85$ & $415$ \\
AESPA & Random feasible, 64 & $0.9301$ & $0.3769$ & $12.4$ & $26.82$ & $547$ \\
\bottomrule
\end{tabular}}
\ifdim\wd\resulttable>\linewidth\resizebox{\linewidth}{!}{\usebox{\resulttable}}\else\usebox{\resulttable}\fi
\caption{Candidate filtering on AEW and AESPA, seed 0. The DEAS-BoN, filter, and control arms use batch 32; the FM-BC row is the seed-0 evaluation of Table~\ref{tab:fmseeds}. Each $+$ filter row uses the same sample count as the DEAS-BoN row above it. All methods share the regional FM-BC checkpoint; DEAS-BoN arms share the standard-reward critic. Rates and errors summarize the full regional test population, and CPA is the median episode minimum clearance. FDE$_{\neg R}$ conditions on missing the punctual deadline. Least violation changes the fallback when no sample passes; random feasible removes value ranking. These samples do not receive PAR's replacement-action limits.}
\label{tab:filtered}
\end{table}

%% file: tables/filtered_other_regions.tex
\begin{table}[!htbp]
\centering\small
\setlength{\tabcolsep}{3pt}
\sbox{\resulttable}{%
\begin{tabular}{llccccc}
\toprule
Region & Method & $\reachg\uparrow$ & Collision rate $C\downarrow$ & CPA (m) & $\gcade\downarrow$ & FDE$_{\neg R}$ (m) \\
\midrule
AEPBC & FM-BC & $0.8978$ & $0.2764$ & $56.7$ & $24.21$ & $502$ \\
AEPBC & DEAS-BoN, $N=16$ & $0.9562$ & $0.2747$ & $57.3$ & $28.94$ & $458$ \\
AEPBC & \quad $+$ filter & $0.9559$ & $0.2456$ & $57.2$ & $28.77$ & $459$ \\
AEPBC & DEAS-BoN, $N=64$ & $0.9647$ & $0.2830$ & $53.6$ & $31.98$ & $410$ \\
AEPBC & \quad $+$ filter & $0.9635$ & $0.2280$ & $53.4$ & $31.65$ & $402$ \\
AEPBC & DEAS-BoN, $N=256$ & $0.9708$ & $0.2868$ & $52.5$ & $35.22$ & $383$ \\
AEPBC & \quad $+$ filter & $0.9713$ & $0.2088$ & $52.9$ & $34.76$ & $383$ \\
AEPBC & Filter 64, least violation & $0.9602$ & $0.2074$ & $53.5$ & $31.16$ & $401$ \\
AEPBC & Random feasible, 64 & $0.8962$ & $0.2292$ & $55.0$ & $24.18$ & $503$ \\
\midrule
AWW & FM-BC & $0.7632$ & $0.2514$ & $80.8$ & $27.16$ & $401$ \\
AWW & DEAS-BoN, $N=16$ & $0.8264$ & $0.2539$ & $79.3$ & $28.23$ & $360$ \\
AWW & \quad $+$ filter & $0.8268$ & $0.2221$ & $79.4$ & $28.20$ & $361$ \\
AWW & DEAS-BoN, $N=64$ & $0.8488$ & $0.2498$ & $81.6$ & $29.08$ & $359$ \\
AWW & \quad $+$ filter & $0.8480$ & $0.2115$ & $81.8$ & $28.98$ & $361$ \\
AWW & DEAS-BoN, $N=256$ & $0.8606$ & $0.2559$ & $80.6$ & $30.49$ & $327$ \\
AWW & \quad $+$ filter & $0.8598$ & $0.1960$ & $81.1$ & $30.37$ & $330$ \\
AWW & Filter 64, least violation & $0.8476$ & $0.2025$ & $81.7$ & $28.93$ & $366$ \\
AWW & Random feasible, 64 & $0.7637$ & $0.2033$ & $78.8$ & $27.31$ & $402$ \\
\bottomrule
\end{tabular}}
\ifdim\wd\resulttable>\linewidth\resizebox{\linewidth}{!}{\usebox{\resulttable}}\else\usebox{\resulttable}\fi
\caption{Candidate filtering on AEPBC and AWW, with the same seed-0 protocol, metric definitions, and arm order as Table~\ref{tab:filtered}. Each filtered DEAS-BoN arm is compared with its corresponding unfiltered sample count.}
\label{tab:filtered_other}
\end{table}

%% file: tables/arrival_time.tex
\begin{table}[!htbp]
\centering\small
\setlength{\tabcolsep}{4pt}
\sbox{\resulttable}{%
\begin{tabular}{llrrrrrr}
\toprule
Region & Method & Reaching $n$ & Mean (\%) & Median (\%) & $p_{10}$ (\%) & $p_{90}$ (\%) & Earlier \\
\midrule
AEW & FM-BC & $7554$ & $-4.01$ & $-5.17$ & $-25.33$ & $16.68$ & $0.610$ \\
AEW & $+$ PAR & $7575$ & $-3.96$ & $-4.85$ & $-25.19$ & $17.11$ & $0.604$ \\
\midrule
AESPA & FM-BC & $7021$ & $-6.85$ & $-8.56$ & $-31.39$ & $17.20$ & $0.671$ \\
AESPA & $+$ PAR & $7037$ & $-6.27$ & $-7.96$ & $-31.17$ & $18.75$ & $0.658$ \\
\midrule
AEPBC & FM-BC & $4170$ & $-3.95$ & $-6.00$ & $-33.61$ & $23.85$ & $0.606$ \\
AEPBC & $+$ PAR & $4202$ & $-5.07$ & $-6.22$ & $-33.99$ & $22.91$ & $0.614$ \\
\midrule
AWW & FM-BC & $2337$ & $3.83$ & $-1.99$ & $-35.38$ & $45.81$ & $0.495$ \\
AWW & $+$ PAR & $2408$ & $1.09$ & $-2.68$ & $-36.44$ & $41.41$ & $0.518$ \\
\bottomrule
\end{tabular}}
\ifdim\wd\resulttable>\linewidth\resizebox{\linewidth}{!}{\usebox{\resulttable}}\else\usebox{\resulttable}\fi
\caption{Arrival time relative to human replay on the same episode: $100(\tau_g-\tau_g^{\mathrm{expert}})/\tau_g^{\mathrm{expert}}$. Each cell is the mean of three per-seed summaries; counts are rounded. Non-reaching episodes are excluded separately for each method, so the populations differ. This pairs each policy with the human reference, not with the other policy on a common subset.}
\label{tab:arrivaltime}
\end{table}

%% file: tables/kinematics.tex
\begin{table}[!htbp]
\centering\small
\setlength{\tabcolsep}{4pt}
\sbox{\resulttable}{%
\begin{tabular}{llrrrrrr}
\toprule
Region & Method & min $\tilde\kappa$ & max $\tilde\kappa$ & min $a$ & max $a$ & min drift & max drift \\
\midrule
AEW & Human & $-0.0310$ & $0.0266$ & $-0.5584$ & $0.4901$ & $-0.2304$ & $0.4621$ \\
AEW & FM-BC & $-0.0076$ & $0.0055$ & $-0.1705$ & $0.1526$ & $-0.0520$ & $0.2820$ \\
AEW & $+$ PAR & $-0.0136$ & $0.0205$ & $-0.1748$ & $0.1588$ & $-0.0636$ & $0.2564$ \\
\midrule
AESPA & Human & $-0.0579$ & $0.0547$ & $-0.5468$ & $0.4752$ & $-0.5720$ & $0.5659$ \\
AESPA & FM-BC & $-0.0132$ & $0.0150$ & $-0.1592$ & $0.1422$ & $-0.2988$ & $0.2408$ \\
AESPA & $+$ PAR & $-0.0281$ & $0.0342$ & $-0.1676$ & $0.1532$ & $-0.2963$ & $0.2291$ \\
\midrule
AEPBC & Human & $-0.0517$ & $0.0451$ & $-0.4771$ & $0.4135$ & $-0.4578$ & $0.5285$ \\
AEPBC & FM-BC & $-0.0138$ & $0.0135$ & $-0.1338$ & $0.1181$ & $-0.2554$ & $0.2952$ \\
AEPBC & $+$ PAR & $-0.0250$ & $0.0316$ & $-0.1359$ & $0.1217$ & $-0.2285$ & $0.2539$ \\
\midrule
AWW & Human & $-0.0500$ & $0.0433$ & $-0.4595$ & $0.3979$ & $-0.4864$ & $0.5834$ \\
AWW & FM-BC & $-0.0329$ & $0.0229$ & $-0.1600$ & $0.1420$ & $-0.4495$ & $0.5170$ \\
AWW & $+$ PAR & $-0.0349$ & $0.0365$ & $-0.1579$ & $0.1428$ & $-0.3536$ & $0.4009$ \\
\bottomrule
\end{tabular}}
\ifdim\wd\resulttable>\linewidth\resizebox{\linewidth}{!}{\usebox{\resulttable}}\else\usebox{\resulttable}\fi
\caption{Human replay, FM-BC, and PAR kinematics. Each cell averages per-episode extrema, then averages policy seeds (one replay for the human). Acceleration is in m/s$^2$ and drift in radians. $\tilde\kappa$ is the legacy heading-based statistic; divide by the 10\,s timestep to obtain heading change per travelled metre above its speed threshold. These reference-window summaries include unchanged nominal actions.}
\label{tab:kinematics}
\end{table}

%% file: tables/learning_kinematics.tex
\begin{table}[!htbp]
\centering\small
\setlength{\tabcolsep}{4pt}
\sbox{\resulttable}{%
\begin{tabular}{llrrrrrr}
\toprule
Region & Method & min $\tilde\kappa$ & max $\tilde\kappa$ & min $a$ & max $a$ & min drift & max drift \\
\midrule
AEW & FM-BC & $-0.0076$ & $0.0055$ & $-0.1705$ & $0.1526$ & $-0.0520$ & $0.2820$ \\
AEW & NF-BC & $-0.0198$ & $0.0159$ & $-0.3447$ & $0.3031$ & $-0.1615$ & $0.3908$ \\
AEW & ReinFlow & $-0.0117$ & $0.0088$ & $-0.1641$ & $0.1487$ & $-0.1984$ & $0.4094$ \\
AEW & HR-PPO & $-0.0136$ & $0.0090$ & $-0.0515$ & $0.0631$ & $-0.1062$ & $0.2922$ \\
AEW & PPO & $-0.1093$ & $0.1378$ & $-0.2958$ & $0.6312$ & $-2.1146$ & $2.0384$ \\
AEW & QC-FQL & $-0.0054$ & $0.0041$ & $-0.1620$ & $0.1455$ & $-0.0322$ & $0.2407$ \\
AEW & GLRP (one-shot) & $-0.4580$ & $0.4638$ & $-0.3939$ & $0.4145$ & $-0.2352$ & $0.2544$ \\
AEW & TrAISformer (one-shot) & $-0.2433$ & $0.2274$ & $-0.3109$ & $0.3117$ & $-0.2436$ & $0.3660$ \\
\midrule
AESPA & FM-BC & $-0.0132$ & $0.0150$ & $-0.1592$ & $0.1422$ & $-0.2988$ & $0.2408$ \\
AESPA & NF-BC & $-0.0338$ & $0.0318$ & $-0.3270$ & $0.2883$ & $-0.4602$ & $0.4250$ \\
AESPA & ReinFlow & $-0.0169$ & $0.0202$ & $-0.1624$ & $0.1460$ & $-0.4285$ & $0.3890$ \\
AESPA & HR-PPO & $-0.0180$ & $0.0118$ & $-0.0413$ & $0.0577$ & $-0.2244$ & $0.2498$ \\
AESPA & PPO & $-0.1462$ & $0.1094$ & $-0.2111$ & $0.0959$ & $-1.8930$ & $2.1720$ \\
AESPA & QC-FQL & $-0.0096$ & $0.0128$ & $-0.1559$ & $0.1398$ & $-0.2048$ & $0.1671$ \\
AESPA & GLRP (one-shot) & $-0.4956$ & $0.4764$ & $-0.4734$ & $0.5027$ & $-0.4450$ & $0.5490$ \\
AESPA & TrAISformer (one-shot) & $-0.2748$ & $0.2573$ & $-0.3659$ & $0.3655$ & $-0.3366$ & $0.3601$ \\
\midrule
AEPBC & FM-BC & $-0.0138$ & $0.0135$ & $-0.1338$ & $0.1181$ & $-0.2554$ & $0.2952$ \\
AEPBC & NF-BC & $-0.0379$ & $0.0325$ & $-0.3201$ & $0.2787$ & $-0.4401$ & $0.5057$ \\
AEPBC & ReinFlow & $-0.0179$ & $0.0175$ & $-0.1353$ & $0.1237$ & $-0.3747$ & $0.4224$ \\
AEPBC & HR-PPO & $-0.0187$ & $0.0134$ & $-0.0457$ & $0.0591$ & $-0.2109$ & $0.2701$ \\
AEPBC & PPO & $-0.0861$ & $0.1265$ & $-0.2967$ & $0.8809$ & $-1.6912$ & $1.7310$ \\
AEPBC & QC-FQL & $-0.0073$ & $0.0100$ & $-0.1228$ & $0.1096$ & $-0.1586$ & $0.1840$ \\
AEPBC & GLRP (one-shot) & $-0.4600$ & $0.4756$ & $-0.4763$ & $0.5121$ & $-0.4061$ & $0.4375$ \\
AEPBC & TrAISformer (one-shot) & $-0.2384$ & $0.2271$ & $-0.3848$ & $0.3798$ & $-0.3105$ & $0.3385$ \\
\midrule
AWW & FM-BC & $-0.0329$ & $0.0229$ & $-0.1600$ & $0.1420$ & $-0.4495$ & $0.5170$ \\
AWW & NF-BC & $-0.0562$ & $0.0423$ & $-0.4261$ & $0.3729$ & $-0.7420$ & $0.7532$ \\
AWW & ReinFlow & $-0.0368$ & $0.0256$ & $-0.1648$ & $0.1503$ & $-0.5916$ & $0.6521$ \\
AWW & HR-PPO & $-0.0230$ & $0.0120$ & $-0.0764$ & $0.0777$ & $-0.3460$ & $0.3877$ \\
AWW & PPO & $-0.0042$ & $0.0199$ & $-0.2388$ & $1.3943$ & $-1.1691$ & $0.9326$ \\
AWW & QC-FQL & $-0.0177$ & $0.0160$ & $-0.1228$ & $0.1129$ & $-0.3311$ & $0.3858$ \\
AWW & GLRP (one-shot) & $-0.4759$ & $0.4844$ & $-0.5297$ & $0.5771$ & $-0.4437$ & $0.5117$ \\
AWW & TrAISformer (one-shot) & $-0.2724$ & $0.2806$ & $-0.4608$ & $0.4602$ & $-0.6113$ & $0.5994$ \\
\bottomrule
\end{tabular}}
\ifdim\wd\resulttable>\linewidth\resizebox{\linewidth}{!}{\usebox{\resulttable}}\else\usebox{\resulttable}\fi
\caption{Kinematic summaries for all learning baselines, with the same conventions as Table~\ref{tab:kinematics}. Three-seed means of per-episode extrema; values near low speed and AIS heading discontinuities can dominate these statistics. They do not establish hydrodynamic realism. Trajectory-prediction rows are separate evaluations, three seeds each, executed open-loop from the first step.}
\label{tab:learningkinematics}
\end{table}

%% file: tables/selection_kinematics.tex
\begin{table}[!htbp]
\centering\small
\setlength{\tabcolsep}{3pt}
\sbox{\resulttable}{%
\begin{tabular}{llrrrrrr}
\toprule
Region & Method & min $\tilde\kappa$ & max $\tilde\kappa$ & min $a$ & max $a$ & min drift & max drift \\
\midrule
AEW & FM-BC & $-0.0076$ & $0.0055$ & $-0.1705$ & $0.1526$ & $-0.0520$ & $0.2820$ \\
AEW & DEAS-BoN, $N=16$ & $-0.0079$ & $0.0061$ & $-0.3069$ & $0.2728$ & $-0.0568$ & $0.2788$ \\
AEW & FM-BC (seed 0) & $-0.0075$ & $0.0055$ & $-0.1732$ & $0.1545$ & $-0.0481$ & $0.2846$ \\
AEW & Filtered DEAS-BoN, $N=256$ & $-0.0146$ & $0.0126$ & $-0.5192$ & $0.4754$ & $-0.1288$ & $0.3518$ \\
\midrule
AESPA & FM-BC & $-0.0132$ & $0.0150$ & $-0.1592$ & $0.1422$ & $-0.2988$ & $0.2408$ \\
AESPA & DEAS-BoN, $N=16$ & $-0.0149$ & $0.0164$ & $-0.2848$ & $0.2488$ & $-0.2768$ & $0.2419$ \\
AESPA & FM-BC (seed 0) & $-0.0130$ & $0.0147$ & $-0.1576$ & $0.1410$ & $-0.2914$ & $0.2336$ \\
AESPA & Filtered DEAS-BoN, $N=256$ & $-0.0281$ & $0.0278$ & $-0.4768$ & $0.4315$ & $-0.3733$ & $0.3460$ \\
\midrule
AEPBC & FM-BC & $-0.0138$ & $0.0135$ & $-0.1338$ & $0.1181$ & $-0.2554$ & $0.2952$ \\
AEPBC & DEAS-BoN, $N=16$ & $-0.0142$ & $0.0145$ & $-0.2239$ & $0.1935$ & $-0.2306$ & $0.2883$ \\
AEPBC & FM-BC (seed 0) & $-0.0135$ & $0.0139$ & $-0.1339$ & $0.1189$ & $-0.2401$ & $0.2913$ \\
AEPBC & Filtered DEAS-BoN, $N=256$ & $-0.0238$ & $0.0221$ & $-0.4051$ & $0.3517$ & $-0.2882$ & $0.3484$ \\
\midrule
AWW & FM-BC & $-0.0329$ & $0.0229$ & $-0.1600$ & $0.1420$ & $-0.4495$ & $0.5170$ \\
AWW & DEAS-BoN, $N=16$ & $-0.0295$ & $0.0224$ & $-0.2302$ & $0.2049$ & $-0.4358$ & $0.5102$ \\
AWW & FM-BC (seed 0) & $-0.0351$ & $0.0243$ & $-0.1597$ & $0.1439$ & $-0.4562$ & $0.5323$ \\
AWW & Filtered DEAS-BoN, $N=256$ & $-0.0324$ & $0.0237$ & $-0.3655$ & $0.3248$ & $-0.4874$ & $0.5634$ \\
\bottomrule
\end{tabular}}
\ifdim\wd\resulttable>\linewidth\resizebox{\linewidth}{!}{\usebox{\resulttable}}\else\usebox{\resulttable}\fi
\caption{Kinematics of value selection and candidate filtering. The FM-BC and DEAS-BoN rows average three policy seeds; the FM-BC (seed 0) row accompanies filtered DEAS-BoN, which uses seed 0. Each value first averages per-episode extrema. Units and low-speed limitations follow Table~\ref{tab:kinematics}.}
\label{tab:selectionkinematics}
\end{table}

%% file: tables/contact_cases.tex
\begin{table}[!htbp]
\centering\small
\setlength{\tabcolsep}{4pt}
\sbox{\resulttable}{%
\begin{tabular}{lrrr}
\toprule
Partner & Exposed vessel-episodes & FM-BC pairs & PAR pairs \\
\midrule
Cargo & $80041$ & $785$ & $30$ \\
Tanker & $164941$ & $1059$ & $55$ \\
Service & $3321$ & $10$ & $1$ \\
Fishing operations & $6766$ & $49$ & $0$ \\
Passenger & $457$ & $4$ & $0$ \\
Other & $10301$ & $87$ & $5$ \\
Unknown & $12190$ & $75$ & $3$ \\
\midrule
Moving & $45700$ & $232$ & $56$ \\
Anchored & $232317$ & $1837$ & $38$ \\
\bottomrule
\end{tabular}}
\ifdim\wd\resulttable>\linewidth\resizebox{\linewidth}{!}{\usebox{\resulttable}}\else\usebox{\resulttable}\fi
\caption{AEW seed-0 collision-case scan, 7,600 episodes. A pair is an episode and partner vessel with at least one sampled collision. Exposure counts a vessel within 1 nautical mile at least once along the FM-BC trajectory. The motion split is a separate partition of the same population. Anchored means maximum recorded episode speed below 0.26\,m/s. FM-BC counts come from a separate rollout of the same seed-0 checkpoint.}
\label{tab:caseclasses}
\end{table}

%% file: tables/replay_throughput.tex
\begin{table}[!htbp]
\centering\small
\setlength{\tabcolsep}{3pt}
\sbox{\resulttable}{%
\begin{tabular}{llrrrr}
\toprule
Control & Simulator / hardware & Worlds & Median & Minimum & Maximum \\
\midrule
Single & \sns{}, 1 process & $1{,}024$ & $207$ & $207$ & $207$ \\
Single & \sns{}, up to 32 processes & $64$ & $3{,}153$ & $3{,}139$ & $3{,}157$ \\
Single & \jns{}, RTX 3090 & $4{,}096$ & $3{,}105{,}403$ & $3{,}103{,}203$ & $3{,}108{,}024$ \\
Single & \jns{}, H100 NVL & $4{,}096$ & $9{,}215{,}237$ & $9{,}199{,}142$ & $9{,}219{,}120$ \\
All vessels & \jns{}, RTX 3090 & $1{,}024$ & $53{,}591{,}266$ & $53{,}503{,}247$ & $53{,}639{,}737$ \\
All vessels & \jns{}, H100 NVL & $4{,}096$ & $124{,}156{,}972$ & $124{,}109{,}573$ & $124{,}192{,}629$ \\
\bottomrule
\end{tabular}}
\ifdim\wd\resulttable>\linewidth\resizebox{\linewidth}{!}{\usebox{\resulttable}}\else\usebox{\resulttable}\fi
\caption{Peak observed replay throughput across the tested world counts, in valid controlled-vessel steps per second. Each entry summarizes three timing repeats at the world count with the highest median. Single control counts the ego only; all-vessel control counts every valid vessel, including stationary ones. GPU measurements omit learned policy inference, reward, and full benchmark metrics; CPU steps include built-in metrics. Termination and sampled populations also differ (Appendix~\ref{app:compute}).}
\label{tab:replaythroughput}
\end{table}

%% file: dynamics_details.tex
\section{Dynamics models and the external-effect term}
\label{app:dynamics}

Equation~\ref{eq:dynamics} defines the delta action space used throughout the
main results. This appendix states the three further dynamics models the
benchmark supports, together with the sea-current term that couples them to an
external forcing field. Each model is given by its forward law and its inverse,
because the inverse is what extracts expert actions from a recorded track. A
model whose inverse cannot reproduce a recording cannot be trained on that
recording by behavior cloning.

\subsection{Notation}
\label{app:dyn-notation}

Positions $p_k=(x_k,y_k)$ are in the local metric frame of
Equation~\ref{eq:dynamics} and headings $\psi_k$ follow the mathematical
convention ($0$ east, counter-clockwise), with step $\Delta t=10$\,s and step
index $k$. We write
\begin{equation}
 f(\psi)=(\cos\psi,\ \sin\psi),\qquad
 n(\psi)=(-\sin\psi,\ \cos\psi),\qquad
 \Delta p_k=p_{k+1}-p_k,
\end{equation}
so $f(\psi)$ is the forward (bow) unit vector and $n(\psi)$ the port unit
vector, $90^\circ$ to the left of the bow. The operator
$\operatorname{wrap}(\theta)=((\theta+\pi)\bmod 2\pi)-\pi$ maps an angle to
$(-\pi,\pi]$, and $\operatorname{atan2}(y,x)$ is the two-argument arctangent.
The admissible action set $\mathcal A$ is Equation~\ref{eq:bound}.

\paragraph{Reproducing a recording.}
Each model below is stated as a pair. The inverse reads two consecutive
recorded states and returns the action that should join them; the forward law
takes a state and an action and returns the next state. Running the two in
sequence answers a simple question: if we extract the action from a recorded
step and then execute it in the simulator, do we arrive at the position the
vessel actually reached?

When the answer is yes at every step, the model reproduces the recording
exactly, and the demonstrations it extracts are consistent with the simulator
that will execute them. When the answer is no, the extracted action does not
take the vessel where the recording says it went. A BC policy trained on such
data inherits that gap before it makes any error of its own, so the gap limits
what the policy can achieve. Table~\ref{tab:dynamics} reports, for each model,
whether it reproduces a recording exactly.

\subsection{The external-effect term}

Let $U_k=U(p_k,t_k)$ be the surface current, a two-dimensional velocity
sampled at the vessel's own position and wall-clock time at the \emph{start} of
the step. We take it from the HYCOM global ocean
reanalysis~\citep{chassignet2007hycom}, whose surface velocity fields are
interpolated in space and time to the query. The drift is decomposed on the vessel's own axes,
\begin{equation}
 g^{f}_k=\Delta t\,c_f\langle U_k,f(\psi_k)\rangle,\qquad
 g^{p}_k=\Delta t\,c_p\langle U_k,n(\psi_k)\rangle,\qquad
 \delta_k=g^{f}_k f(\psi_k)+g^{p}_k n(\psi_k),
 \label{eq:drift}
\end{equation}
where $g^{f}_k$ and $g^{p}_k$ are the displacements in metres contributed along
the bow and to port, $\delta_k$ is their vector sum, and the dimensionless
gains $c_f,c_p$ give the fraction of the water's motion taken up along each
axis. Because $f$ and $n$ are orthonormal, the isotropic case collapses to a
heading-free expression,
\begin{equation}
 c_f=c_p=c\quad\Longrightarrow\quad \delta_k=\Delta t\,c\,U_k .
 \label{eq:isotropic}
\end{equation}
We use $c_f=c_p=1$, so that the vessel is advected by the surface current. This
is a kinematic assertion rather than a fitted hull coefficient. The port gain
$c_p$ cannot be fitted from these recordings, because it never enters the
extracted action of any model below, so asserting it is preferable to reporting
a value the data does not constrain.

The field is queried with a coverage weight. Outside its spatial or temporal
domain we set $\delta_k=0$, which asserts \emph{no push} rather than \emph{no
current}. The two are different claims and only the first is supported.

\subsection{Querying the current field}
\label{app:current-query}

The reanalysis supplies eastward and northward surface velocity on a regular
longitude--latitude grid at a nominal $1/12^\circ$ spacing, about $9$\,km at
this latitude, with three-hourly output. A query $(p_k,t_k)$ is answered in
three steps. The position is converted from the region's local metric frame
back to geographic coordinates; the two velocity components are interpolated
trilinearly, bilinearly over the four surrounding grid nodes and linearly
between the two bracketing snapshots in time; and the result is rotated into
the local frame to give $U_k$ in metres per second. Interpolation is applied to
the velocity components separately, so speed and direction are both derived
quantities rather than interpolated directly.

Each query also returns a coverage weight $w_k\in[0,1]$, the product of the
in-domain indicators for the spatial and temporal axes and the validity of the
surrounding nodes, and the drift of Equation~\ref{eq:drift} is scaled by it.
Land and out-of-domain nodes are invalid, so $w_k$ falls to zero near the
boundary rather than stepping, and a vessel leaving coverage is not given a
discontinuous shove.

Figure~\ref{fig:hycom-overview} places all eight benchmark regions in the field
at one instant. It is drawn by sampling \texttt{HycomCurrentField} exactly as
the dynamics query it, so the plotted field is the one the vessels are advected
by rather than a separate rendering of the source product. Over the full extent
shown the speed averages $0.317$\,kn and peaks at $1.651$\,kn. The four
navigation regions lie along the strait, where the through-flow is strongest
and most nearly aligned with the traffic lanes, while the four anchorage groups
sit in weaker water to the north. The current term therefore matters most in
exactly the regions the dynamics study uses.

\definecolor{regionnav}{HTML}{D55E00}
\definecolor{regionanch}{HTML}{CC79A7}
\begin{figure}[t]
\centering
\includegraphics[width=\textwidth]{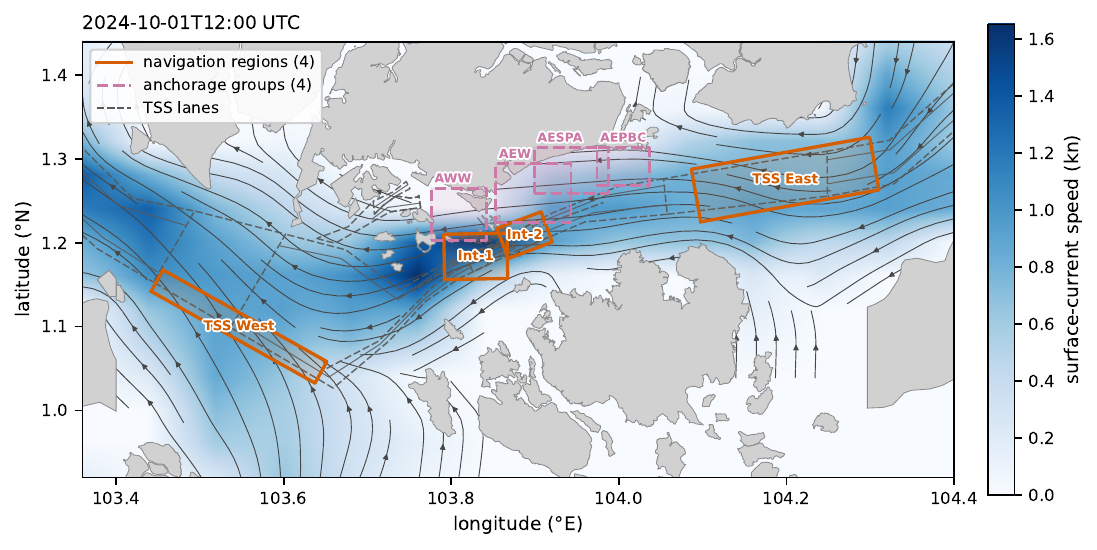}
\caption{All eight benchmark regions in the Singapore Strait, over the HYCOM
surface current at one instant. Shading gives current speed in knots and
streamlines give direction, sampled over $103.36$--$104.40^\circ$E and
$0.92$--$1.44^\circ$N through the same query path the dynamics use. Solid
\textcolor{regionnav}{vermillion} outlines mark the four navigation regions (TSS West, TSS East,
Int-1, Int-2), drawn from each configuration's \texttt{region\_wkt} so the
rotations are the real ones rather than axis-aligned approximations; dashed
\textcolor{regionanch}{purple} outlines mark the four anchorage groups (AWW, AEW, AESPA, AEPBC). The
two families differ in both hue and line style so they remain distinguishable
in grayscale and under colour-vision deficiency. The neutral basemap behind
them is the traffic separation scheme: 16 lanes, 21 separation lines, and 4
separation zones, taken from OpenStreetMap \texttt{seamark:*} data
(\copyright{} OpenStreetMap contributors, ODbL). The strongest flow follows the
strait through the navigation regions, while the anchorage groups lie in weaker
water to the north.}
\label{fig:hycom-overview}
\end{figure}

\begin{figure}[t]
\centering
\includegraphics[width=\textwidth]{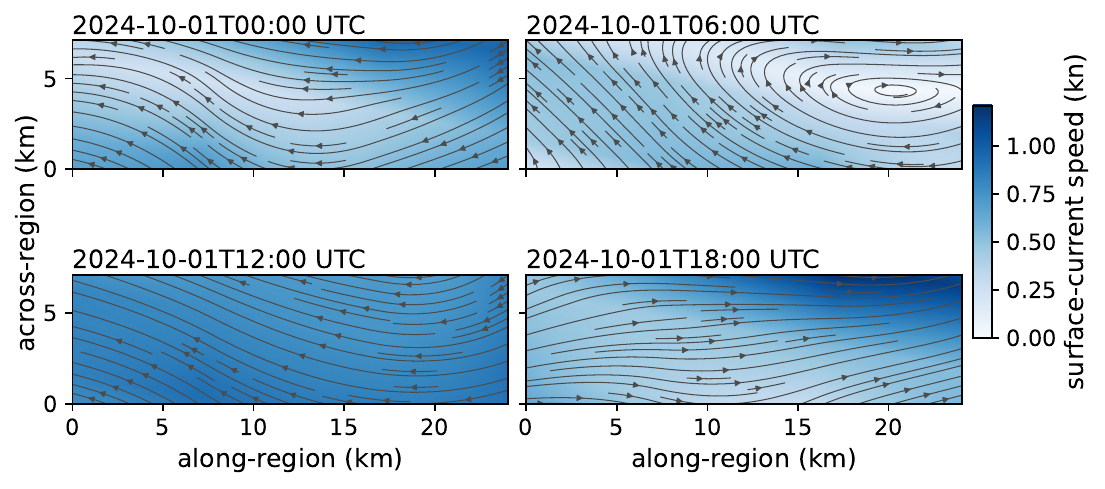}
\caption{HYCOM surface current over the TSS East region on 1 October 2024, at
four times through the day. Shading gives current speed in knots and
streamlines give direction, both in the region's local metric frame. The flow
reverses with the tide over the day and is not uniform across the region; the
hatched wedge marks positions outside the field's coverage, where the drift
term is set to zero. Vessel tracks in this region are advected by the field
sampled at their own position and time, as described in
Appendix~\ref{app:current-query}.}
\label{fig:hycom}
\end{figure}

Figure~\ref{fig:hycom} shows the resulting field over TSS East. The
interpolation is what makes the field continuous at vessel scale; it does not
add information. The region is roughly $24$\,km along by $7$\,km across, so it
spans only a few native cells, and the field it supplies is best read as a
slowly varying regional set rather than resolved local structure. This is the
same reason the port gain $c_p$ is asserted rather than fitted: neither the
recordings nor the field constrain circulation at the scale of a single
vessel's manoeuvre.

\subsection{Delta dynamics with current}

The action of Equation~\ref{eq:dynamics}, inherited from
\sns{}~\citep{pham2025shipnavisim} and advected by $\delta_k$:
\begin{align}
 \text{forward:}\quad & p_{k+1}=p_k+(\Delta x_k,\Delta y_k)+\delta_k,
 && \psi_{k+1}=\operatorname{wrap}(\psi_k+\Delta\psi_k),\\
 \text{inverse:}\quad & (\Delta x_k,\Delta y_k)=\Delta p_k-\delta_k,
 && \Delta\psi_k=\operatorname{wrap}(\psi_{k+1}-\psi_k).
\end{align}
Three action components act against three state targets, so the extracted
action always reproduces the recorded step exactly. With $U\equiv0$ the model is
bit-identical to Equation~\ref{eq:dynamics}, which we verify numerically on
every region frame, so the current-free results remain directly comparable to
the main text.

\subsection{Port-glide dynamics}

Forward thrust and turn, $a_k=(D_k,\Delta\psi_k)$, where $D_k$ is the commanded
forward displacement in metres along the bow. The delta model of the previous
subsection commands displacement on all three degrees of freedom, but a
conventional surface ship carries no actuator on the sway axis: propeller and
rudder produce surge and yaw, and sway follows from them rather than being
commanded, which is what makes such a vessel underactuated in the sense
of~\citet{do2010underactuated}. The two-component models below take that constraint
seriously, and they differ only in what they then do with the heading channel.
We define port-glide here rather than adopting it from prior work; it is the
most direct way to drop the sway command while keeping the recorded heading,
which is read from the recording:
\begin{align}
 \text{forward:}\quad & \hat p_{k+1}=p_k+D_k f(\psi_k)+\delta_k,
 && \psi_{k+1}=\operatorname{wrap}(\psi_k+\Delta\psi_k),\\
 \text{inverse:}\quad & D_k=\langle\Delta p_k,f(\psi_k)\rangle-g^{f}_k,
 && \Delta\psi_k=\operatorname{wrap}(\psi_{k+1}-\psi_k).
\end{align}
Two action components act against three state targets, and the model's own
output $\hat p_{k+1}$ is written separately from the recorded $p_{k+1}$ because
here the two differ. Only motion along the bow can be commanded, so any
sideways motion in the recording that the current does not account for cannot
be reproduced. Executing the extracted action therefore leaves
$\hat p_{k+1}\neq p_{k+1}$, short of the recording by the sideways distance
\begin{equation}
 \big|\langle\Delta p_k,n(\psi_k)\rangle-g^{p}_k\big|,
 \label{eq:sideways}
\end{equation}
which is generally not zero. The model is asserting that a vessel cannot be
\emph{commanded} sideways, which is correct, and in addition that it cannot
\emph{move} sideways, which is not: an underactuated hull still acquires sway
through drift, leeway, and the yaw--sway coupling of its own
turn~\citep{do2010underactuated}. Conflating the two is a modelling choice rather than a
defect, but the residual accumulates over a trajectory, so under this model the
expert does not follow its own recorded track.

\subsection{Water-track dynamics}

The missing sideways motion of Equation~\ref{eq:sideways} disappears if the
heading is \emph{defined} as the direction of motion through the water rather
than read from the recording. Assuming $c_f=c_p$ so that
Equation~\ref{eq:isotropic} applies and the drift is the heading-free
$\delta_k$, let
\begin{equation}
 w_k=\Delta p_k-\delta_k
 \label{eq:watertrack}
\end{equation}
be the through-water displacement, that is, the part of the motion not
attributable to the current. The action is again $(D_k,\Delta\psi_k)$, but the
model turns before it translates:
\begin{align}
 \text{forward:}\quad & \psi_{k+1}=\operatorname{wrap}(\psi_k+\Delta\psi_k),
 && p_{k+1}=p_k+D_k f(\psi_{k+1})+\delta_k,\\
 \text{inverse:}\quad & D_k=\lVert w_k\rVert\ \ge 0,
 && \Delta\psi_k=\operatorname{wrap}\!\big(\operatorname{atan2}(w_{k,y},w_{k,x})-\psi_k\big).
\end{align}
Exactness is then an identity rather than an approximation, because
$D_k f(\psi_{k+1})\equiv w_k$ by construction:
\begin{equation}
 p_k+D_k f(\psi_{k+1})+\delta_k=p_k+w_k+\delta_k=p_k+\Delta p_k=p_{k+1},
\end{equation}
so any recorded track is reproduced to floating-point precision, whatever
sideways motion it contains.

Three consequences follow. First, this is the zero-sideslip assumption: a hull
with no leeway points along its own water track, and the angle between heading
and course over ground, the drift angle in the sense
of~\citet{fossen2011handbook}, is then produced by the current rather than
asserted. Water-track keeps the same two-component, no-sway-actuator command
set as port-glide~\citep{do2010underactuated}, but recovers the missing sideways motion
by letting the heading follow the through-water displacement instead of fixing
it from the recording, so underactuation costs nothing in reconstruction.
Second, $\psi_k$ becomes the course of the step just taken, a post-step
quantity, whereas the models above carry a pre-step one. Third, the model
rewrites the heading channel of the data and not only the action column, so the
observation, the hull orientation used for clearance geometry, and the
curvature statistics all follow the redefined heading. Water-track results are
therefore comparable to the others on position-based metrics such as goal
reaching, but not on clearance metrics, which depend on hull orientation.

Two degenerate cases are handled explicitly. A through-water step shorter than
$0.5$\,m has no defined course and holds the previous heading, and the final
step of a window inherits the heading of its predecessor.

\subsection{Nomoto dynamics}
\label{app:nomoto}

Water-track with the heading command passed through a first-order lag, in the
spirit of the classical first-order steering model of \citet{nomoto1957steering};
see \citet{fossen2011handbook} for a modern treatment. The action set remains
surge and yaw only~\citep{do2010underactuated}, and the lag is what the two available
actuators impose on the yaw channel: a rudder order changes the turn rate over
several steps rather than instantly. The position law, $D_k$, and the drift are
unchanged, and only the heading channel differs. With momentum parameter $\alpha\in[0,1)$ and action
$a_k=(D_k,a^{\psi}_k)$, where $a^{\psi}_k$ is the commanded turn,
\begin{align}
 \text{forward:}\quad & \Delta\psi_k=\alpha\,\Delta\psi_{k-1}+(1-\alpha)\,a^{\psi}_k,\\
 \text{inverse:}\quad & a^{\psi}_k=\frac{\Delta\psi_k-\alpha\,\Delta\psi_{k-1}}{1-\alpha},
\end{align}
with $\Delta\psi_{-1}=0$, and with $\Delta\psi_k$ obtained from the water-track
inverse above. At $\alpha=0$ the model reduces exactly to water-track. For
$\alpha>0$ a held command settles onto a constant turn rate, which is the
defining behavior of a first-order steering response. Recordings are
reproduced exactly for every $\alpha$ in range. The inverse needs only the previous heading increment
$\Delta\psi_{k-1}$, so no additional state variable is carried.

We use this discrete lag rather than a discretization of the continuous
first-order equation $T\dot r+r=K\delta^{\mathrm{rud}}$ of
\citet{nomoto1957steering}, in which $r$ is the yaw rate,
$\delta^{\mathrm{rud}}$ the rudder angle, $T$ a time constant and $K$ a gain.
That equation carries the instantaneous yaw rate, which two headings cannot
recover, since $(\psi_k-\psi_{k-1})/\Delta t$ is the average rate over the step
and the two differ substantially during transients. More importantly, $K$ and
$T$ are not identifiable from these recordings, because the rudder is never
observed and every parameter pair reproduces the heading sequence exactly.
There is consequently no true time constant that the discrete form fails to
honor. The parameter $\alpha$ plays the role of $e^{-\Delta t/T}$ and is swept
rather than fitted.

The inverse divides by $1-\alpha$, so the admissible range of $a^{\psi}$ must
exceed the turn limit of $\mathcal A$ by a factor $(1+\alpha)/(1-\alpha)$.
Retaining the bound of Equation~\ref{eq:bound} unchanged clips the extracted
action and silently destroys the reconstruction.

\input{figures/dynamics_vectors}

\subsection{Summary}

\begin{table}[h]
\centering
\small
\begin{tabular}{@{}llc>{\raggedright\arraybackslash}p{3.7cm}>{\raggedright\arraybackslash}p{3.0cm}@{}}
\toprule
Model & Action & Dim. & Heading & Reproduces a recording \\
\midrule
Delta       & $(\Delta x,\Delta y,\Delta\psi)$ & 3 & taken from the recording, controlled independently & exactly \\
Port-glide  & $(D,\Delta\psi)$                 & 2 & taken from the recording & no, misses sideways motion \\
Water-track & $(D,\Delta\psi)$                 & 2 & estimated from the movement & exactly \\
Nomoto      & $(D,a^{\psi})$                   & 2 & estimated from the movement, then filtered & exactly \\
\bottomrule
\end{tabular}
\caption{The four dynamics models. The last column asks whether executing the
extracted action lands the vessel where the recording says it went. Delta
achieves this with a third degree of freedom; water-track achieves it instead
by assuming the vessel cannot slip sideways. Port-glide cannot, and the motion
it fails to reproduce limits what a BC policy trained on it can achieve.}
\label{tab:dynamics}
\end{table}

These models support two comparisons: whether a vessel's heading is an
independent control or is determined by where the vessel is going, and what an
external forcing field costs a policy under either assumption.

%% file: figures/dynamics_vectors.tex
\begin{figure}[t]
\centering
\begin{tikzpicture}[vecpic]

\begin{scope}[shift={(0,0)}]
  \node[ttl] at (1.75,2.15) {Delta};
  \coordinate (dtend) at (2.75,0.80);
  \hullglyph{20}
  \node[lbl,black!60] at (-0.10,-0.62) {$p_k$};
  \hullatc{(dtend)}{58}
  \node[lbl,black!60] at ($(dtend)+(-0.05,-0.60)$) {$p_{k+1}$};
  \draw[bow] (0,0) -- (20:1.15);
  \draw[vec,vecact]   (0,0) -- (2.0,1.35)
        node[lbl,vecact,above left=-2pt] {$(\Delta x_k,\Delta y_k)$};
  \draw[vec,vecdrift] (2.0,1.35) -- (dtend)
        node[lbl,vecdrift,above=0pt,pos=0.45] {$\delta_k$};
  \draw[vec,vecgnd]   (0,0) -- (dtend);
  \draw[turn] (20:0.72) arc (20:58:0.72);
  \node[lbl,vecturn] at (64:1.15) {$\Delta\psi_k$};
\end{scope}

\begin{scope}[shift={(6.3,0)}]
  \node[ttl] at (1.75,2.15) {Port-glide};
  \coordinate (pgend) at ($(20:2.1)+(-32:0.95)$);
  \coordinate (rec)   at ($(pgend)+(110:0.95)$);
  \hullglyph{20}
  \node[lbl,black!60] at (-0.10,-0.62) {$p_k$};
  \ghosthullatc{(rec)}{38}
  \hullatc{(pgend)}{38}
  \draw[bow] (0,0) -- (20:2.6);
  \draw[turn] (20:0.80) arc (20:38:0.80);
  \node[lbl,vecturn] at (58:0.86) {$\Delta\psi_k$};
  \draw[vec,vecact]   (0,0) -- (20:2.1)
        node[lbl,vecact,above left=-3pt,pos=0.98] {$D_k f(\psi_k)$};
  \draw[vec,vecdrift] (20:2.1) -- (pgend)
        node[lbl,vecdrift,above=0pt,pos=0.45] {$\delta_k$};
  \draw[vec,vecgnd]   (0,0) -- (pgend);
  \draw[{Stealth[length=1.6mm]}-{Stealth[length=1.6mm]},vecmiss,line width=0.7pt]
        (pgend) -- (rec);
  \node[lbl,vecmiss,right=1pt] at ($(pgend)!0.5!(rec)$) {missed};
  \node[lbl,black!55,above=4pt] at (rec) {recorded $p_{k+1}$};
\end{scope}

\begin{scope}[shift={(0,-4.6)}]
  \node[ttl] at (1.75,2.15) {Water-track};
  \coordinate (wtend) at ($(55:2.0)+(-20:1.0)$);
  \hullglyph{55}
  \node[lbl,black!60] at (-0.10,-0.62) {$p_k$};
  \hullatc{(wtend)}{55}
  \node[lbl,black!60] at ($(wtend)+(-0.05,-0.60)$) {$p_{k+1}$};
  \draw[bow] (0,0) -- (20:1.55);
  \draw[turn] (20:0.95) arc (20:55:0.95);
  \node[lbl,vecturn] at (43:1.22) {$\Delta\psi_k$};
  \draw[vec,vecact]   (0,0) -- (55:2.0)
        node[lbl,vecact,above left=-3pt,pos=0.6] {$D_k f(\psi_{k+1})$};
  \draw[vec,vecdrift] (55:2.0) -- (wtend)
        node[lbl,vecdrift,above=0pt,pos=0.45] {$\delta_k$};
  \draw[vec,vecgnd]   (0,0) -- (wtend);
\end{scope}

\begin{scope}[shift={(6.3,-4.6)}]
  \node[ttl] at (1.75,2.15) {Nomoto};
  \coordinate (nmend) at ($(42:2.0)+(-28:1.0)$);
  \hullglyph{42}
  \node[lbl,black!60] at (-0.10,-0.62) {$p_k$};
  \hullatc{(nmend)}{42}
  \node[lbl,black!60] at ($(nmend)+(-0.05,-0.60)$) {$p_{k+1}$};
  \draw[bow] (0,0) -- (20:1.55);
  \draw[turn,dotted] (20:1.25) arc (20:66:1.25);
  \node[lbl,vecturn] at (73:1.44) {$a^{\psi}_k$};
  \draw[-{Stealth[length=1.6mm]},black!70,line width=0.5pt]
        (20:0.95) arc (20:42:0.95);
  \node[lbl,black!70] at (31:1.30) {$\Delta\psi_k$};
  \draw[vec,vecact]   (0,0) -- (42:2.0)
        node[lbl,vecact,above left=-3pt,pos=0.55] {$D_k f(\psi_{k+1})$};
  \draw[vec,vecdrift] (42:2.0) -- (nmend)
        node[lbl,vecdrift,above=0pt,pos=0.45] {$\delta_k$};
  \draw[vec,vecgnd]   (0,0) -- (nmend);
\end{scope}

\end{tikzpicture}
\caption{The forward law of each dynamics model, as a vector sum. From the
position $p_k$, the commanded displacement is \textcolor{vecact}{blue} and the
commanded turn \textcolor{vecturn}{green}, while the drift $\delta_k$
contributed by the sea current is \textcolor{vecdrift}{red}. Their result,
\textcolor{vecgnd}{grey}, is the step over ground the model produces, ending at
the solid hull; the dashed ray is the bow $f(\psi_k)$. Delta commands displacement on both axes and turns
independently of it. Port-glide instead takes its heading from the AIS
recording rather than choosing it, and can only push along that bow. Any
sideways motion in the recording is therefore out of reach, so the vessel ends
up somewhere other than the recorded $p_{k+1}$, drawn as a dashed hull,
indicated in \textcolor{vecmiss}{pale red} (Equation~\ref{eq:sideways}). The other
three reproduce the recorded step $\Delta p_k=p_{k+1}-p_k$ exactly. Water-track
turns first and defines the bow as the direction of travel through the water,
and Nomoto does the same with a first-order lag, so that its commanded turn
$a^{\psi}_k$ (dotted) and the turn $\Delta\psi_k$ actually applied (grey)
differ.}
\label{fig:dynamics-vectors}
\end{figure}
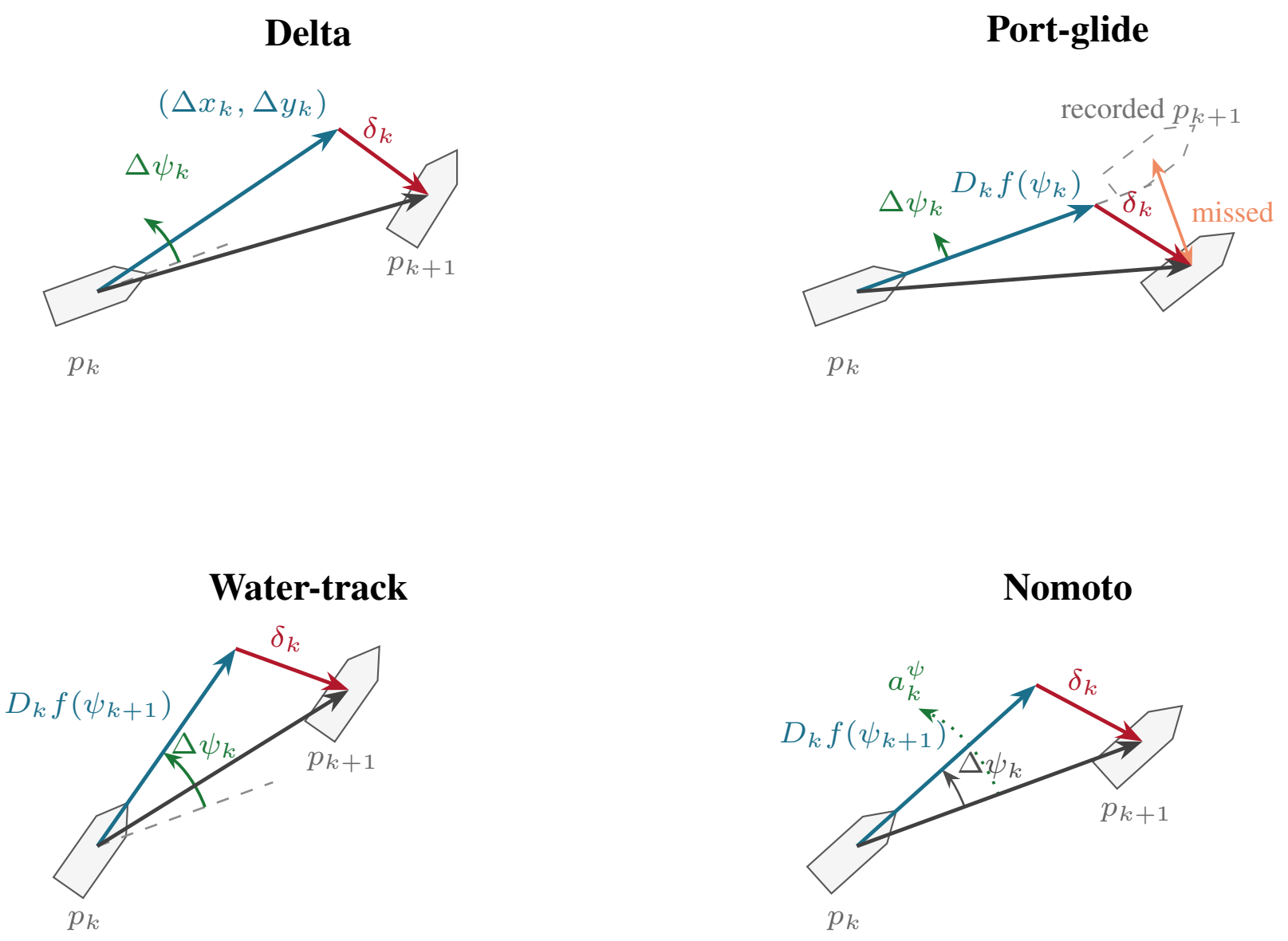

%% file: sections/dynamics_anchorage.tex
\subsection{Dynamics models and the external effect}
\label{app:dynamics-anchorage}
We compare the dynamics model and an unmodelled current in the anchorage
regions AEW and AESPA with FM-BC under the anchorage protocol of
Section~\ref{sec:anchorage_results}, scored with the metrics of
Section~\ref{sec:benchmark} on the full test populations of 7,600 and 7,068
episodes. The current comparison also carries the TSS East rows, so
Table~\ref{tab:dynamics-anchorage-current} reports both settings together.
Each row is a separately trained FM-BC policy whose demonstrations were
extracted through, and whose actions are integrated by, the named model of
Appendix~\ref{app:dynamics}; its checkpoint is the best of the last ten by
$\reachg$, with ties broken by $\gcade$, as in Appendix~\ref{app:methods}. The
forward displacement of the two-component models is bounded at 300\,m on AEW
and 280\,m on AESPA, and the current is not observed by the policy.

\begin{table}[!htbp]
\centering\small
\setlength{\tabcolsep}{4pt}
\sbox{\resulttable}{%
\begin{tabular}{llccccccc}
\toprule
Region & Dynamics model & $\reachg\uparrow$ & Collision rate $C\downarrow$ & CPA (m) & $\gcade\downarrow$ & FDE$_{\neg R}$ (m) & Replay $\reachg$ & Replay $C$ \\
\midrule
AEW & Delta & $0.8880$ & $0.2007$ & $88.3$ & $23.19$ & $429$ & $1.0000$ & $0.0000$ \\
AEW & Port-glide & $0.4433$ & $0.2849$ & $61.6$ & $42.79$ & $945$ & $0.3986$ & $0.1993$ \\
AEW & Water-track & $0.8817$ & $0.2332$ & $64.7$ & $25.32$ & $474$ & $1.0000$ & $0.0171$ \\
AEW & Nomoto & $0.8782$ & $0.2371$ & $64.9$ & $25.54$ & $463$ & $1.0000$ & $0.0170$ \\
\midrule
AESPA & Delta & $0.9317$ & $0.4502$ & $8.7$ & $26.83$ & $552$ & $1.0000$ & $0.0000$ \\
AESPA & Port-glide & $0.2591$ & $0.5074$ & $-1.0$ & $44.65$ & $809$ & $0.2292$ & $0.4428$ \\
AESPA & Water-track & $0.9380$ & $0.4690$ & $5.8$ & $28.31$ & $693$ & $1.0000$ & $0.0204$ \\
AESPA & Nomoto & $0.9345$ & $0.4713$ & $5.1$ & $28.78$ & $711$ & $1.0000$ & $0.0199$ \\
\bottomrule
\end{tabular}}
\ifdim\wd\resulttable>\linewidth\resizebox{\linewidth}{!}{\usebox{\resulttable}}\else\usebox{\resulttable}\fi
\caption{Dynamics models with no current. The Delta rows correspond to Table~\ref{tab:fmseeds}. FDE$_{\neg R}$ is as in Section~\ref{sec:benchmark}. The replay columns replay the human tracks open-loop through the same model; they are a reference, not an upper bound.}
\label{tab:dynamics-anchorage-models}
\end{table}

\begin{table}[!htbp]
\centering\small
\setlength{\tabcolsep}{3pt}
\sbox{\resulttable}{%
\begin{tabular}{llccccccccccc}
\toprule
 & & \multicolumn{5}{c}{Trained without current} & & \multicolumn{5}{c}{Trained with current} \\
\cmidrule(lr){3-7}\cmidrule(lr){9-13}
Region & Dynamics model & $\reachg\uparrow$ & $C\downarrow$ & CPA (m) & $\gcade\downarrow$ & FDE$_{\neg R}$ (m) & & $\reachg\uparrow$ & $C\downarrow$ & CPA (m) & $\gcade\downarrow$ & FDE$_{\neg R}$ (m) \\
\midrule
AEW & Delta & $0.6563$ & $0.3149$ & $49.6$ & $32.62$ & $525$ & & $0.6682$ & $0.3175$ & $48.9$ & $31.31$ & $495$ \\
AEW & Port-glide & $0.3639$ & $0.3471$ & $41.8$ & $48.88$ & $969$ & & $0.3647$ & $0.3325$ & $46.3$ & $47.23$ & $914$ \\
AEW & Water-track & $0.7300$ & $0.2337$ & $70.2$ & $33.43$ & $518$ & & $0.8067$ & $0.2180$ & $74.5$ & $27.66$ & $475$ \\
AEW & Nomoto & $0.7539$ & $0.2351$ & $70.4$ & $33.83$ & $505$ & & $0.8092$ & $0.2175$ & $74.0$ & $27.56$ & $471$ \\
\midrule
AESPA & Delta & $0.6142$ & $0.5545$ & $-8.2$ & $35.32$ & $508$ & & $0.6882$ & $0.5447$ & $-6.2$ & $34.02$ & $516$ \\
AESPA & Port-glide & $0.2204$ & $0.5904$ & $-11.0$ & $50.94$ & $977$ & & $0.2732$ & $0.5576$ & $-8.0$ & $50.48$ & $993$ \\
AESPA & Water-track & $0.7600$ & $0.4707$ & $5.0$ & $33.85$ & $519$ & & $0.8838$ & $0.4431$ & $8.6$ & $30.10$ & $565$ \\
AESPA & Nomoto & $0.7586$ & $0.4641$ & $5.9$ & $34.01$ & $518$ & & $0.8820$ & $0.4440$ & $8.9$ & $30.45$ & $583$ \\
\midrule
TSS-East & Delta & $0.9108$ & $0.1423$ & $262.6$ & $119.06$ & $3712$ & & ${0.9203}$ & ${0.1332}$ & $285.1$ & ${114.45}$ & ${3562}$ \\
TSS-East & Port-glide & $0.3758$ & $0.1255$ & $290.8$ & $111.57$ & $3500$ & & ${0.4052}$ & ${0.1251}$ & $295.2$ & ${106.90}$ & ${3368}$ \\
TSS-East & Water-track & ${0.9313}$ & ${0.1192}$ & $307.9$ & $107.51$ & $3408$ & & $0.9176$ & $0.1205$ & $316.4$ & ${103.15}$ & ${3270}$ \\
TSS-East & Nomoto & $0.7472$ & $0.1183$ & $298.2$ & $109.06$ & $3475$ & & ${0.8373}$ & ${0.1169}$ & $309.4$ & ${103.51}$ & ${3267}$ \\
\bottomrule
\end{tabular}}
\ifdim\wd\resulttable>\linewidth\resizebox{\linewidth}{!}{\usebox{\resulttable}}\else\usebox{\resulttable}\fi
\caption{An unmodelled current. Every cell is evaluated with the HYCOM current; the two column groups differ only in whether the policy was trained without or with it. Metrics as in Table~\ref{tab:dynamics-anchorage-models}.}
\label{tab:dynamics-anchorage-current}
\end{table}

Without a current, Delta, Water-track and Nomoto reach similar $\reachg$ in both regions (0.878--0.888 on AEW and 0.932--0.938 on AESPA), while Port-glide drops to 0.443 on AEW and 0.259 on AESPA (Table~\ref{tab:dynamics-anchorage-models}). Human replay through Port-glide reaches only 0.399 and 0.229, because the model cannot reproduce the recording's sideways motion (Equation~\ref{eq:sideways}); the Port-glide policy exceeds its replay, so the replay is a reference rather than an upper bound. Under the current, training with it raises $\reachg$ by $+0.077$/$+0.124$ for Water-track, $+0.055$/$+0.123$ for Nomoto, $+0.012$/$+0.074$ for Delta and $+0.001$/$+0.053$ for Port-glide on AEW/AESPA (Table~\ref{tab:dynamics-anchorage-current}), so the recovery depends on the model and the region. TSS East, in the same table, moves far less: $+0.090$ for Nomoto, $+0.029$ for Port-glide, $+0.010$ for Delta and $-0.014$ for Water-track. The anchorages are the harder case for an unmodelled current, since a vessel manoeuvring slowly among moored hulls accumulates drift over a long approach, whereas a transiting vessel crosses the field quickly. Three caveats apply. Water-track and Nomoto orient the hull by the through-water heading (Section~\ref{sec:dynamics}), so their $C$ and CPA are not directly comparable with the Delta rows: replay through them already has $C=0.017$ on AEW and $0.020$ on AESPA, against zero under Delta. Finally, the HYCOM record joins two products on 5 September 2024, so the anchorage training window is mostly non-tidal and the test window tidal; train--test differences under the current carry this confound.

\input{tables/ext_condition_table}

The current can enter in three independent places, and
Table~\ref{tab:radius-conditioning} separates them. It is in the
\emph{simulator} for every row, so every policy is evaluated in it. It is in
the \emph{expert labels} when the vault was built with it, so that the
imitation targets come from inverting the dynamics under the current; for the
through-water models the heading is itself defined as the through-water course,
so the current changes what the policy is asked to reproduce and not only the
physics it meets. It is in the \emph{policy input} when the observation carries
the local current as world-frame components. The bottom row of each block
therefore learned from still-water labels and was then placed in a current, a
transfer result whose absolute value is not directly comparable with the two
matched rows above it. Rollouts ran at the 100\,m radius with the 1{,}000 extra
steps the protocol grants, and the wider radii are recomputed from each
episode's closest approach, which is not truncated at arrival.

Conditioning the policy on the current cuts $\gcade$ for
every dynamics model, from $114.5$ to $79.7$\,m for Delta, $103.2$ to $80.9$ for
Water-track, $106.9$ to $82.9$ for Port-glide and $103.5$ to $83.2$ for Nomoto.
Goal rate moves the other way for the two-component models, falling from
$91.8$\% to $72.2$\% for Water-track and from $83.7$\% to $47.4$\% for Nomoto at
a 200\,m radius, and the gap closes as the radius is relaxed.

The right panel of Figure~\ref{fig:drift-current} shows why the two measures disagree. The goal
signal is strong enough that an unconditioned policy still arrives. The current
carries it off the demonstrated line for most of the transit, and it turns back
towards the goal only at the end, so goal rate records an arrival while
$\gcade$ records the detour that preceded it. A policy that cannot observe the
current is not failing to reach its destination; it is reaching it by a route
the expert never took.

The TSS East dynamics comparison trains one policy per model while keeping the
other settings fixed. Port-glide, the only model that cannot reproduce the
recording, reaches 41.6\% of goals; the three exact-replay models reach
88.2--96.0\%. Collision rate, closest approach, and tracking error differ
little, so the main effect of the dynamics choice is whether the vessel arrives.
\input{dynamics_results}

%% file: tables/ext_condition_table.tex
\begin{table}[t]
\centering
\small
\setlength{\tabcolsep}{4pt}
\begin{tabular}{lcccrrrr}
\toprule
& \multicolumn{3}{c}{Sea current in} & \multicolumn{3}{c}{Goal rate at radius} & gc-ADE \\
\cmidrule(lr){2-4} \cmidrule(lr){5-7}
Motion model & simulator & expert labels & policy input & 200\,m & 500\,m & 1000\,m & (m) \\
\midrule
Delta-current (3-D) & $\checkmark$ & $\checkmark$ & $\checkmark$ & \textbf{93.57} & \textbf{97.46} & \textbf{98.94} & \textbf{79.7} \\
 & $\checkmark$ & $\checkmark$ & $\times$ & 92.03 & 95.88 & 97.50 & 114.5 \\
 & $\checkmark$ & $\times$ & $\times$ & 91.08 & 95.90 & 97.73 & 119.1 \\
\addlinespace
Water-track (2-D) & $\checkmark$ & $\checkmark$ & $\checkmark$ & 72.15 & 90.52 & 96.10 & \textbf{80.9} \\
 & $\checkmark$ & $\checkmark$ & $\times$ & 91.76 & 95.89 & 97.59 & 103.2 \\
 & $\checkmark$ & $\times$ & $\times$ & \textbf{93.13} & \textbf{96.99} & \textbf{98.22} & 107.5 \\
\addlinespace
Port-glide (2-D) & $\checkmark$ & $\checkmark$ & $\checkmark$ & 26.80 & 62.59 & 91.09 & \textbf{82.9} \\
 & $\checkmark$ & $\checkmark$ & $\times$ & \textbf{40.52} & \textbf{79.00} & 94.65 & 106.9 \\
 & $\checkmark$ & $\times$ & $\times$ & 37.58 & 77.69 & \textbf{95.17} & 111.6 \\
\addlinespace
Nomoto (2-D) & $\checkmark$ & $\checkmark$ & $\checkmark$ & 47.36 & 82.67 & 93.27 & \textbf{83.2} \\
 & $\checkmark$ & $\checkmark$ & $\times$ & \textbf{83.73} & \textbf{94.27} & 96.55 & 103.5 \\
 & $\checkmark$ & $\times$ & $\times$ & 74.72 & 94.21 & \textbf{96.58} & 109.1 \\
\bottomrule
\end{tabular}
\caption{Goal rate against the radius it is scored at, on the 92-day held-out split (23,566 episodes per cell). Goal rate is the percentage of episodes whose closest approach fell within the stated radius. Every row is evaluated in the current; the three check columns say where else it enters. Best per column within each dynamics model in bold.}
\label{tab:radius-conditioning}
\end{table}

%% file: dynamics_results.tex
\begin{table}[tbp]
\centering\scriptsize
\setlength{\tabcolsep}{4pt}
\sbox{\resulttable}{%
\begin{tabular}{lccccc}
\toprule
Dynamics & $\reachg\uparrow$ & Collision rate $C\downarrow$ & CPA (m) & $\gcade\downarrow$ & FDE$_{\neg R}\downarrow$ (m) \\
\midrule
Delta & $0.9495$ & $0.1014$ & $359.5$ & $78.54$ & $2452$ \\
Port-glide & $0.4160$ & $0.1071$ & $329.6$ & $80.77$ & $2517$ \\
Water-track & $0.9596$ & $0.0985$ & $358.7$ & $78.27$ & $2452$ \\
Nomoto & $0.8817$ & $0.1017$ & $349.1$ & $80.22$ & $2510$ \\
\bottomrule
\end{tabular}
}
\ifdim\wd\resulttable>\linewidth\resizebox{\linewidth}{!}{\usebox{\resulttable}}\else\usebox{\resulttable}\fi
\caption{Results of FM-BC with different dynamics models on TSS East, solo regime, with no sea current. Columns follow Table~\ref{tab:region_main_solo}.}
\label{tab:dynamics-models}
\vskip -12pt
\end{table}

%% file: navigation_results.tex
\section{Single-agent results in the navigation regions}
\label{app:navresults}

Section~\ref{sec:tss_results} drives every ego at once in the four navigation
regions of Figure~\ref{fig:hycom-overview}. This appendix reports the matched
single-agent runs, in which one ego is under policy control and all other
traffic log-replays, so that the two can be read against each other: the
single-agent rows say how much of any failure belongs to the policy rather than
to the multi-agent setting. Both regimes use the same fixed 1{,}000-step
horizon of Appendix~\ref{app:termination} and the same 200\,m re-scoring, and
policies are trained separately per region.

\input{tables/main_solo}

\begin{figure}[!htbp]
\centering
\includegraphics[width=0.85\linewidth]{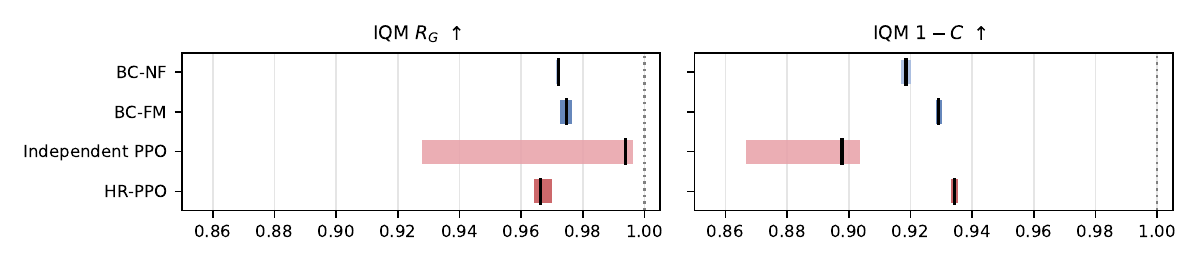}
\caption{\rev{\textbf{Aggregate over the four navigation regions} (4 regions $\times$ 3 seeds; single-agent; the checkpoints of Table~\ref{tab:region_main_solo}): interquartile mean (IQM) with 95\% stratified-bootstrap CIs, computed as in Appendix~\ref{app:agg}; dotted line: human replay. $\reachg$ is scored at 200\,m.}}
\label{fig:agg-nav-solo}
\end{figure}

\begin{figure}[!htbp]
\centering
\includegraphics[width=0.85\linewidth]{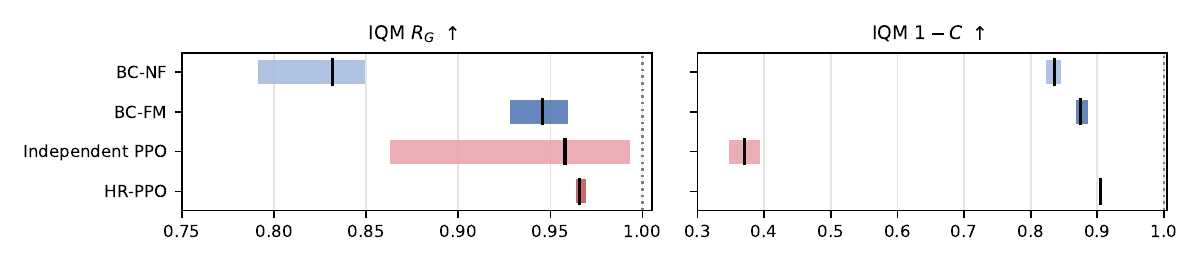}
\caption{\rev{\textbf{Aggregate over the four navigation regions, multi-agent} (4 regions $\times$ 3 seeds; every vessel under policy control; the checkpoints of Table~\ref{tab:region_main_fleet}): interquartile mean (IQM) with 95\% stratified-bootstrap CIs, computed as in Appendix~\ref{app:agg}; dotted line: human replay. $\reachg$ is scored at 200\,m.}}
\label{fig:agg-nav-fleet}
\end{figure}

\rev{Figures~\ref{fig:agg-nav-solo} and~\ref{fig:agg-nav-fleet} aggregate the single-agent and multi-agent rows (Tables~\ref{tab:region_main_solo} and~\ref{tab:region_main_fleet}) over the four regions and three seeds in the same way as the anchorage aggregate of Appendix~\ref{app:agg}.}

Two patterns are visible. First, imitation transfers to these regions: BC-NF
and BC-FM reach 90.4--99.8\% of goals, and the two intersections are easier
than the straits, where episodes are longer and goals further away (FDE
607--898\,m at the intersections against 2,017--2,423\,m in the straits).
Second, the reinforcement learning baselines are the weakest rows and fail
region by region rather than uniformly: Independent PPO reaches 93.5\% of goals
in TSS West but 13.5\% in Intersection 2, and its TSS West $\gcade$ of 899.43
shows it reaching goals by a route unlike the demonstrated one.

\input{tables/kinematics_solo}
\input{tables/kinematics_fleet}

Tables~\ref{tab:region_kin_solo} and~\ref{tab:region_kin_fleet} report the
kinematic results under the two regimes. BC-FM is the smoothest method in
almost every cell, and the reinforcement learning rows carry the largest
accelerations and drift angles, consistent with their tracking errors.
Curvature, acceleration and drift all reach larger values under joint control
for every method, which is expected when neighbours manoeuvre rather than
replay.

\input{figures/arrival_delay}

%% file: tables/main_solo.tex
\begin{table}[!htbp]
\centering\small
\setlength{\tabcolsep}{4pt}
\sbox{\resulttable}{%
\begin{tabular}{llcccccc}
\toprule
Region & Method & $\reachg\uparrow$ & Collision rate $C\downarrow$ & CPA (m) & $\gcade\downarrow$ & FDE$_{\neg R}$ (m) \\
\midrule
Intersection 1 & BC-NF & $0.9984\pm0.0001$ & $0.0742\pm0.0012$ & $269.9\pm1.1$ & $43.95\pm0.05$ & $852$ \\
Intersection 1 & BC-FM & $0.9967\pm0.0018$ & $0.0666\pm0.0009$ & $283.9\pm0.8$ & $46.61\pm0.31$ & $909$ \\
Intersection 1 & Independent PPO & $0.9325\pm0.1144$ & $0.1378\pm0.0539$ & $174.0\pm23.2$ & $572.04\pm0.92$ & $6083$ \\
Intersection 1 & HR-PPO & $0.9962\pm0.0010$ & $0.0729\pm0.0020$ & $280.4\pm2.9$ & $52.62\pm0.53$ & $1036$ \\
\midrule
Intersection 2 & BC-NF & $0.9975\pm0.0006$ & $0.0654\pm0.0020$ & $290.2\pm2.7$ & $37.07\pm0.25$ & $601$ \\
Intersection 2 & BC-FM & $0.9982\pm0.0006$ & $0.0544\pm0.0005$ & $312.0\pm0.7$ & $37.96\pm0.22$ & $619$ \\
Intersection 2 & Independent PPO & $0.6999\pm0.2739$ & $0.2804\pm0.1940$ & $97.8\pm96.9$ & $362.97\pm140.96$ & $3352$ \\
Intersection 2 & HR-PPO & $0.9963\pm0.0004$ & $0.0585\pm0.0009$ & $311.5\pm0.8$ & $45.39\pm0.72$ & $707$ \\
\midrule
TSS East & BC-NF & $0.9095\pm0.0029$ & $0.1206\pm0.0030$ & $313.6\pm7.8$ & $75.73\pm0.47$ & $2319$ \\
TSS East & BC-FM & $0.9456\pm0.0078$ & $0.1043\pm0.0021$ & $345.2\pm2.1$ & $79.08\pm0.63$ & $2457$ \\
TSS East & Independent PPO & $0.9912\pm0.0018$ & $0.0920\pm0.0079$ & $444.1\pm1.4$ & $912.39\pm2.92$ & $17004$ \\
TSS East & HR-PPO & $0.9367\pm0.0067$ & $0.1100\pm0.0010$ & $341.2\pm2.9$ & $84.98\pm2.01$ & $2607$ \\
\midrule
TSS West & BC-NF & $0.9465\pm0.0015$ & $0.0887\pm0.0032$ & $483.0\pm4.5$ & $72.06\pm0.80$ & $2053$ \\
TSS West & BC-FM & $0.9514\pm0.0047$ & $0.0752\pm0.0016$ & $503.5\pm3.8$ & $73.25\pm0.30$ & $2113$ \\
TSS West & Independent PPO & $0.9967\pm0.0008$ & $0.0493\pm0.0135$ & $729.0\pm82.1$ & $894.47\pm2.24$ & $16703$ \\
TSS West & HR-PPO & $0.4196\pm0.1274$ & $0.0345\pm0.0038$ & $755.6\pm50.6$ & $94.84\pm5.55$ & $2218$ \\
\bottomrule
\end{tabular}}
\ifdim\wd\resulttable>\linewidth\resizebox{\linewidth}{!}{\usebox{\resulttable}}\else\usebox{\resulttable}\fi
\caption{Four-method comparison on the four TSS and intersection regions, with one vessel under policy control and all other traffic replayed from the logs. $\reachg$ is scored at a goal radius $r_G=200$\,m. Each cell is the mean over seeds, with $\pm$ the standard deviation where all three seeds were evaluated; a dash marks a cell that has not been evaluated.}
\label{tab:region_main_solo}
\end{table}

%% file: tables/kinematics_solo.tex
\begin{table}[!htbp]
\centering\small
\setlength{\tabcolsep}{4pt}
\sbox{\resulttable}{%
\begin{tabular}{llrrrrrr}
\toprule
Region & Method & min $\tilde\kappa$ & max $\tilde\kappa$ & min $a$ & max $a$ & min drift & max drift \\
\midrule
Intersection 1 & BC-NF & $-0.0079\pm0.0001$ & $0.0082\pm0.0001$ & $-0.5000\pm0.0061$ & $0.4498\pm0.0050$ & $-0.1150\pm0.0050$ & $0.1297\pm0.0007$ \\
Intersection 1 & BC-FM & $-0.0041\pm0.0001$ & $0.0046\pm0.0000$ & $-0.2675\pm0.0006$ & $0.2395\pm0.0012$ & $-0.0961\pm0.0054$ & $0.1062\pm0.0043$ \\
Intersection 1 & Independent PPO & $-0.0070\pm0.0023$ & $0.0326\pm0.0271$ & $-0.1304\pm0.1166$ & $1.5542\pm0.0551$ & $-1.7279\pm0.3083$ & $1.2727\pm0.4292$ \\
Intersection 1 & HR-PPO & $-0.0023\pm0.0001$ & $0.0026\pm0.0001$ & $-0.0569\pm0.0019$ & $0.0767\pm0.0026$ & $-0.1128\pm0.0017$ & $0.1100\pm0.0145$ \\
\midrule
Intersection 2 & BC-NF & $-0.0056\pm0.0001$ & $0.0058\pm0.0002$ & $-0.5000\pm0.0023$ & $0.4411\pm0.0005$ & $-0.0861\pm0.0047$ & $0.1110\pm0.0064$ \\
Intersection 2 & BC-FM & $-0.0026\pm0.0001$ & $0.0031\pm0.0002$ & $-0.2159\pm0.0017$ & $0.1910\pm0.0013$ & $-0.0561\pm0.0015$ & $0.0746\pm0.0030$ \\
Intersection 2 & Independent PPO & $-0.0171\pm0.0140$ & $0.0761\pm0.1022$ & $-0.4865\pm0.3944$ & $1.6399\pm0.0653$ & $-1.9463\pm0.4360$ & $2.1608\pm0.6974$ \\
Intersection 2 & HR-PPO & $-0.0013\pm0.0001$ & $0.0013\pm0.0001$ & $-0.0583\pm0.0028$ & $0.0819\pm0.0024$ & $-0.0601\pm0.0039$ & $0.0615\pm0.0054$ \\
\midrule
TSS East & BC-NF & $-0.0057\pm0.0001$ & $0.0062\pm0.0005$ & $-0.7632\pm0.0161$ & $0.6450\pm0.0217$ & $-0.1589\pm0.0408$ & $0.1121\pm0.0053$ \\
TSS East & BC-FM & $-0.0019\pm0.0001$ & $0.0021\pm0.0001$ & $-0.2745\pm0.0029$ & $0.2444\pm0.0028$ & $-0.0844\pm0.0097$ & $0.0718\pm0.0003$ \\
TSS East & Independent PPO & $-0.0244\pm0.0048$ & $0.0120\pm0.0082$ & $-0.1093\pm0.0203$ & $1.5097\pm0.0014$ & $-2.5067\pm0.6779$ & $2.6987\pm0.3925$ \\
TSS East & HR-PPO & $-0.0013\pm0.0005$ & $0.0014\pm0.0002$ & $-0.0563\pm0.0004$ & $0.0860\pm0.0005$ & $-0.1094\pm0.0023$ & $0.0588\pm0.0112$ \\
\midrule
TSS West & BC-NF & $-0.0031\pm0.0001$ & $0.0076\pm0.0001$ & $-0.8140\pm0.0281$ & $0.7218\pm0.0248$ & $-0.1182\pm0.0008$ & $0.0832\pm0.0032$ \\
TSS West & BC-FM & $-0.0009\pm0.0002$ & $0.0053\pm0.0010$ & $-0.3223\pm0.0185$ & $0.2785\pm0.0150$ & $-0.0985\pm0.0131$ & $0.0527\pm0.0047$ \\
TSS West & Independent PPO & $-0.0064\pm0.0074$ & $0.0175\pm0.0089$ & $-0.1725\pm0.0489$ & $1.4932\pm0.0066$ & $-3.0202\pm0.0401$ & $3.0037\pm0.0732$ \\
TSS West & HR-PPO & $-0.0455\pm0.0077$ & $0.0140\pm0.0013$ & $-0.0922\pm0.0104$ & $0.1115\pm0.0099$ & $-0.2964\pm0.0873$ & $0.2503\pm0.0422$ \\
\bottomrule
\end{tabular}}
\ifdim\wd\resulttable>\linewidth\resizebox{\linewidth}{!}{\usebox{\resulttable}}\else\usebox{\resulttable}\fi
\caption{Smallest and largest curvature, acceleration and drift reached during an episode, with one vessel under policy control and all other traffic replayed from the logs. Each cell averages over episodes, then over seeds; a dash marks a cell that has not been evaluated.}
\label{tab:region_kin_solo}
\end{table}

%% file: tables/kinematics_fleet.tex
\begin{table}[!htbp]
\centering\small
\setlength{\tabcolsep}{4pt}
\sbox{\resulttable}{%
\begin{tabular}{llrrrrrr}
\toprule
Region & Method & min $\tilde\kappa$ & max $\tilde\kappa$ & min $a$ & max $a$ & min drift & max drift \\
\midrule
Intersection 1 & BC-NF & $-0.0445\pm0.0052$ & $0.0611\pm0.0088$ & $-0.9731\pm0.0464$ & $0.8198\pm0.0335$ & $-0.9462\pm0.0589$ & $1.0831\pm0.0957$ \\
Intersection 1 & BC-FM & $-0.0196\pm0.0093$ & $0.0342\pm0.0250$ & $-0.4109\pm0.0125$ & $0.2443\pm0.0027$ & $-0.9511\pm0.1289$ & $0.8168\pm0.0409$ \\
Intersection 1 & Independent PPO & $-0.0235\pm0.0134$ & $0.2130\pm0.1449$ & $-1.7383\pm0.1230$ & $1.7489\pm0.1290$ & $-2.8694\pm0.2526$ & $2.8628\pm0.2823$ \\
Intersection 1 & HR-PPO & $-0.0032\pm0.0005$ & $0.0028\pm0.0001$ & $-0.2641\pm0.0049$ & $0.0714\pm0.0014$ & $-0.1210\pm0.0042$ & $0.1254\pm0.0135$ \\
\midrule
Intersection 2 & BC-NF & $-0.0583\pm0.0108$ & $0.0634\pm0.0101$ & $-1.2678\pm0.2012$ & $1.0525\pm0.1628$ & $-1.0948\pm0.0168$ & $1.1418\pm0.0351$ \\
Intersection 2 & BC-FM & $-0.0271\pm0.0137$ & $0.0941\pm0.0128$ & $-0.3757\pm0.0074$ & $0.2027\pm0.0056$ & $-0.6663\pm0.0809$ & $1.1049\pm0.1136$ \\
Intersection 2 & Independent PPO & $-0.0642\pm0.0335$ & $0.0806\pm0.0940$ & $-1.5510\pm0.3307$ & $1.7143\pm0.0918$ & $-3.0145\pm0.0207$ & $3.0026\pm0.0672$ \\
Intersection 2 & HR-PPO & $-0.0047\pm0.0020$ & $0.0031\pm0.0012$ & $-0.1883\pm0.0110$ & $0.0832\pm0.0036$ & $-0.1044\pm0.0557$ & $0.2106\pm0.1160$ \\
\midrule
TSS East & BC-NF & $-0.0296\pm0.0050$ & $0.0462\pm0.0130$ & $-1.2675\pm0.0205$ & $1.0861\pm0.0233$ & $-1.0671\pm0.1083$ & $1.1866\pm0.1353$ \\
TSS East & BC-FM & $-0.0123\pm0.0004$ & $0.0231\pm0.0139$ & $-0.5099\pm0.0115$ & $0.2696\pm0.0026$ & $-0.9031\pm0.1376$ & $0.5572\pm0.2056$ \\
TSS East & Independent PPO & $-0.1938\pm0.2005$ & $0.0442\pm0.0463$ & $-1.7619\pm0.0647$ & $1.6548\pm0.0370$ & $-2.8474\pm0.3188$ & $2.9369\pm0.1939$ \\
TSS East & HR-PPO & $-0.0013\pm0.0006$ & $0.0013\pm0.0003$ & $-0.3992\pm0.0067$ & $0.0849\pm0.0015$ & $-0.1147\pm0.0033$ & $0.0618\pm0.0104$ \\
\midrule
TSS West & BC-NF & $-0.0166\pm0.0075$ & $0.0236\pm0.0035$ & $-1.3468\pm0.0160$ & $1.1532\pm0.0380$ & $-0.5837\pm0.1109$ & $1.0453\pm0.2205$ \\
TSS West & BC-FM & $-0.0083\pm0.0066$ & $0.0077\pm0.0006$ & $-0.4430\pm0.0237$ & $0.3131\pm0.0027$ & $-0.4533\pm0.3729$ & $0.8552\pm0.5272$ \\
TSS West & Independent PPO & $-0.0111\pm0.0054$ & $0.0314\pm0.0145$ & $-1.5944\pm0.1408$ & $1.5012\pm0.0118$ & $-3.0516\pm0.0280$ & $3.0342\pm0.0573$ \\
TSS West & HR-PPO & $-0.0435\pm0.0075$ & $0.0149\pm0.0010$ & $-0.2520\pm0.0375$ & $0.1112\pm0.0090$ & $-0.3052\pm0.0813$ & $0.2557\pm0.0426$ \\
\bottomrule
\end{tabular}}
\ifdim\wd\resulttable>\linewidth\resizebox{\linewidth}{!}{\usebox{\resulttable}}\else\usebox{\resulttable}\fi
\caption{Smallest and largest curvature, acceleration and drift reached during an episode, with every vessel under policy control at the same time. Each cell averages over episodes, then over seeds; a dash marks a cell that has not been evaluated.}
\label{tab:region_kin_fleet}
\end{table}

%% file: figures/arrival_delay.tex
\begin{figure}[p]
\centering
{\small\itshape Solo}\\[0.3em]
\includegraphics[width=0.48\textwidth]{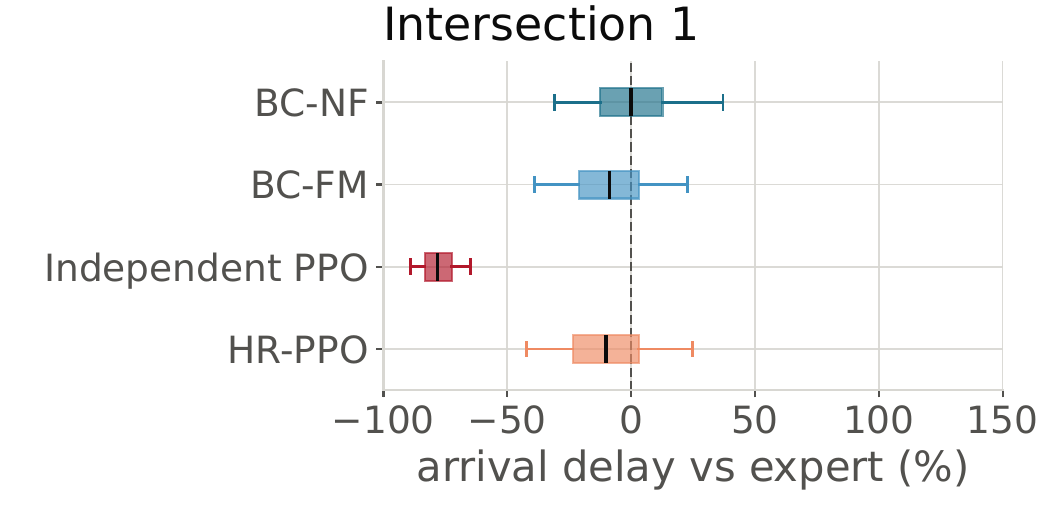}\hfill
\includegraphics[width=0.48\textwidth]{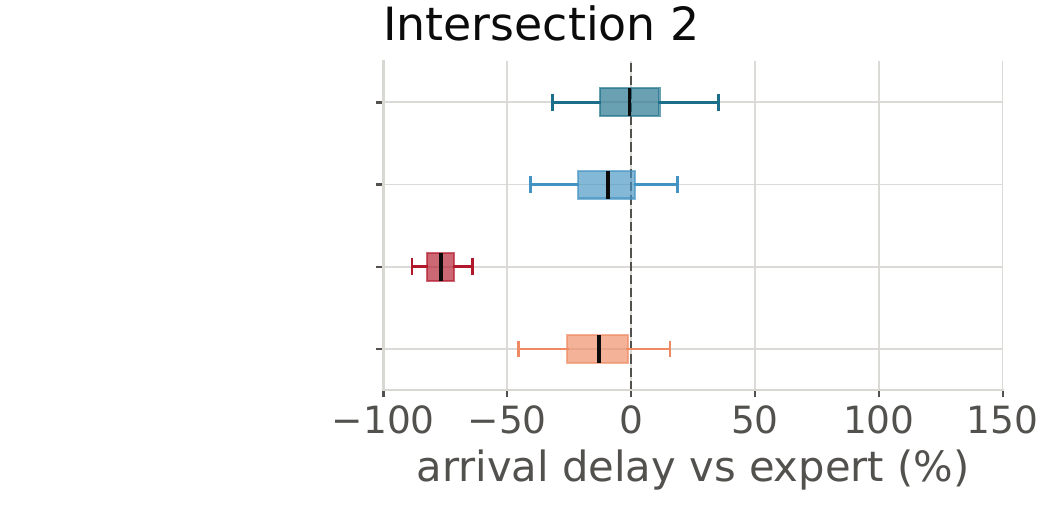}\\[0.4em]
\includegraphics[width=0.48\textwidth]{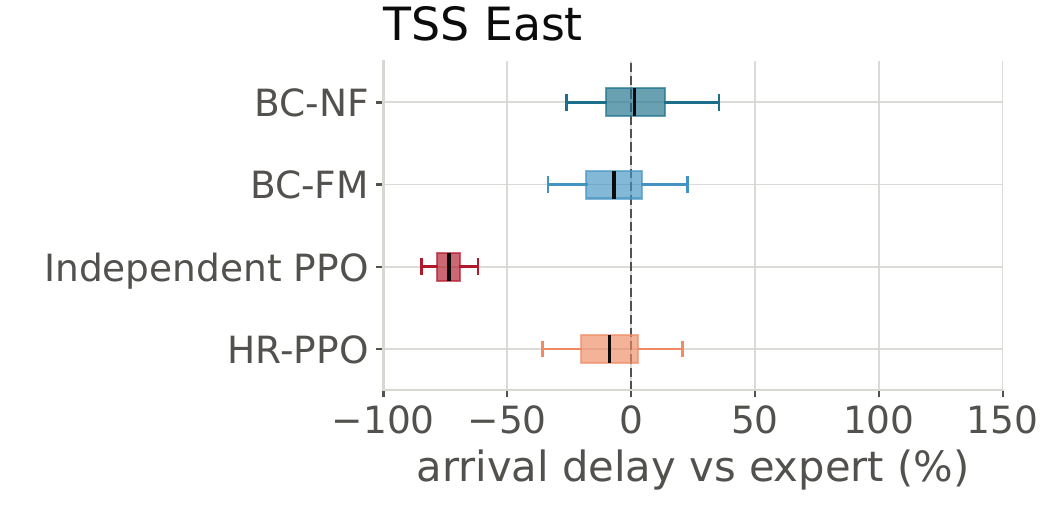}\hfill
\includegraphics[width=0.48\textwidth]{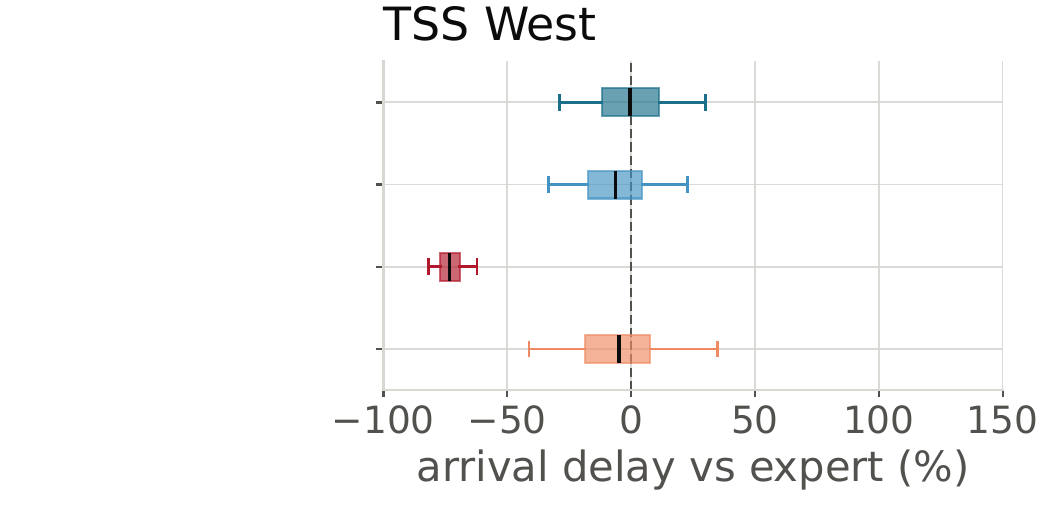}

\vspace{1.1em}
{\small\itshape Fleet}\\[0.3em]
\includegraphics[width=0.48\textwidth]{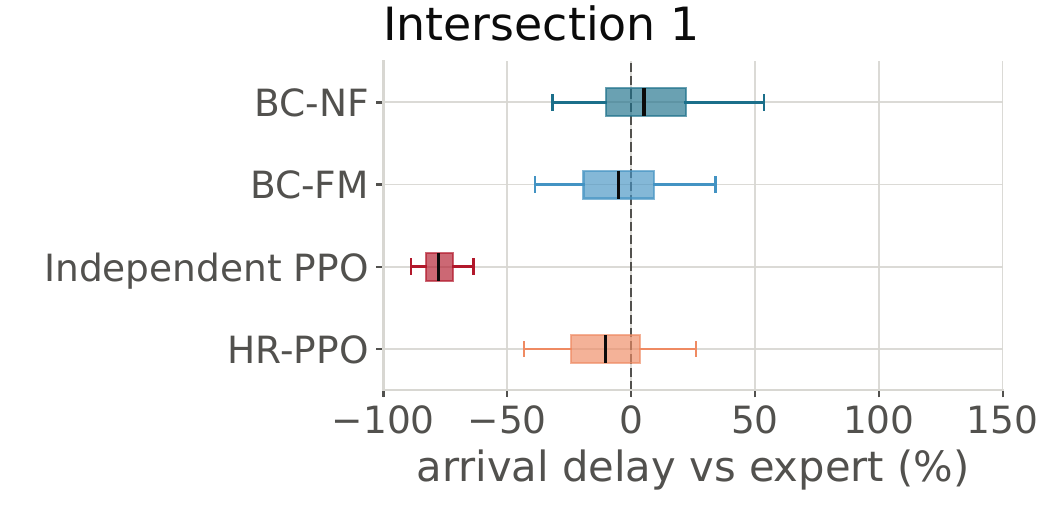}\hfill
\includegraphics[width=0.48\textwidth]{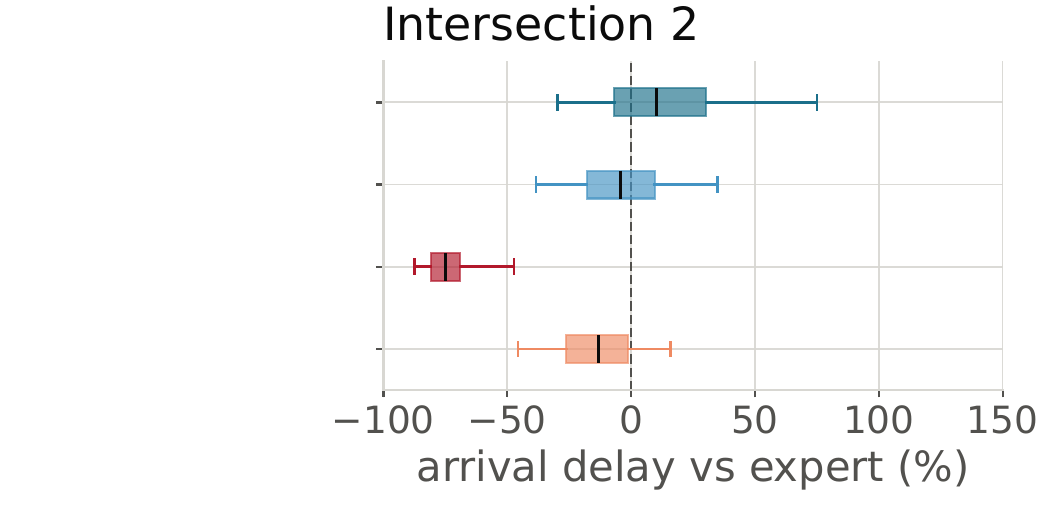}\\[0.4em]
\includegraphics[width=0.48\textwidth]{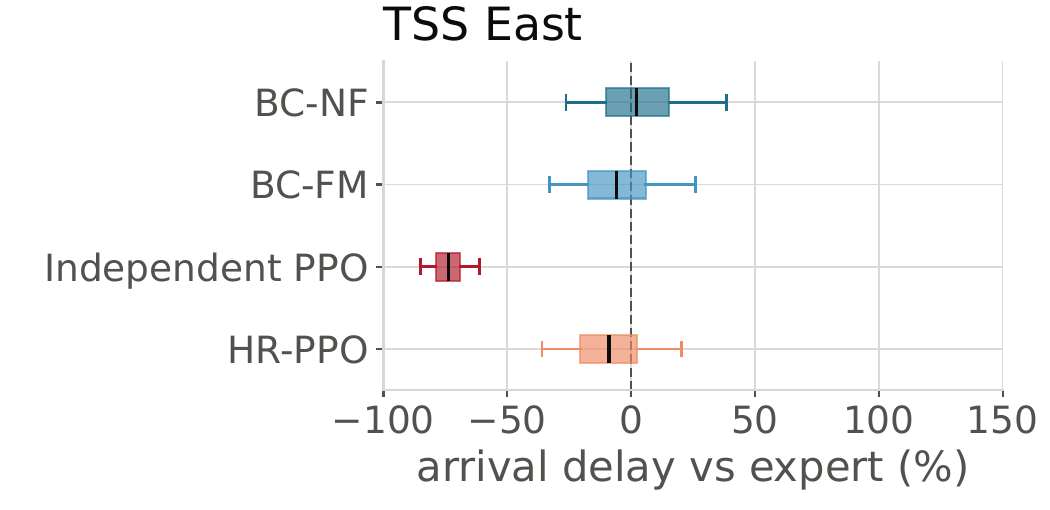}\hfill
\includegraphics[width=0.48\textwidth]{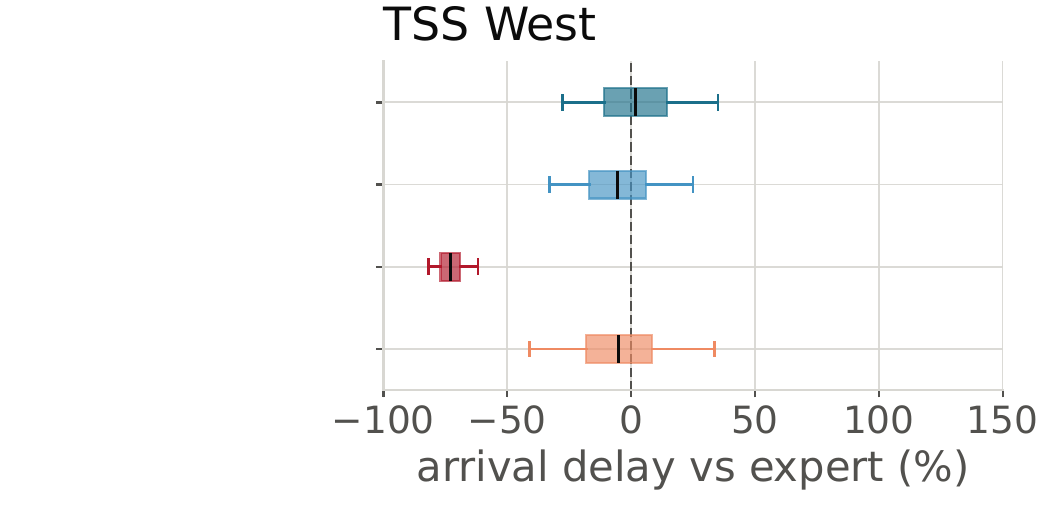}

\caption{Arrival delay against the expert, over episodes, for the four learned
methods in each navigation region under both control regimes. Negative values
are earlier than the recorded transit and positive values later; the dashed
line marks the expert, and each box shows the interquartile range with the
median.}
\label{fig:arrival-delay}
\end{figure}

%% file: sections/12_rebuttal_appendix.tex
\FloatBarrier
\section{Validation-style checkpoint selection}
\label{app:sel}
{\color{blue}
The results in Section~\ref{sec:anchorage_results} are \emph{test-selected}: each run's checkpoint was chosen on the test set, and PAR's settings were fixed on test runs (Appendix~\ref{app:envelope}). \rev{This appendix repeats the comparison so that no selection uses the reporting months. FM-BC, NF-BC and PPO use their existing runs with the checkpoint re-selected. QC-FQL, ReinFlow and HR-PPO are retrained with the October-selected FM-BC, DEAS-BoN is re-evaluated with it, and PAR's setting is re-selected on October. Appendix~\ref{app:sel_tp} separately reports retrained GLRP and TrAISformer models.}

\paragraph{Recommended protocol.} We recommend that benchmark users train on January--September (273 days), select checkpoints and hyper-parameters on October (31 days), and report on November--December (61 days).

\paragraph{Setup.} All policies were trained on January--September. \rev{For the six methods with checkpoint schedules (FM-BC, NF-BC, QC-FQL, ReinFlow, HR-PPO, PPO; 72 runs), each of the last 10 checkpoints} had already been evaluated on all 92 test days. We split these stored per-episode results by date. We use October only to select the checkpoint, with the paper's rule (highest $\reachg$, ties broken by $\gcade$), and November--December only to report. The re-analysis first reproduces all 40 reported region $\times$ method cells exactly, and the rule re-selects the paper's checkpoint in all 72 learned-method runs (6 methods $\times$ 4 regions $\times$ 3 seeds).

\paragraph{Methods that depend on FM-BC.} \rev{QC-FQL distills from a frozen FM-BC policy, ReinFlow starts from FM-BC and is regularized toward it, HR-PPO is penalized toward it, and DEAS-BoN samples its candidates from it (Appendix~\ref{app:methods}). In the runs of Section~\ref{sec:anchorage_results}, this FM-BC checkpoint was test-selected. October selection picks the same FM-BC checkpoint in 4 of the 12 region $\times$ seed runs. For the other 8, we retrained QC-FQL, ReinFlow and HR-PPO with the October-selected FM-BC checkpoint, with the same seeds and training settings, and evaluated their last 10 checkpoints as before. We re-evaluated DEAS-BoN with the October-selected FM-BC checkpoint and unchanged critics.}

\paragraph{October protocol versus test selection.} \rev{Table~\ref{tab:sel_oct} scores the October protocol and the paper's test-selected results on the same November--December episodes. Over the 28 method $\times$ region cells, the test-selected results are higher in $\reachg$ by 0.19 percentage points (pp) on average (median 0.13\,pp, range $-0.17$ to $+1.15$\,pp; the largest difference is ReinFlow on AESPA). With the stored runs of the six methods with checkpoint schedules, the checkpoint choice alone raised $\reachg$ by 0.13\,pp on average (maximum 0.32\,pp). Excluding PPO, whose seeds diverge, the collision rate differs by at most 1.58\,pp and $\gcade$ by at most 0.76\,m (both for ReinFlow on AEPBC). For PPO, the differences are at most 0.66\,pp and 6.1\,m. With both scored on November--December, the ranking of two methods changes only where their gap is smaller than the seed standard deviation: one $\reachg$ pair (FM-BC versus FM-BC + PAR on AEW, gaps below 0.2\,pp) and four collision-rate pairs. In this ranking, FM-BC + PAR uses C3 at the October-selected FM-BC checkpoints (Table~\ref{tab:sel_par}).}

\paragraph{Fixed final checkpoint.} Table~\ref{tab:sel_final} applies no selection at all and uses each run's last checkpoint on the full 92-day test set. FM-BC and NF-BC lose 0.1--0.3\,pp of $\reachg$, and HR-PPO and PPO lose at most \rev{0.4\,pp}. In contrast, ReinFlow loses \rev{11--22\,pp on AEW, AESPA and AEPBC, where 7 of its 9 runs degrade late in training}, and QC-FQL loses 3.7\,pp on AWW\rev{, where all three runs degrade late}. \rev{These losses are consistent with late-training degradation, and October selection also avoids the degraded checkpoints}.

\paragraph{PAR setting re-selected on October.} \rev{We re-selected PAR's setting on October data. We fixed the rule before the runs: on the October episodes of AEW and AESPA, with FM-BC seed~0 at its October-selected checkpoint, we take from the 11 settings of Table~\ref{tab:envelopes} the one with the lowest collision rate averaged over the two regions, among the settings whose $\reachg$ is at most 0.5\,pp below FM-BC's in both regions. Nine settings pass this condition. The rule selects E0, the setting without turn or speed-change limits (mean October collision rate 0.58\%), and C3 ranks last of the nine (2.06\%). For E0, goal-rule radii of 200, 400 and 800\,m give the same $\reachg$ and the same collisions on the AEW October episodes, so we keep 400\,m. We then evaluated FM-BC, FM-BC + PAR (E0) and FM-BC + PAR (C3) at the October-selected FM-BC checkpoints of all three seeds and scored them on November--December (Table~\ref{tab:sel_par}). The three are scored on identical episodes, so their differences are paired. The collision rate falls from $0.2045/0.4582/0.2768/0.2762$ for FM-BC to $0.0026/0.0089/0.0061/0.0043$ with E0 and to $0.0120/0.0328/0.0202/0.0244$ with C3 (AEW/AESPA/AEPBC/AWW). E0 removes 97.8--98.7\% of FM-BC's collisions and C3 removes 91.2--94.1\%. $\reachg$ changes by $+1.0/+1.4/+2.6/+7.4$\,pp with E0 and by $-0.1/-0.5/+1.1/+3.6$\,pp with C3. Relative to the test-selected FM-BC checkpoints, the paired change in $\reachg$ moves by at most 0.43\,pp for E0 and 0.28\,pp for C3.}

\begin{table}[!htbp]\color{blue}
\centering\scriptsize
\setlength{\tabcolsep}{3pt}
\resizebox{\linewidth}{!}{%
\begin{tabular}{llccccccc}
\toprule
& & \multicolumn{3}{c}{$\reachg\uparrow$} & \multicolumn{2}{c}{$C\downarrow$} & \multicolumn{2}{c}{$\gcade\downarrow$ (m)} \\
\cmidrule(lr){3-5}\cmidrule(lr){6-7}\cmidrule(lr){8-9}
Region & Method & October protocol & test-selected & difference & October protocol & test-selected & October protocol & test-selected \\
\midrule
AEW & FM-BC & $0.8955\pm0.0064$ & $0.8969\pm0.0056$ & $+0.0014$ & $0.2045\pm0.0055$ & $0.2028\pm0.0071$ & $23.08\pm0.10$ & $23.10\pm0.09$ \\
AEW & NF-BC & $0.7541\pm0.0015$ & $0.7569\pm0.0009$ & $+0.0028$ & $0.2267\pm0.0040$ & $0.2250\pm0.0046$ & $26.42\pm0.21$ & $26.42\pm0.19$ \\
AEW & ReinFlow & $0.9206\pm0.0128$ & $0.9266\pm0.0079$ & $+0.0060$ & $0.2224\pm0.0105$ & $0.2154\pm0.0143$ & $25.48\pm0.44$ & $25.45\pm0.61$ \\
AEW & HR-PPO & $0.8804\pm0.0142$ & $0.8823\pm0.0103$ & $+0.0019$ & $0.2241\pm0.0145$ & $0.2230\pm0.0092$ & $29.94\pm0.99$ & $30.01\pm0.84$ \\
AEW & PPO & $0.3365\pm0.4782$ & $0.3365\pm0.4782$ & $+0.0000$ & $0.4465\pm0.1785$ & $0.4465\pm0.1785$ & $242.38\pm51.84$ & $242.38\pm51.84$ \\
AEW & QC-FQL & $0.9181\pm0.0039$ & $0.9194\pm0.0013$ & $+0.0012$ & $0.2005\pm0.0041$ & $0.2019\pm0.0001$ & $23.48\pm0.21$ & $23.51\pm0.18$ \\
AEW & DEAS-BoN & $0.9673\pm0.0039$ & $0.9666\pm0.0044$ & $-0.0007$ & $0.2043\pm0.0024$ & $0.2045\pm0.0027$ & $31.81\pm0.11$ & $31.84\pm0.10$ \\
\midrule
AESPA & FM-BC & $0.9303\pm0.0008$ & $0.9310\pm0.0004$ & $+0.0007$ & $0.4492\pm0.0025$ & $0.4511\pm0.0038$ & $26.77\pm0.04$ & $26.76\pm0.06$ \\
AESPA & NF-BC & $0.7987\pm0.0063$ & $0.8005\pm0.0074$ & $+0.0018$ & $0.4926\pm0.0106$ & $0.4911\pm0.0081$ & $29.56\pm0.10$ & $29.59\pm0.07$ \\
AESPA & ReinFlow & $0.9372\pm0.0168$ & $0.9488\pm0.0201$ & $+0.0115$ & $0.4765\pm0.0254$ & $0.4641\pm0.0237$ & $29.69\pm0.41$ & $29.90\pm0.38$ \\
AESPA & HR-PPO & $0.9507\pm0.0004$ & $0.9491\pm0.0022$ & $-0.0017$ & $0.4536\pm0.0053$ & $0.4536\pm0.0035$ & $32.24\pm0.24$ & $32.33\pm0.13$ \\
AESPA & PPO & $0.0070\pm0.0038$ & $0.0079\pm0.0023$ & $+0.0009$ & $0.4782\pm0.1310$ & $0.4848\pm0.1305$ & $190.03\pm7.07$ & $184.21\pm13.90$ \\
AESPA & QC-FQL & $0.9592\pm0.0013$ & $0.9615\pm0.0024$ & $+0.0023$ & $0.4487\pm0.0071$ & $0.4472\pm0.0010$ & $27.81\pm0.09$ & $27.83\pm0.10$ \\
AESPA & DEAS-BoN & $0.9788\pm0.0023$ & $0.9787\pm0.0022$ & $-0.0001$ & $0.4522\pm0.0023$ & $0.4538\pm0.0037$ & $36.57\pm0.19$ & $36.55\pm0.20$ \\
\midrule
AEPBC & FM-BC & $0.8986\pm0.0066$ & $0.9002\pm0.0070$ & $+0.0016$ & $0.2782\pm0.0032$ & $0.2753\pm0.0010$ & $24.57\pm0.09$ & $24.56\pm0.07$ \\
AEPBC & NF-BC & $0.7669\pm0.0037$ & $0.7684\pm0.0016$ & $+0.0015$ & $0.3151\pm0.0070$ & $0.3137\pm0.0087$ & $26.96\pm0.32$ & $26.96\pm0.32$ \\
AEPBC & ReinFlow & $0.9196\pm0.0081$ & $0.9287\pm0.0175$ & $+0.0091$ & $0.2958\pm0.0052$ & $0.2800\pm0.0092$ & $27.52\pm0.30$ & $26.76\pm0.42$ \\
AEPBC & HR-PPO & $0.9140\pm0.0061$ & $0.9140\pm0.0041$ & $+0.0000$ & $0.2987\pm0.0096$ & $0.3015\pm0.0114$ & $30.46\pm0.12$ & $30.55\pm0.45$ \\
AEPBC & PPO & $0.6685\pm0.3043$ & $0.6697\pm0.3039$ & $+0.0012$ & $0.3291\pm0.0499$ & $0.3310\pm0.0495$ & $247.14\pm59.73$ & $247.42\pm59.59$ \\
AEPBC & QC-FQL & $0.9404\pm0.0046$ & $0.9437\pm0.0020$ & $+0.0034$ & $0.2716\pm0.0073$ & $0.2744\pm0.0002$ & $25.05\pm0.04$ & $25.02\pm0.04$ \\
AEPBC & DEAS-BoN & $0.9537\pm0.0032$ & $0.9544\pm0.0030$ & $+0.0007$ & $0.2742\pm0.0055$ & $0.2751\pm0.0064$ & $29.39\pm0.14$ & $29.39\pm0.13$ \\
\midrule
AWW & FM-BC & $0.7784\pm0.0125$ & $0.7807\pm0.0140$ & $+0.0024$ & $0.2777\pm0.0086$ & $0.2803\pm0.0105$ & $27.23\pm0.12$ & $27.26\pm0.14$ \\
AWW & NF-BC & $0.7126\pm0.0024$ & $0.7137\pm0.0020$ & $+0.0011$ & $0.2777\pm0.0024$ & $0.2766\pm0.0024$ & $27.18\pm0.13$ & $27.18\pm0.14$ \\
AWW & ReinFlow & $0.8249\pm0.0199$ & $0.8232\pm0.0164$ & $-0.0017$ & $0.2764\pm0.0049$ & $0.2786\pm0.0132$ & $27.89\pm0.28$ & $28.47\pm0.26$ \\
AWW & HR-PPO & $0.8499\pm0.0272$ & $0.8519\pm0.0331$ & $+0.0019$ & $0.2978\pm0.0077$ & $0.2937\pm0.0149$ & $31.85\pm0.17$ & $32.01\pm0.51$ \\
AWW & PPO & $0.9832\pm0.0208$ & $0.9838\pm0.0214$ & $+0.0006$ & $0.2893\pm0.0314$ & $0.2928\pm0.0383$ & $298.29\pm17.10$ & $292.16\pm26.83$ \\
AWW & QC-FQL & $0.8549\pm0.0037$ & $0.8607\pm0.0036$ & $+0.0058$ & $0.2622\pm0.0033$ & $0.2615\pm0.0036$ & $26.47\pm0.14$ & $26.47\pm0.23$ \\
AWW & DEAS-BoN & $0.8361\pm0.0065$ & $0.8346\pm0.0086$ & $-0.0015$ & $0.2723\pm0.0013$ & $0.2736\pm0.0039$ & $28.39\pm0.18$ & $28.41\pm0.26$ \\
\bottomrule
\end{tabular}}
\caption{\rev{November--December (61 days) results under the October protocol and at the paper's test-selected checkpoints; mean $\pm$ standard deviation over 3 seeds. FM-BC, NF-BC and PPO: the same runs with the checkpoint selected on October. QC-FQL, ReinFlow and HR-PPO: the FM-BC checkpoint and the method's own checkpoint are both selected on October (8 of the 12 runs retrained). DEAS-BoN: re-evaluated with the October-selected FM-BC checkpoint. ``Difference'' is test-selected minus October-protocol $\reachg$.}}
\label{tab:sel_oct}
\end{table}

\begin{table}[!htbp]\color{blue}
\centering\scriptsize
\setlength{\tabcolsep}{3pt}
\resizebox{\linewidth}{!}{%
\begin{tabular}{llccccccc}
\toprule
& & \multicolumn{3}{c}{$\reachg\uparrow$} & \multicolumn{2}{c}{$C\downarrow$} & \multicolumn{2}{c}{$\gcade\downarrow$ (m)} \\
\cmidrule(lr){3-5}\cmidrule(lr){6-7}\cmidrule(lr){8-9}
Region & Method & test-selected & final & difference & test-selected & final & test-selected & final \\
\midrule
AEW & FM-BC & $0.8944\pm0.0057$ & $0.8933\pm0.0065$ & $+0.0011$ & $0.2036\pm0.0036$ & $0.2033\pm0.0027$ & $23.18\pm0.08$ & $23.18\pm0.10$ \\
AEW & NF-BC & $0.7545\pm0.0003$ & $0.7521\pm0.0007$ & $+0.0024$ & $0.2231\pm0.0028$ & $0.2237\pm0.0029$ & $26.54\pm0.13$ & $26.54\pm0.13$ \\
AEW & ReinFlow & $0.9246\pm0.0059$ & $0.8147\pm0.1025$ & $+0.1100$ & $0.2132\pm0.0095$ & $0.3777\pm0.1481$ & $25.48\pm0.62$ & $32.50\pm6.46$ \\
AEW & HR-PPO & $0.8796\pm0.0094$ & $0.8760\pm0.0148$ & $+0.0036$ & $0.2200\pm0.0080$ & $0.2229\pm0.0131$ & $30.25\pm0.86$ & $30.07\pm0.93$ \\
AEW & PPO & $0.3382\pm0.4746$ & $0.3378\pm0.4746$ & $+0.0004$ & $0.4329\pm0.1714$ & $0.4303\pm0.1696$ & $242.01\pm56.28$ & $242.42\pm56.32$ \\
AEW & QC-FQL & $0.9199\pm0.0013$ & $0.9106\pm0.0061$ & $+0.0093$ & $0.2025\pm0.0019$ & $0.2033\pm0.0031$ & $23.57\pm0.15$ & $23.60\pm0.14$ \\
\midrule
AESPA & FM-BC & $0.9313\pm0.0004$ & $0.9287\pm0.0024$ & $+0.0026$ & $0.4510\pm0.0016$ & $0.4492\pm0.0008$ & $26.76\pm0.07$ & $26.72\pm0.06$ \\
AESPA & NF-BC & $0.7961\pm0.0036$ & $0.7938\pm0.0033$ & $+0.0024$ & $0.4897\pm0.0041$ & $0.4888\pm0.0024$ & $29.52\pm0.02$ & $29.52\pm0.06$ \\
AESPA & ReinFlow & $0.9482\pm0.0201$ & $0.8003\pm0.1374$ & $+0.1479$ & $0.4638\pm0.0228$ & $0.6238\pm0.1528$ & $30.00\pm0.34$ & $37.61\pm6.86$ \\
AESPA & HR-PPO & $0.9466\pm0.0015$ & $0.9480\pm0.0014$ & $-0.0014$ & $0.4576\pm0.0041$ & $0.4587\pm0.0045$ & $32.78\pm0.12$ & $32.42\pm0.56$ \\
AESPA & PPO & $0.0075\pm0.0034$ & $0.0039\pm0.0004$ & $+0.0035$ & $0.4883\pm0.1432$ & $0.3818\pm0.0751$ & $185.71\pm13.65$ & $187.58\pm11.54$ \\
AESPA & QC-FQL & $0.9616\pm0.0014$ & $0.9593\pm0.0025$ & $+0.0023$ & $0.4461\pm0.0020$ & $0.4474\pm0.0016$ & $27.85\pm0.09$ & $27.89\pm0.05$ \\
\midrule
AEPBC & FM-BC & $0.8977\pm0.0064$ & $0.8963\pm0.0058$ & $+0.0013$ & $0.2779\pm0.0018$ & $0.2769\pm0.0016$ & $24.16\pm0.10$ & $24.16\pm0.09$ \\
AEPBC & NF-BC & $0.7621\pm0.0031$ & $0.7591\pm0.0028$ & $+0.0030$ & $0.3163\pm0.0106$ & $0.3178\pm0.0091$ & $26.92\pm0.23$ & $26.90\pm0.23$ \\
AEPBC & ReinFlow & $0.9256\pm0.0182$ & $0.7084\pm0.0125$ & $+0.2172$ & $0.2845\pm0.0078$ & $0.5191\pm0.0044$ & $26.30\pm0.58$ & $36.95\pm0.90$ \\
AEPBC & HR-PPO & $0.9049\pm0.0049$ & $0.9048\pm0.0072$ & $+0.0001$ & $0.3070\pm0.0097$ & $0.3024\pm0.0050$ & $30.56\pm0.50$ & $30.48\pm0.25$ \\
AEPBC & PPO & $0.6672\pm0.3032$ & $0.6669\pm0.3032$ & $+0.0003$ & $0.3362\pm0.0546$ & $0.3350\pm0.0547$ & $246.84\pm59.39$ & $246.79\pm59.32$ \\
AEPBC & QC-FQL & $0.9428\pm0.0021$ & $0.9399\pm0.0028$ & $+0.0029$ & $0.2746\pm0.0022$ & $0.2723\pm0.0049$ & $24.62\pm0.05$ & $24.62\pm0.06$ \\
\midrule
AWW & FM-BC & $0.7774\pm0.0123$ & $0.7745\pm0.0105$ & $+0.0029$ & $0.2539\pm0.0061$ & $0.2560\pm0.0059$ & $27.27\pm0.10$ & $27.26\pm0.10$ \\
AWW & NF-BC & $0.6989\pm0.0014$ & $0.6970\pm0.0008$ & $+0.0019$ & $0.2569\pm0.0059$ & $0.2571\pm0.0074$ & $27.56\pm0.06$ & $27.55\pm0.12$ \\
AWW & ReinFlow & $0.8219\pm0.0145$ & $0.8198\pm0.0200$ & $+0.0022$ & $0.2547\pm0.0133$ & $0.2510\pm0.0085$ & $28.40\pm0.03$ & $27.88\pm0.41$ \\
AWW & HR-PPO & $0.8370\pm0.0371$ & $0.8423\pm0.0261$ & $-0.0053$ & $0.2680\pm0.0120$ & $0.2722\pm0.0082$ & $32.04\pm0.51$ & $31.86\pm0.49$ \\
AWW & PPO & $0.9849\pm0.0195$ & $0.9842\pm0.0189$ & $+0.0007$ & $0.2691\pm0.0349$ & $0.2637\pm0.0347$ & $293.18\pm26.13$ & $299.15\pm16.69$ \\
AWW & QC-FQL & $0.8583\pm0.0031$ & $0.8217\pm0.0094$ & $+0.0367$ & $0.2399\pm0.0033$ & $0.2494\pm0.0037$ & $26.39\pm0.18$ & $26.22\pm0.25$ \\
\bottomrule
\end{tabular}}
\caption{\rev{Full 92-day test set: test-selected (paper) checkpoint versus each run's final checkpoint; mean $\pm$ standard deviation over 3 seeds. ``Difference'' is selected minus final $\reachg$.}}
\label{tab:sel_final}
\end{table}

\begin{table}[!htbp]\color{blue}
\centering\scriptsize
\setlength{\tabcolsep}{3pt}
\resizebox{\linewidth}{!}{%
\begin{tabular}{lrccccccccc}
\toprule
& & \multicolumn{3}{c}{$\reachg\uparrow$} & \multicolumn{3}{c}{$C\downarrow$} & \multicolumn{3}{c}{$\gcade$ (m)} \\
\cmidrule(lr){3-5}\cmidrule(lr){6-8}\cmidrule(lr){9-11}
Region & Episodes & FM-BC & + PAR (E0) & + PAR (C3) & FM-BC & + PAR (E0) & + PAR (C3) & FM-BC & + PAR (E0) & + PAR (C3) \\
\midrule
AEW & 5080 & $0.8976\pm0.0080$ & $0.9075\pm0.0070$ & $0.8970\pm0.0047$ & $0.2045\pm0.0065$ & $0.0026\pm0.0005$ & $0.0120\pm0.0014$ & $22.99\pm0.11$ & $23.10\pm0.10$ & $22.93\pm0.13$ \\
AESPA & 4791 & $0.9296\pm0.0028$ & $0.9439\pm0.0006$ & $0.9244\pm0.0066$ & $0.4582\pm0.0041$ & $0.0089\pm0.0012$ & $0.0328\pm0.0006$ & $26.86\pm0.02$ & $27.23\pm0.03$ & $27.06\pm0.02$ \\
AEPBC & 2856 & $0.9006\pm0.0077$ & $0.9262\pm0.0033$ & $0.9114\pm0.0058$ & $0.2768\pm0.0025$ & $0.0061\pm0.0014$ & $0.0202\pm0.0018$ & $24.58\pm0.10$ & $24.77\pm0.08$ & $24.61\pm0.02$ \\
AWW & 1546 & $0.7822\pm0.0054$ & $0.8566\pm0.0049$ & $0.8185\pm0.0042$ & $0.2762\pm0.0055$ & $0.0043\pm0.0004$ & $0.0244\pm0.0038$ & $27.29\pm0.17$ & $27.54\pm0.44$ & $27.53\pm0.19$ \\
\bottomrule
\end{tabular}}
\caption{\rev{FM-BC, FM-BC + PAR with the October-selected setting (E0), and FM-BC + PAR (C3) at the October-selected FM-BC checkpoints, scored on November--December; mean $\pm$ standard deviation over 3 seeds. The three arms are scored on identical episodes.}}
\label{tab:sel_par}

\end{table}

\subsection{Retrained trajectory-prediction baselines}
\label{app:sel_tp}
The trajectory-prediction baselines are selected differently from our policies: GLRP and TrAISformer chose their checkpoints on 28 random January--September days and therefore trained on 245 days, while our policies trained on all 273. \rev{We retrained both on all 273 days with the same architectures, training settings, and adaptations as in Appendix~\ref{app:methods}. Each selects its checkpoint on October with its own published rule (GLRP: lowest validation ADE; TrAISformer: lowest validation loss) and is scored on November--December (Table~\ref{tab:sel_tp}; October and full-test results in Table~\ref{tab:sel_tp_of}). The stored and retrained models are scored on the same episodes, and each retrained run uses the same GPU type as the stored run it is compared with.}

\rev{On November--December, $\reachg$ rises in 6 of the 8 region $\times$ model cells. GLRP on AEW drops from 0.538 to 0.481 because of one retrained seed whose one-shot plans arrive late. Its November--December $\reachg$ is 0.291, against 0.563 and 0.588 for the other two seeds, and it has a strict reach rate of 0.20, a median delay of $+5$ steps, and an FDE of 218\,m. That seed's training ended normally (early stopping, with a validation ADE comparable to the other seeds). The collision rate changes by at most 1.9\,pp and $\gcade$ by at most 1.7\,m. No ordering among FM-BC, GLRP, and TrAISformer changes on any metric, split, or region.}

\begin{table}[!htbp]\color{blue}
\centering\scriptsize
\setlength{\tabcolsep}{4pt}
\begin{tabular}{lllccc}
\toprule
Region & Method & Model & $\reachg\uparrow$ & $C\downarrow$ & $\gcade\downarrow$ (m) \\
\midrule
AEW & GLRP & stored & $0.5377 \pm 0.0111$ & $0.2487 \pm 0.0037$ & $16.84 \pm 0.32$ \\
AEW & GLRP & retrained & $0.4807 \pm 0.1648$ & $0.2496 \pm 0.0137$ & $16.55 \pm 0.35$ \\
AEW & TrAISformer & stored & $0.2262 \pm 0.0098$ & $0.5102 \pm 0.0118$ & $38.57 \pm 0.88$ \\
AEW & TrAISformer & retrained & $0.2488 \pm 0.0081$ & $0.5072 \pm 0.0055$ & $36.92 \pm 0.35$ \\
AEW & FM-BC & October-selected & $0.8955 \pm 0.0064$ & $0.2045 \pm 0.0055$ & $23.08 \pm 0.10$ \\
\midrule
AESPA & GLRP & stored & $0.6008 \pm 0.0638$ & $0.5232 \pm 0.0113$ & $22.36 \pm 0.62$ \\
AESPA & GLRP & retrained & $0.6148 \pm 0.0631$ & $0.5046 \pm 0.0101$ & $22.27 \pm 0.14$ \\
AESPA & TrAISformer & stored & $0.1388 \pm 0.0079$ & $0.7668 \pm 0.0103$ & $58.90 \pm 1.12$ \\
AESPA & TrAISformer & retrained & $0.1521 \pm 0.0067$ & $0.7620 \pm 0.0039$ & $58.64 \pm 1.21$ \\
AESPA & FM-BC & October-selected & $0.9303 \pm 0.0008$ & $0.4492 \pm 0.0025$ & $26.77 \pm 0.04$ \\
\midrule
AEPBC & GLRP & stored & $0.4775 \pm 0.0242$ & $0.3372 \pm 0.0063$ & $20.71 \pm 0.27$ \\
AEPBC & GLRP & retrained & $0.5291 \pm 0.0223$ & $0.3454 \pm 0.0033$ & $20.73 \pm 0.42$ \\
AEPBC & TrAISformer & stored & $0.1716 \pm 0.0103$ & $0.5125 \pm 0.0084$ & $47.57 \pm 0.68$ \\
AEPBC & TrAISformer & retrained & $0.1748 \pm 0.0136$ & $0.5215 \pm 0.0126$ & $48.59 \pm 0.66$ \\
AEPBC & FM-BC & October-selected & $0.8986 \pm 0.0066$ & $0.2782 \pm 0.0032$ & $24.57 \pm 0.09$ \\
\midrule
AWW & GLRP & stored & $0.5580 \pm 0.0269$ & $0.3262 \pm 0.0143$ & $20.50 \pm 0.42$ \\
AWW & GLRP & retrained & $0.5466 \pm 0.0692$ & $0.3226 \pm 0.0049$ & $20.43 \pm 0.17$ \\
AWW & TrAISformer & stored & $0.1971 \pm 0.0167$ & $0.4476 \pm 0.0056$ & $53.30 \pm 0.85$ \\
AWW & TrAISformer & retrained & $0.2089 \pm 0.0093$ & $0.4513 \pm 0.0126$ & $52.24 \pm 0.73$ \\
AWW & FM-BC & October-selected & $0.7784 \pm 0.0125$ & $0.2777 \pm 0.0086$ & $27.23 \pm 0.12$ \\
\bottomrule
\end{tabular}
\caption{\rev{Trajectory-prediction baselines on November--December (61 days), mean $\pm$ standard deviation over 3 seeds. ``Stored'': the paper's models (245 training days, checkpoint selected on 28 random January--September days). ``Retrained'': all 273 training days, checkpoint selected on October. FM-BC uses its October-selected checkpoints.}}
\label{tab:sel_tp}
\end{table}

\begin{table}[!htbp]\color{blue}
\centering\scriptsize
\setlength{\tabcolsep}{3pt}
\resizebox{\linewidth}{!}{%
\begin{tabular}{lllcccccc}
\toprule
& & & \multicolumn{3}{c}{October (31 days)} & \multicolumn{3}{c}{Full test (92 days)} \\
\cmidrule(lr){4-6}\cmidrule(lr){7-9}
Region & Method & Model & $\reachg\uparrow$ & $C\downarrow$ & $\gcade\downarrow$ (m) & $\reachg\uparrow$ & $C\downarrow$ & $\gcade\downarrow$ (m) \\
\midrule
AEW & GLRP & stored & $0.5190 \pm 0.0324$ & $0.2341 \pm 0.0053$ & $16.85 \pm 0.21$ & $0.5315 \pm 0.0048$ & $0.2439 \pm 0.0024$ & $16.84 \pm 0.26$ \\
AEW & GLRP & retrained & $0.5044 \pm 0.1846$ & $0.2426 \pm 0.0209$ & $16.33 \pm 0.22$ & $0.4886 \pm 0.1714$ & $0.2473 \pm 0.0155$ & $16.48 \pm 0.30$ \\
AEW & TrAISformer & stored & $0.2259 \pm 0.0179$ & $0.5020 \pm 0.0035$ & $38.49 \pm 1.24$ & $0.2261 \pm 0.0125$ & $0.5075 \pm 0.0069$ & $38.54 \pm 1.00$ \\
AEW & TrAISformer & retrained & $0.2511 \pm 0.0229$ & $0.4937 \pm 0.0059$ & $36.92 \pm 0.34$ & $0.2495 \pm 0.0125$ & $0.5027 \pm 0.0052$ & $36.92 \pm 0.33$ \\
AEW & FM-BC & October-selected & $0.8898 \pm 0.0067$ & $0.2042 \pm 0.0030$ & $23.31 \pm 0.09$ & $0.8936 \pm 0.0065$ & $0.2044 \pm 0.0032$ & $23.16 \pm 0.09$ \\
\midrule
AESPA & GLRP & stored & $0.5862 \pm 0.0666$ & $0.4928 \pm 0.0040$ & $22.33 \pm 0.59$ & $0.5961 \pm 0.0647$ & $0.5134 \pm 0.0089$ & $22.35 \pm 0.61$ \\
AESPA & GLRP & retrained & $0.5844 \pm 0.0673$ & $0.4980 \pm 0.0187$ & $21.99 \pm 0.31$ & $0.6050 \pm 0.0645$ & $0.5025 \pm 0.0122$ & $22.18 \pm 0.19$ \\
AESPA & TrAISformer & stored & $0.1427 \pm 0.0053$ & $0.7595 \pm 0.0082$ & $58.95 \pm 1.40$ & $0.1401 \pm 0.0070$ & $0.7644 \pm 0.0091$ & $58.92 \pm 1.20$ \\
AESPA & TrAISformer & retrained & $0.1446 \pm 0.0082$ & $0.7617 \pm 0.0042$ & $58.65 \pm 1.29$ & $0.1497 \pm 0.0072$ & $0.7619 \pm 0.0031$ & $58.64 \pm 1.22$ \\
AESPA & FM-BC & October-selected & $0.9332 \pm 0.0004$ & $0.4504 \pm 0.0038$ & $26.78 \pm 0.09$ & $0.9312 \pm 0.0004$ & $0.4496 \pm 0.0007$ & $26.77 \pm 0.05$ \\
\midrule
AEPBC & GLRP & stored & $0.4576 \pm 0.0352$ & $0.3287 \pm 0.0088$ & $20.21 \pm 0.17$ & $0.4710 \pm 0.0278$ & $0.3344 \pm 0.0017$ & $20.55 \pm 0.21$ \\
AEPBC & GLRP & retrained & $0.5229 \pm 0.0227$ & $0.3465 \pm 0.0253$ & $19.87 \pm 0.46$ & $0.5271 \pm 0.0219$ & $0.3457 \pm 0.0080$ & $20.45 \pm 0.40$ \\
AEPBC & TrAISformer & stored & $0.1528 \pm 0.0134$ & $0.5153 \pm 0.0180$ & $48.22 \pm 0.74$ & $0.1655 \pm 0.0110$ & $0.5134 \pm 0.0002$ & $47.78 \pm 0.70$ \\
AEPBC & TrAISformer & retrained & $0.1680 \pm 0.0099$ & $0.5187 \pm 0.0082$ & $49.00 \pm 0.83$ & $0.1726 \pm 0.0122$ & $0.5206 \pm 0.0108$ & $48.73 \pm 0.71$ \\
AEPBC & FM-BC & October-selected & $0.8941 \pm 0.0053$ & $0.2805 \pm 0.0043$ & $23.32 \pm 0.34$ & $0.8971 \pm 0.0058$ & $0.2790 \pm 0.0023$ & $24.17 \pm 0.09$ \\
\midrule
AWW & GLRP & stored & $0.5800 \pm 0.0378$ & $0.2581 \pm 0.0034$ & $20.10 \pm 0.79$ & $0.5662 \pm 0.0306$ & $0.3010 \pm 0.0103$ & $20.35 \pm 0.54$ \\
AWW & GLRP & retrained & $0.5543 \pm 0.1187$ & $0.2445 \pm 0.0132$ & $19.46 \pm 0.11$ & $0.5494 \pm 0.0865$ & $0.2937 \pm 0.0044$ & $20.07 \pm 0.09$ \\
AWW & TrAISformer & stored & $0.2074 \pm 0.0232$ & $0.3759 \pm 0.0095$ & $50.04 \pm 0.13$ & $0.2009 \pm 0.0170$ & $0.4211 \pm 0.0049$ & $52.09 \pm 0.57$ \\
AWW & TrAISformer & retrained & $0.2239 \pm 0.0200$ & $0.3682 \pm 0.0168$ & $49.91 \pm 0.17$ & $0.2145 \pm 0.0111$ & $0.4205 \pm 0.0084$ & $51.38 \pm 0.40$ \\
AWW & FM-BC & October-selected & $0.7728 \pm 0.0111$ & $0.2111 \pm 0.0061$ & $27.28 \pm 0.31$ & $0.7763 \pm 0.0120$ & $0.2531 \pm 0.0053$ & $27.25 \pm 0.11$ \\
\bottomrule
\end{tabular}}
\caption{\rev{As Table~\ref{tab:sel_tp}, on October (the selection month for the retrained models and FM-BC) and on the full 92-day test set.}}
\label{tab:sel_tp_of}
\end{table}
}

\FloatBarrier
\section{More closely matched replay throughput}
\label{app:thr}
{\color{blue}
The transition-only measurement of Section~\ref{sec:throughput} (Appendix~\ref{app:compute}) and the public CPU implementation of \sns{} perform different work. We therefore measured a \rev{more closely matched} replay workload on the same GPUs.

\paragraph{\rev{More closely matched workload.}} It uses the same 29,858 AEW scenarios (22,258 training and 7,600 test) as the transition-only runner, in the same longest-first order, consumed once per repeat, with the same 128 vessel slots, 10\,s step, automatic reset, world counts (1--4,096), and three timed repeats. On top of that runner it:
(i)~consumes every element of the policy observation (ego history, all 10 neighbours, every feature channel, and the goal), with the history length ($H=10$) and neighbour count of the CPU reference;
(ii)~computes per-step tracking, kinematics, near-miss, and hull-collision metrics, all checked to equal our evaluator's values on 256 sampled expert rollouts; and
(iii)~applies the CPU reference's termination (200\,m goal, region exit, or log end), which gives 79.95 steps per episode against the CPU's 79.94.
Steps per second count valid controlled ego steps. We report the peak over world counts of the median of three repeats. Figure~\ref{fig:thr_matched} shows the throughput at every world count. 

\paragraph{FM-BC closed-loop rollout.} We also time the unmodified evaluation driver on FM-BC (AEW seed 0), with full observations, metrics, and policy inference at every step.

\begin{table}[!htbp]\color{blue}
\centering\small
\setlength{\tabcolsep}{4pt}
\begin{tabular}{lcc}
\toprule
Workload & RTX 3090 & H100 NVL \\
\midrule
Transition-only microbenchmark (Section~\ref{sec:throughput}) & 3,109,757 (986$\times$) & 9,212,347 (2,922$\times$) \\
More closely matched replay (full obs., metrics, termination) & 534,635 (170$\times$) & 1,277,186 (405$\times$) \\
FM-BC closed-loop rollout, policy inference every step & 22,404 (batch 128) & 56,791 (batch 512) \\
\bottomrule
\end{tabular}
\caption{\rev{Ego steps per second (peak median of 3 repeats). Ratios are against the stored CPU reference of \sns{}: 3,153 steps per second with 32 processes (207 with one process) on an AMD EPYC 7702P. The more closely matched replay peaks at 1,024 worlds on the RTX 3090 and 4,096 on the H100 NVL.}}
\label{tab:thr}
\end{table}

\begin{figure}[!htbp]\color{blue}
\centering
\includegraphics[width=0.55\linewidth]{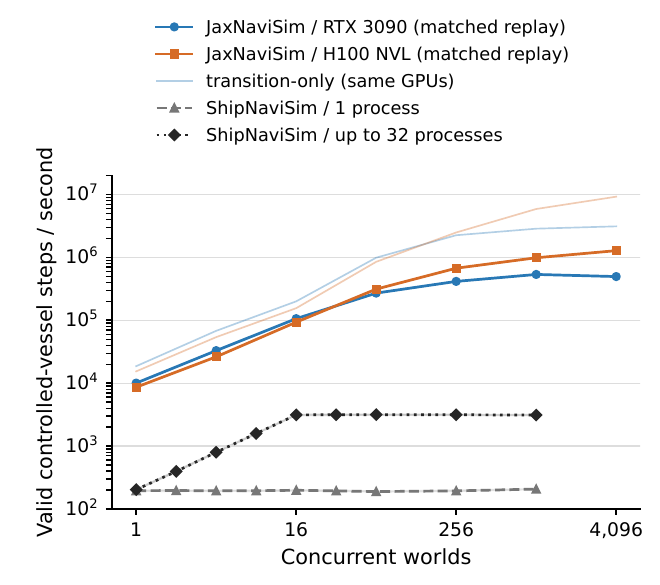}
\caption{\rev{Throughput of the more closely matched replay workload against the number of concurrent worlds (one controlled ego per world): median and minimum--maximum over three repeats. Faint lines: the transition-only microbenchmark of Section~\ref{sec:throughput}, measured again on the same GPUs. The \sns{} CPU lines are the stored reference of Figure~\ref{fig:replaythroughput}a. The peaks are the values of Table~\ref{tab:thr}.}}
\label{fig:thr_matched}
\end{figure}

}

\FloatBarrier
\section{Robustness of PAR}
\label{app:rob}
{\color{blue}
All tests use FM-BC seed 0 on the full AEW and AESPA test sets, without and with PAR (configuration C3, unchanged). Collisions are always scored on the true states. Each zero-perturbation run reproduces the reference exactly on the same GPU type; the AESPA PAR reference is an L40S run, since the stored H100 NVL cell could not be re-run on the same GPU type. ``Reduction kept'' is $(C_{\text{FM-BC}}(L)-C_{\text{PAR}}(L))/(C_{\text{FM-BC,clean}}-C_{\text{PAR,clean}})$ for perturbation level $L$.

\paragraph{Sea current.} The true motion integrates the HYCOM surface current at every step. PAR's own-ship forecast adds only the current measured at the decision step, held constant, and \rev{neighbours are forecast from their observed over-ground positions (finite differences, Appendix~\ref{app:controller})}. The current lowers FM-BC's $\reachg$ from 0.8929 to 0.6521 on AEW and from 0.9304 to 0.6075 on AESPA. With PAR, $\reachg$ is 0.7120 and 0.7566, respectively, which is 6 and 15 points above FM-BC alone. The collision rate with PAR is 0.0186 and 0.0441, against 0.3197 and 0.5587 for FM-BC (Table~\ref{tab:rob_current}). Under the current, PAR still removes 94\% (AEW) and 92\% (AESPA) of FM-BC's collisions, the same fractions as on clean data (94\% and 92\%). We do not report ``reduction kept'' here, because the current raises FM-BC's own collision rate.

\paragraph{Delayed observations.} Every observed neighbour state is 1--3 steps (10--30\,s) old. PAR keeps at least 95\% of its collision reduction, with $\reachg$ within 0.4\,pp of the clean run (Table~\ref{tab:rob_obs}).

\paragraph{Noisy positions: a negative result.} We add Gaussian noise, independent per vessel and per step, to the observed neighbour positions. Speed and heading stay as logged, because AIS speed and course over ground are not derived from differenced positions. The perturbed observation feeds both the policy and PAR's forecast. At $\sigma=10$\,m PAR keeps only about 6\% of its reduction. At $\sigma\geq25$\,m it collides more than FM-BC alone, and it also loses 4--14\,pp of $\reachg$. FM-BC alone is insensitive (within 0.3\,pp). \rev{PAR estimates each neighbour's velocity by differencing its two latest observed positions, 10\,s apart, and extrapolates it over the 160\,s horizon with zero clearance margin, so independent position noise becomes a large velocity error. The results are consistent with corrupted forecasts: on AEW the fallback rate rises} from 0.14 to 2.5 per episode at 10\,m and to 7.3 at 50\,m.

\begin{table}[!htbp]\color{blue}
\centering\scriptsize
\setlength{\tabcolsep}{4pt}
\begin{tabular}{llcccccc}
\toprule
& & \multicolumn{3}{c}{FM-BC} & \multicolumn{3}{c}{FM-BC + PAR} \\
\cmidrule(lr){3-5}\cmidrule(lr){6-8}
Region & Setting & $\reachg$ & $C$ & $\gcade$ & $\reachg$ & $C$ & $\gcade$ \\
\midrule
AEW & clean & 0.8929 & 0.1968 & 23.08 & 0.8925 & 0.0114 & 22.97 \\
AEW & HYCOM current & 0.6521 & 0.3197 & 32.57 & 0.7120 & 0.0186 & 32.02 \\
\midrule
AESPA & clean & 0.9304 & 0.4568 & 26.91 & 0.9236 & 0.0362 & 27.12 \\
AESPA & HYCOM current & 0.6075 & 0.5587 & 35.67 & 0.7566 & 0.0441 & 35.09 \\
\bottomrule
\end{tabular}
\caption{\rev{HYCOM surface current (delta dynamics with current, unit gains), FM-BC seed 0, full test set. Clean rows are the same-GPU zero-perturbation references; a zero-current run reproduces them exactly.}}
\label{tab:rob_current}
\end{table}

\begin{table}[!htbp]\color{blue}
\centering\scriptsize
\setlength{\tabcolsep}{4pt}
\begin{tabular}{llccccccc}
\toprule
& & \multicolumn{3}{c}{FM-BC} & \multicolumn{3}{c}{FM-BC + PAR} & \\
\cmidrule(lr){3-5}\cmidrule(lr){6-8}
Region & Neighbour observation & $\reachg$ & $C$ & $\gcade$ & $\reachg$ & $C$ & $\gcade$ & Reduction kept \\
\midrule
AEW & clean & 0.8929 & 0.1968 & 23.08 & 0.8925 & 0.0114 & 22.97 & 100.0 \% \\
AEW & noise $\sigma=$\,10\,m & 0.8917 & 0.1941 & 23.09 & 0.8426 & 0.1820 & 25.22 & 6.5 \% \\
AEW & noise $\sigma=$\,25\,m & 0.8929 & 0.1941 & 23.09 & 0.7913 & 0.2605 & 27.51 & $-$35.8 \% \\
AEW & noise $\sigma=$\,50\,m & 0.8922 & 0.1962 & 23.09 & 0.7579 & 0.2863 & 29.05 & $-$48.6 \% \\
AEW & delay 1 step (10\,s) & 0.8929 & 0.1964 & 23.09 & 0.8916 & 0.0142 & 22.97 & 98.3 \% \\
AEW & delay 2 steps (20\,s) & 0.8941 & 0.1963 & 23.09 & 0.8932 & 0.0180 & 23.02 & 96.2 \% \\
AEW & delay 3 steps (30\,s) & 0.8932 & 0.1967 & 23.09 & 0.8936 & 0.0201 & 23.01 & 95.2 \% \\
AESPA & clean & 0.9304 & 0.4568 & 26.91 & 0.9236 & 0.0362 & 27.12 & 100.0 \% \\
AESPA & noise $\sigma=$\,10\,m & 0.9293 & 0.4547 & 26.91 & 0.8797 & 0.4284 & 29.13 & 6.3 \% \\
AESPA & noise $\sigma=$\,25\,m & 0.9281 & 0.4550 & 26.91 & 0.8295 & 0.5020 & 30.84 & $-$11.2 \% \\
AESPA & noise $\sigma=$\,50\,m & 0.9302 & 0.4551 & 26.91 & 0.7906 & 0.5338 & 32.07 & $-$18.7 \% \\
AESPA & delay 1 step (10\,s) & 0.9308 & 0.4553 & 26.92 & 0.9270 & 0.0426 & 27.11 & 98.1 \% \\
AESPA & delay 2 steps (20\,s) & 0.9304 & 0.4570 & 26.91 & 0.9254 & 0.0518 & 27.11 & 96.3 \% \\
AESPA & delay 3 steps (30\,s) & 0.9297 & 0.4560 & 26.92 & 0.9250 & 0.0542 & 27.19 & 95.5 \% \\
\bottomrule
\end{tabular}
\caption{\rev{Perturbed neighbour observations, FM-BC seed 0, full test set. Noise is Gaussian on observed neighbour positions, independent per vessel and step; delay replaces every neighbour channel with its value 1--3 steps earlier. A negative ``reduction kept'' means PAR collides more than FM-BC alone.}}
\label{tab:rob_obs}
\end{table}

\paragraph{Goal-handling thresholds.} PAR's goal rule (Appendix~\ref{app:controller}, Goal handling) has two thresholds. The \emph{goal-rule radius} (400\,m) is the distance from the goal within which the rule applies. Within this radius, a feasible policy chunk is no longer kept automatically if its predicted progress toward the goal over the four committed actions is below the \emph{progress threshold} (10.4\,m). The progress threshold was not tuned because it is the benchmark's anchored-speed limit (0.26\,m/s) multiplied by the 40\,s of one action chunk. We varied each one at a time (radius 200/800\,m; threshold 5.2/20.8\,m) on AEW, seed 0, full test set (Table~\ref{tab:rob_goal}). The collision rate stays within 1.12--1.16\% (default 1.14\%), and $\reachg$ changes by at most $+0.33$ points.

\begin{table}[!htbp]\color{blue}
\centering\scriptsize
\setlength{\tabcolsep}{4pt}
\begin{tabular}{lcccccc}
\toprule
& \multicolumn{3}{c}{Full test set} & \multicolumn{3}{c}{November--December} \\
\cmidrule(lr){2-4}\cmidrule(lr){5-7}
PAR setting (AEW, seed 0) & $\reachg$ & $C$ & $\gcade$ & $\reachg$ & $C$ & $\gcade$ \\
\midrule
default: radius 400 m, threshold 10.4 m (stored) & 0.8925 & 0.0114 & 22.97 & 0.8907 & 0.0122 & 22.83 \\
goal-rule radius 200 m & 0.8928 & 0.0116 & 22.97 & 0.8911 & 0.0124 & 22.84 \\
goal-rule radius 800 m & 0.8928 & 0.0112 & 22.97 & 0.8909 & 0.0118 & 22.84 \\
progress threshold 5.2 m & 0.8926 & 0.0112 & 22.97 & 0.8915 & 0.0118 & 22.83 \\
progress threshold 20.8 m & 0.8958 & 0.0116 & 22.98 & 0.8949 & 0.0124 & 22.86 \\
\bottomrule
\end{tabular}
\caption{\rev{Sensitivity of PAR to its goal-handling thresholds, one at a time. The default row is the stored C3 run, evaluated on the same GPU type as the variants.}}
\label{tab:rob_goal}
\end{table}

}

\FloatBarrier
\section{Aggregate results over the anchorage regions}
\label{app:agg}
{\color{blue}
Following~\citet{agarwal2021deep}, we aggregate the 12 runs of each method (4 anchorage regions $\times$ 3 seeds) and report the interquartile mean (IQM, the mean of the middle 6 of the 12 scores) and the mean, with 95\% stratified-bootstrap confidence intervals (50,000 replicates, resampling seeds within each region).

\paragraph{Normalization.} Human replay reaches every goal by construction ($\reachg=1$) and has $C=0$, so $\reachg$ is already a fraction of the replay's score. We use the collision-free rate $1-C$ so that higher is better on both axes. \rev{$\gcade$ is not aggregated: the aggregate focuses on goal reaching and collision avoidance, and $\gcade$ stays in the per-region tables.} We keep $\reachg$ and $1-C$ separate and do not combine them into one score, because the three methods with the highest $\reachg$ (DEAS-BoN, QC-FQL, ReinFlow) \rev{barely change collisions (IQM $1-C$ within 0.01 of FM-BC's)}, whereas PAR raises IQM $1-C$ from 0.734 to 0.979 at similar $\reachg$.

\paragraph{Robustness of the orderings.} The orderings by IQM are unchanged when checkpoints are selected on October and scored on November--December (Table~\ref{tab:agg_b}), except that HR-PPO and FM-BC swap on $\reachg$, with overlapping intervals.

\paragraph{Probability of improvement.} Figure~\ref{fig:poi} shows the average probability that a run of each method improves on a run of FM-BC in the same region. QC-FQL improves on FM-BC with probability 1.00 on $\reachg$ and 0.86 on $1-C$; FM-BC + PAR with 0.60 on $\reachg$ and 1.00 on $1-C$; DEAS-BoN with 1.00 on $\reachg$ and 0.51 on $1-C$.

\begin{table}[!htbp]\color{blue}
\centering\scriptsize
\setlength{\tabcolsep}{4pt}
\begin{tabular}{lcccc}
\toprule
Method & $\reachg$ IQM [95\% CI] & $\reachg$ mean [95\% CI] & $1-C$ IQM [95\% CI] & $1-C$ mean [95\% CI] \\
\midrule
DEAS-BoN & 0.960 [0.959, 0.961] & 0.933 [0.931, 0.935] & 0.736 [0.734, 0.739] & 0.704 [0.703, 0.706] \\
QC-FQL & 0.931 [0.930, 0.932] & 0.921 [0.920, 0.922] & 0.743 [0.741, 0.744] & 0.709 [0.708, 0.710] \\
ReinFlow & 0.922 [0.916, 0.932] & 0.905 [0.898, 0.912] & 0.730 [0.723, 0.737] & 0.696 [0.689, 0.702] \\
FM-BC + PAR & 0.901 [0.899, 0.902] & 0.886 [0.883, 0.887] & 0.979 [0.977, 0.981] & 0.978 [0.977, 0.979] \\
FM-BC & 0.896 [0.892, 0.900] & 0.875 [0.872, 0.878] & 0.734 [0.731, 0.737] & 0.703 [0.702, 0.705] \\
HR-PPO & 0.894 [0.888, 0.897] & 0.892 [0.885, 0.902] & 0.712 [0.705, 0.720] & 0.687 [0.683, 0.691] \\
NF-BC & 0.758 [0.757, 0.760] & 0.753 [0.752, 0.754] & 0.713 [0.707, 0.719] & 0.679 [0.675, 0.681] \\
GLRP & 0.537 [0.530, 0.558] & 0.541 [0.524, 0.559] & 0.682 [0.678, 0.688] & 0.652 [0.649, 0.655] \\
PPO & 0.497 [0.259, 0.769] & 0.499 [0.380, 0.636] & 0.650 [0.575, 0.700] & 0.618 [0.566, 0.671] \\
TrAISformer & 0.183 [0.173, 0.192] & 0.183 [0.177, 0.189] & 0.490 [0.486, 0.492] & 0.448 [0.445, 0.451] \\
\midrule
Human replay (reference) & 1 & 1 & 1 & 1 \\
\bottomrule
\end{tabular}
\caption{\rev{Aggregates over 4 anchorage regions $\times$ 3 seeds, full 92-day test set, test-selected checkpoints. GLRP and TrAISformer are executed open-loop.}}
\label{tab:agg_a}
\end{table}

\begin{table}[!htbp]\color{blue}
\centering\scriptsize
\setlength{\tabcolsep}{4pt}
\begin{tabular}{lcccc}
\toprule
Method & $\reachg$ IQM [95\% CI] & $\reachg$ mean [95\% CI] & $1-C$ IQM [95\% CI] & $1-C$ mean [95\% CI] \\
\midrule
DEAS-BoN & 0.960 [0.958, 0.963] & 0.934 [0.932, 0.936] & 0.727 [0.725, 0.730] & 0.699 [0.698, 0.701] \\
QC-FQL & 0.929 [0.926, 0.932] & 0.918 [0.916, 0.920] & 0.733 [0.730, 0.737] & 0.704 [0.702, 0.707] \\
ReinFlow & 0.919 [0.912, 0.926] & 0.901 [0.894, 0.907] & 0.714 [0.711, 0.717] & 0.682 [0.676, 0.689] \\
FM-BC + PAR & 0.904 [0.901, 0.908] & 0.888 [0.885, 0.890] & 0.978 [0.976, 0.980] & 0.978 [0.977, 0.979] \\
HR-PPO & 0.898 [0.892, 0.905] & 0.899 [0.892, 0.906] & 0.702 [0.696, 0.707] & 0.681 [0.677, 0.686] \\
FM-BC & 0.897 [0.893, 0.901] & 0.876 [0.872, 0.879] & 0.722 [0.717, 0.726] & 0.698 [0.695, 0.700] \\
NF-BC & 0.761 [0.759, 0.762] & 0.758 [0.756, 0.760] & 0.704 [0.700, 0.706] & 0.672 [0.669, 0.675] \\
GLRP$^\dagger$ & 0.539 [0.533, 0.558] & 0.543 [0.526, 0.560] & 0.668 [0.662, 0.676] & 0.641 [0.637, 0.646] \\
PPO & 0.496 [0.257, 0.770] & 0.499 [0.379, 0.636] & 0.647 [0.575, 0.698] & 0.614 [0.562, 0.666] \\
TrAISformer$^\dagger$ & 0.184 [0.175, 0.193] & 0.183 [0.178, 0.189] & 0.489 [0.482, 0.495] & 0.441 [0.436, 0.445] \\
\bottomrule
\end{tabular}
\caption{\rev{As Table~\ref{tab:agg_a}, but under the October protocol of Appendix~\ref{app:sel}, scored on November--December. FM-BC + PAR uses C3. $^\dagger$GLRP and TrAISformer use their stored models, scored on November--December only.}}
\label{tab:agg_b}
\end{table}

\begin{figure}[!htbp]\color{blue}
\centering
\includegraphics[width=\linewidth]{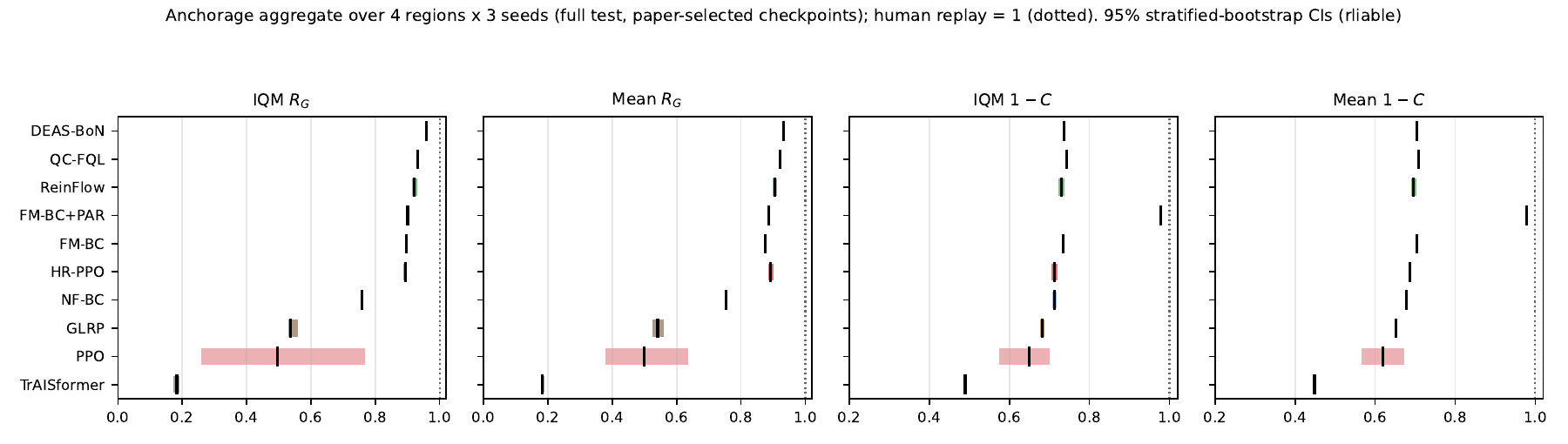}
\caption{\rev{IQM and mean of $\reachg$ and $1-C$ with 95\% stratified-bootstrap CIs, full test set, test-selected checkpoints (the numbers of Table~\ref{tab:agg_a}). Dotted line: human replay.}}
\label{fig:agg_full}
\end{figure}

\begin{figure}[!htbp]\color{blue}
\centering
\includegraphics[width=0.85\linewidth]{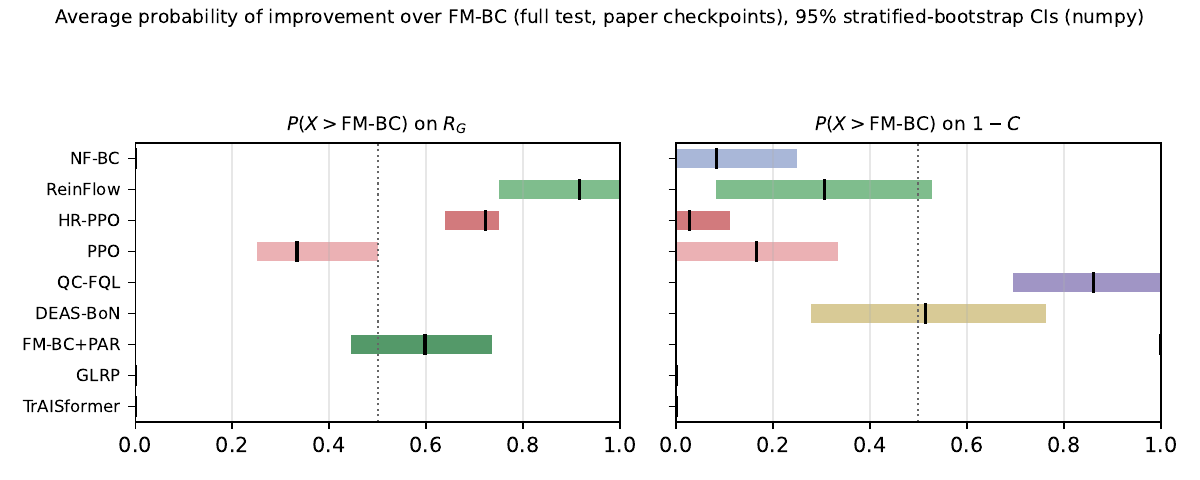}
\caption{\rev{Average probability of improvement over FM-BC, full test set, test-selected checkpoints, with 95\% stratified-bootstrap CIs (2,000 replicates). With 3 seeds per region the estimates are coarse.}}
\label{fig:poi}
\end{figure}
}

\FloatBarrier
\section{Expert-data statistics}
\label{app:exp}
{\color{blue}
Tables~\ref{tab:exp} and~\ref{tab:exp2} characterize the expert (replay) data of each anchorage region, separately for training (January--September) and test (October--December), over the benchmark episodes of Table~\ref{tab:funnel}. Duration is $(T-1)\cdot10$\,s; speed is the logged speed; the heading rate is $|\Delta\psi|/10$\,s. Closest approach is the minimum \emph{center-to-center} distance to any scene vessel. It differs from the hull-based CPA of Section~\ref{sec:benchmark} (the replay's median CPA is given in Table~\ref{tab:headline}). Density counts scene vessels within 1\,nm or 500\,m of the ego at every ego step.

\begin{itemize}[leftmargin=*,nosep]
\item \textbf{Density.} The scenes are dense and mostly stationary: a median of 21--39 vessels lie within 1\,nm of the ego at each step, and about 90\% of these vessel-steps are anchored.
\item \textbf{AESPA} is the most congested region. In about 30\% of its expert passages the ego comes within 100\,m (center-to-center) of another vessel, against 6\% in AEW. This is consistent with AESPA having the highest learned-policy collision rate (FM-BC 45\%).
\item \textbf{Gentle maneuvers.} The median heading rate is 0.035--0.05$^\circ$/s and the 95th percentile is 0.24--0.34$^\circ$/s. This is the reason PAR's replacement actions are rate-limited.

\end{itemize}

\begin{table}[!htbp]\color{blue}
\centering\scriptsize
\setlength{\tabcolsep}{3pt}
\resizebox{\linewidth}{!}{%
\begin{tabular}{lcccc}
\toprule
& \multicolumn{2}{c}{AEW} & \multicolumn{2}{c}{AESPA} \\
\cmidrule(lr){2-3}\cmidrule(lr){4-5}
& train & test & train & test \\
\midrule
Episodes & 22,258 & 7,600 & 20,396 & 7,068 \\
Distinct ego vessels (MMSI) & 3,511 & 1,573 & 4,701 & 1,978 \\
Arrival share (\%) & 23 & 22 & 75 & 76 \\
Ego hours & 5,986 & 2,000 & 8,505 & 2,932 \\
Duration (min), p5/p50/p95 & 7.5/12.7/34.3 & 7.7/12.5/33.2 & 9.5/24.2/43.2 & 9.8/24.2/42.0 \\
Step speed (kn), p5/p50/p95 & 0.70/5.65/9.0 & 0.70/5.70/9.0 & 0.60/4.70/8.55 & 0.60/4.85/8.60 \\
Episode-mean speed (kn), p5/p50/p95 & 2.1/6.3/9.6 & 2.1/6.3/9.6 & 1.9/4.5/8.5 & 1.9/4.5/8.6 \\
$|$Heading rate$|$ ($^\circ$/s), p50/p95/p99 & .035/.24/.47 & .035/.24/.44 & .05/.34/.66 & .05/.33/.65 \\
Episode max heading rate ($^\circ$/s), p50/p95 & 0.24/0.79 & 0.22/0.77 & 0.36/1.31 & 0.36/1.29 \\
Closest approach, center-to-center (m), p5/p25/p50/p95 & 97/142/181/356 & 95/141/180/350 & 61/94/135/367 & 60/92/132/364 \\
Episodes with closest approach $<$100\,m / $<$250\,m (\%) & 6/81 & 6/82 & 29/85 & 31/86 \\
Vessels within 1\,nm per step, p5/p25/p50/p75/p95 & 12/17/21/25/32 & 12/18/22/27/34 & 15/25/39/52/65 & 15/24/39/53/66 \\
Vessels within 500\,m per step, p5/p50/p95 & 0/1/4 & 0/1/4 & 0/2/9 & 0/3/9 \\
Anchored share of vessel-steps within 1\,nm ($<$0.26\,m/s, \%) & 91 & 91 & 92 & 93 \\
Scene vessels per episode, median & 88 & 93 & 126 & 127 \\
Ego tanker/cargo (\%) & 84/16 & 86/14 & 79/21 & 81/19 \\
Ego length (m), p5/p50/p95 & 41/86/225 & 41/86/223 & 41/89/229 & 41/89/229 \\
Ego length share (\%): 30--100/100--150/150--200/200--250/250--300/$\geq$300\,m & 71/12/9/6/1/0 & 72/12/10/5/2/0 & 62/11/12/13/1/1 & 64/12/12/11/1/0 \\
Neighbour types: tanker/cargo/other or unknown (\%) & 53/35/10 & 51/33/15 & 81/15/3 & 81/15/4 \\
Neighbour length (m), p5/p50/p95 & 41/169/330 & 42/150/321 & 37/85/275 & 37/85/251 \\
\bottomrule
\end{tabular}}
\caption{\rev{Expert-data statistics for AEW and AESPA (continued for AEPBC and AWW in Table~\ref{tab:exp2}). p$k$: $k$-th percentile. Step percentiles are read from histogram bins (0.05\,kn and 0.005$^\circ$/s) and give the upper edge of the bin.}}
\label{tab:exp}
\end{table}

\begin{table}[!htbp]\color{blue}
\centering\scriptsize
\setlength{\tabcolsep}{3pt}
\resizebox{\linewidth}{!}{%
\begin{tabular}{lcccc}
\toprule
& \multicolumn{2}{c}{AEPBC} & \multicolumn{2}{c}{AWW} \\
\cmidrule(lr){2-3}\cmidrule(lr){4-5}
& train & test & train & test \\
\midrule
Episodes & 11,985 & 4,219 & 7,404 & 2,454 \\
Distinct ego vessels (MMSI) & 3,016 & 1,377 & 1,535 & 798 \\
Arrival share (\%) & 56 & 54 & 52 & 53 \\
Ego hours & 4,053 & 1,415 & 2,304 & 786 \\
Duration (min), p5/p50/p95 & 8.5/19.0/36.8 & 8.8/18.8/36.2 & 6.3/16.2/38.7 & 6.5/16.8/39.9 \\
Step speed (kn), p5/p50/p95 & 0.55/4.35/8.40 & 0.55/4.40/8.50 & 0.55/3.80/8.30 & 0.50/3.75/8.10 \\
Episode-mean speed (kn), p5/p50/p95 & 1.6/4.2/8.5 & 1.6/4.2/8.7 & 1.7/4.1/8.8 & 1.5/4.1/8.4 \\
$|$Heading rate$|$ ($^\circ$/s), p50/p95/p99 & .045/.29/.57 & .045/.29/.56 & .05/.28/.48 & .05/.28/.48 \\
Episode max heading rate ($^\circ$/s), p50/p95 & 0.29/1.07 & 0.29/1.03 & 0.27/0.81 & 0.28/0.82 \\
Closest approach, center-to-center (m), p5/p25/p50/p95 & 74/115/157/393 & 74/114/157/394 & 82/133/186/412 & 82/128/177/372 \\
Episodes with closest approach $<$100\,m / $<$250\,m (\%) & 16/81 & 17/80 & 10/73 & 12/77 \\
Vessels within 1\,nm per step, p5/p25/p50/p75/p95 & 12/19/25/31/38 & 12/20/26/32/40 & 11/18/22/27/34 & 12/19/23/28/35 \\
Vessels within 500\,m per step, p5/p50/p95 & 0/2/5 & 0/2/5 & 0/2/4 & 0/2/4 \\
Anchored share of vessel-steps within 1\,nm ($<$0.26\,m/s, \%) & 91 & 91 & 89 & 89 \\
Scene vessels per episode, median & 55 & 58 & 67 & 70 \\
Ego tanker/cargo (\%) & 75/25 & 75/25 & 86/14 & 90/10 \\
Ego length (m), p5/p50/p95 & 41/94/229 & 41/94/229 & 40/99/185 & 40/99/183 \\
Ego length share (\%): 30--100/100--150/150--200/200--250/250--300/$\geq$300\,m & 60/13/11/16/1/0 & 60/13/12/15/0/0 & 50/26/21/2/1/0 & 51/28/19/1/0/0 \\
Neighbour types: tanker/cargo/other or unknown (\%) & 69/26/4 & 68/25/6 & 59/17/18 & 61/14/18 \\
Neighbour length (m), p5/p50/p95 & 40/98/289 & 40/98/289 & 31/114/265 & 31/112/260 \\
\bottomrule
\end{tabular}}
\caption{\rev{Expert-data statistics for AEPBC and AWW; conventions as in Table~\ref{tab:exp}.}}
\label{tab:exp2}
\end{table}
}

\FloatBarrier
\section{Dataset comparison}
\label{app:data}
{\color{blue}
Table~\ref{tab:datacmp} gives the data source, the covered areas and the data size of the simulators compared in Table~\ref{tab:comparison}, as stated by each work. For the driving simulators, the size is that of the source dataset. Compared with the closest maritime benchmark, \sns{}, ours covers eight regions instead of one. The main experiments of \sns{} use 8,437 training trajectories, and our four anchorage regions alone have 62,043 training episodes.

\begin{table}[!htbp]\color{blue}
\centering\footnotesize
\setlength{\tabcolsep}{4pt}
\begin{tabular}{l>{\raggedright\arraybackslash}p{3.7cm}>{\raggedright\arraybackslash}p{3.1cm}>{\raggedright\arraybackslash}p{3.0cm}}
\toprule
 & Dataset source & Areas & Size \\
\midrule
nuPlan~\citeyearpar{caesar2021nuplan} & nuPlan & 4 cities & 1,282 driving-hours \\
Nocturne~\citeyearpar{vinitsky2022nocturne} & Waymo Open Motion Dataset & 6 cities & 574 driving-hours \\
Waymax~\citeyearpar{gulino2023waymax} & Waymo Open Motion Dataset & 6 cities & 574 driving-hours \\
GPUDrive~\citeyearpar{kazemkhani2025gpudrive} & Waymo Open Motion Dataset & 6 cities & 574 driving-hours \\
MariNav~\citeyearpar{vaidheeswaran2025marinav} & AIS data from Kpler & Gulf of St.\ Lawrence & $>$10M records \\
\sns{}~\citeyearpar{pham2025shipnavisim} & not stated & a selected area of the Singapore Strait & 59.6M records \\
Zhang et al.~\citeyearpar{zhang2026realistic} & proprietary AIS data (industry collaboration) & a selected area of the Indian Ocean & not stated \\
\midrule
\jns{} (ours) & open AIS data from the Maritime and Port Authority of Singapore & 8 Singapore Strait regions & $\sim$141M records / $\sim$6.7M vessel-hours \\
\bottomrule
\end{tabular}
\caption{\rev{Data behind the simulators of Table~\ref{tab:comparison}, as stated by each work. For the driving simulators, the size is that of the source dataset: nuPlan's from \citet{karnchanachari2024nuplan} and the Waymo Open Motion Dataset's from \citet{ettinger2021large}. ``Not stated'': the work gives no source or no number.}}
\label{tab:datacmp}
\end{table}
}

\FloatBarrier
\section{An example of a PAR avoidance decision}
\label{app:parex}
\begin{figure}[!htbp]\color{blue}
\centering
\includegraphics[width=\linewidth]{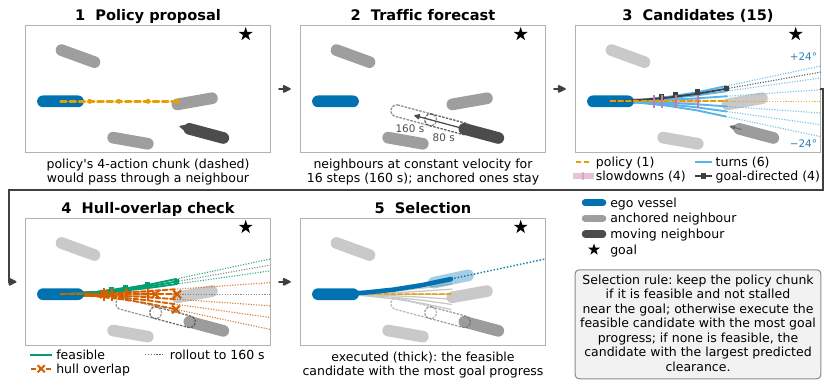}
\caption{\rev{Predictive Action Refinement (PAR) at one decision step (schematic with illustrative geometry). (1) The policy proposes a four-action chunk. (2) PAR forecasts the observed neighbours at constant velocity over 16 steps (160\,s). (3) It builds 15 candidates: the policy chunk, four slowdowns along the same path, six turns ($\pm8^\circ$, $\pm16^\circ$, $\pm24^\circ$, reached gradually), and four goal-directed chunks. (4) It rolls each candidate out over the horizon and marks it infeasible if the predicted capsule hull of the controlled vessel overlaps a forecast neighbour hull. (5) Here the policy chunk is infeasible, so PAR executes the feasible candidate with the most goal progress. The speed ramp of the slowdowns is exaggerated for visibility; Appendix~\ref{app:controller} gives the exact construction.}}
\label{fig:parschem}
\end{figure}

\begin{figure}[!htbp]\color{blue}
\centering
\includegraphics[width=0.92\linewidth]{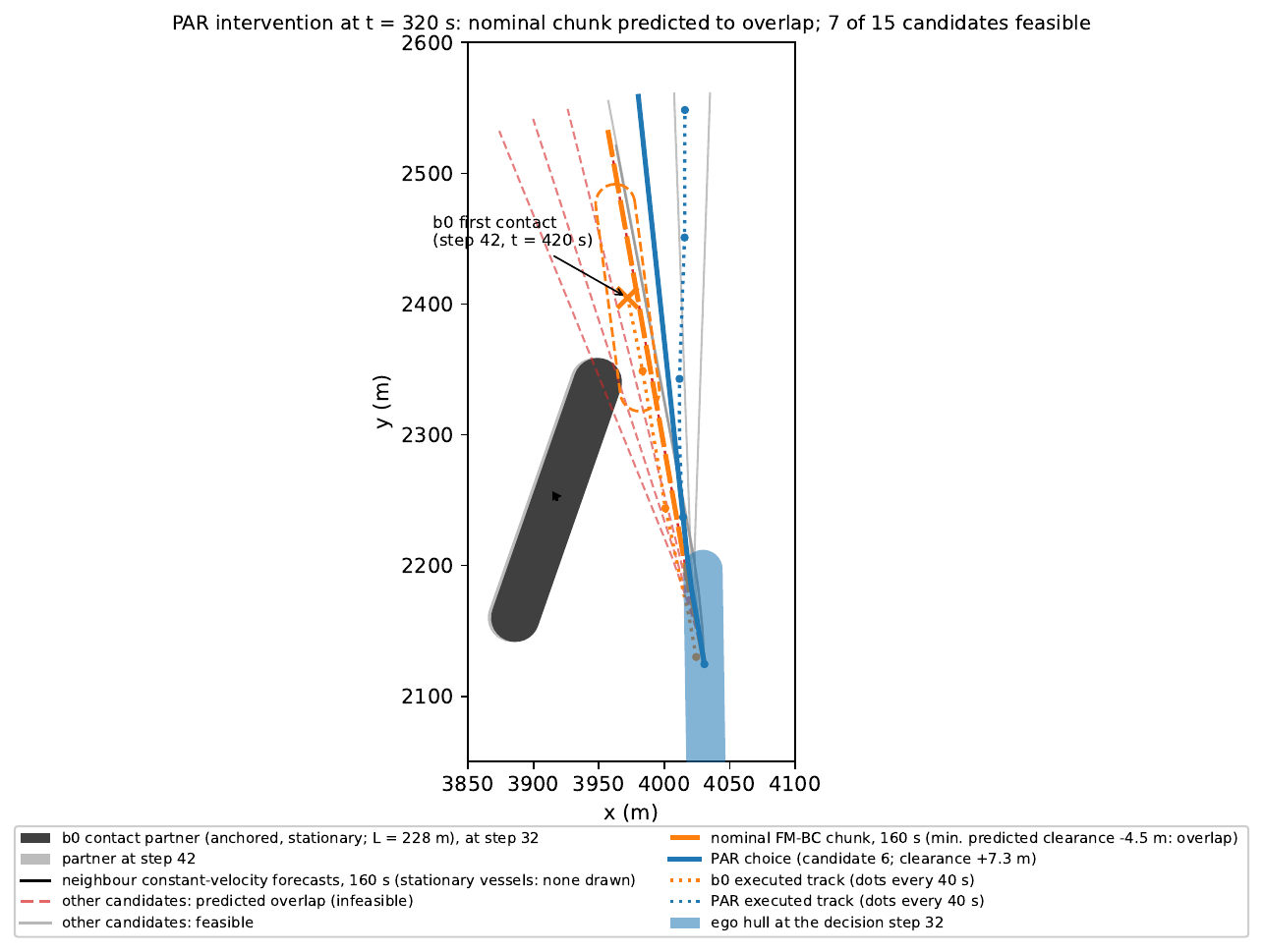}
\caption{\rev{PAR avoidance decision in an AEW test episode of 1 October 2024 (FM-BC seed 0, checkpoint 1.92M; PAR configuration C3). In the legend, ``b0'' is FM-BC without PAR. At $t=320$\,s the policy's proposed 160\,s chunk (orange dashed) has a minimum predicted hull clearance of $-4.5$\,m, an overlap. Seven of the 15 candidates are feasible, and PAR executes the feasible candidate with the most goal progress, a turn with $+7.3$\,m predicted clearance (blue). Without PAR, the same policy contacts an anchored 228\,m vessel at $t=420$\,s (cross). That vessel moves 2.5\,m (0.04\,m/s) between the two times. Neighbour forecasts are constant-velocity; stationary vessels have no drawn forecast.}}
\label{fig:parex}
\end{figure}

{\color{blue}
Figure~\ref{fig:parschem} illustrates the steps of PAR at one decision, and Figure~\ref{fig:parex} shows one AEW test episode with PAR's decision drawn in full. The episode is selected by a deterministic rule: over the AEW seed-0 test episodes in (day, trajectory identifier) order, we take the first in which (i) the stored results show that FM-BC collides and FM-BC + PAR does not, and (ii) a re-run of the episode's own evaluation batch reproduces both outcomes and PAR makes an avoidance intervention (the policy's chunk is predicted infeasible and a different candidate is executed) before FM-BC's first contact. We draw the last such intervention before that contact.
}

\FloatBarrier
\section{Cost-based test-time layers: guided sampling and MPPI}
\label{app:cost}
{\color{blue}
PAR is hand-designed (Section~\ref{sec:geo}). This appendix compares it with two test-time layers that instead minimize an explicit cost: cost-guided sampling of the flow policy, in the manner of guided diffusion planners \citep{zheng2025diffusion}, and model predictive path integral control (MPPI; \citealp{williams2017information}) with the policy as its sampling distribution. Both act on the same frozen FM-BC policies as PAR and use PAR's neighbour forecast and hull check. As in PAR, the policy's own chunk is never modified, and every replacement candidate respects the C3 motion limits (Appendix~\ref{app:envelope}). The appendix therefore compares candidate-generation and selection procedures under the same forecast, check and limits on replacement actions. Both layers are our adaptations of the cited approaches to a frozen flow policy, and all results use the paper's FM-BC policies (four-action chunks).

\paragraph{Cost.} For a candidate action chunk $a$ we use $E(a) = \lambda_c E_{\mathrm{col}}(a) + \lambda_p E_{\mathrm{prog}}(a)$ with $\lambda_c = 10$ and $\lambda_p = 1$. The collision term is $E_{\mathrm{col}}(a) = \sum_{h=1}^{16} \sum_{j} \max(0,\, m_j - d_{hj}(a))^2 / (50\,\mathrm{m})^2$, where $d_{hj}$ is the gap between the controlled vessel's capsule hull at forecast step $h$ and the capsule hull of neighbour $j$ at its constant-velocity forecast. The forecast and the 160\,s rollout of the chunk are PAR's. The gap uses a smoothed segment distance so that it is differentiable. The target gap is $m_j = \min(80\,\mathrm{m},\, d_{0j} - 1\,\mathrm{m})$, where $d_{0j}$ is the current gap, so a neighbour that is already closer than 80\,m is not penalized for its current distance. The progress term is $E_{\mathrm{prog}}(a) = \max(0,\, P(a^{\pi}) - P(a))^2 / (50\,\mathrm{m})^2$, where $P$ is the reduction of the distance to the goal over the four executed actions and $a^{\pi}$ is the policy's own unguided chunk. The cost therefore penalizes only progress lost relative to the policy. The weights and length scales were set once and were not tuned, apart from one $\lambda_p = 0$ setting in the tuning grid below.

\paragraph{Cost-guided sampling.} FM-BC generates a chunk by integrating its velocity field $v$ over $K = 3$ Euler steps from Gaussian noise. At each step we form the one-step estimate of the final sample, $\hat{x}_1 = x + (1 - t)\,v$, decode it to an action chunk, project the chunk onto the motion limits, and take the gradient $g$ of $E$ with respect to $\hat{x}_1$ (no gradient passes through the network). The step is $x \leftarrow x + (v + \gamma\,\hat{g})/K$ with $\hat{g} = -g / \lVert g \rVert$; $\gamma = 0$ recovers the policy's sampler. At each decision the layer forms $S = 16$ candidates: the policy's own unguided sample, left unchanged, and 15 guided samples from fresh noise, each projected onto PAR's C3 motion limits. The policy's own sample is executed if it passes PAR's hull check (capsule hulls, 160\,s forecast, zero margin) and PAR's goal rule. Otherwise, the passing candidate with the greatest progress toward the goal is executed. If no candidate passes, the candidate with the largest minimum forecast clearance is executed and counted as a fallback.

\paragraph{MPPI.} The layer draws $N = 64$ candidates: the policy's own sample, left unchanged, and 63 chunks decoded from fresh noise $z_i \sim \mathcal{N}(0, I)$ by the policy's sampler, each projected onto the motion limits. Candidate $i$ receives the weight $w_i \propto \exp(-E_i / \lambda)$ with $\lambda = \tau \cdot \mathrm{sd}_i(E_i)$ and $\tau = 0.1$. In a second iteration the noise mean is moved to $\mu = \sum_i w_i z_i$, new candidates are drawn from $\mathcal{N}(\mu, I)$, and their weights include the likelihood ratio between $\mathcal{N}(0, I)$ and $\mathcal{N}(\mu, I)$, so the target remains the policy's distribution reweighted by $\exp(-E/\lambda)$. The output is a weighted mean of chunks. To avoid averaging a pass on one side of a neighbour with a pass on the other, the candidates are split by the side of the policy's own sample on which their position after the four executed actions lies, and the mean is taken over the side with the larger total weight. The mean is then projected onto the motion limits. The output is executed if it passes the hull check. Otherwise, the passing candidate with the lowest cost is executed. If no candidate passes, the candidate with the largest minimum forecast clearance is executed (a fallback).

\paragraph{Protocol.} We follow the protocol recommended in Appendix~\ref{app:sel}. The policies are the paper's FM-BC runs (three seeds per region) at their October-selected checkpoints (Appendix~\ref{app:sel}). We tuned the layers' settings only on the October episodes of AEW, seed 0 (2,520 episodes), with a rule fixed before tuning: among the settings whose October $\reachg$ is at least FM-BC's minus 0.5\,pp, take the one with the lowest collision rate (ties broken by $\gcade$). We then froze the selected settings and evaluated them once on November--December, on all four anchorage regions and three seeds. No setting was tuned on another region, seed or month. PAR's settings were fixed on test runs (Appendix~\ref{app:envelope}). For a like-for-like comparison, we therefore re-select PAR's goal-rule radius (the distance from the goal within which its goal rule applies, Appendix~\ref{app:rob}) on the same October episodes by the same rule (described below) and report PAR with this setting. The 160\,s horizon is shared by all layers, and the progress threshold is not a tuned value (Appendix~\ref{app:rob}). The C3 motion limits, also fixed on test runs, are kept throughout, because every layer in this appendix uses them. AEPBC and AWW informed neither PAR's settings nor the cost-based layers' settings.

The tuning grids were as follows. \emph{Cost-guided sampling:} $S \in \{4, 16\}$ and $\gamma \in \{0.05, 0.1, 0.2, 0.5\}$; at $S = 16$ the grid was extended to $\gamma \in \{1, 2, 4\}$ because the selected value lay at its edge, and $\gamma = 4$ fails the $\reachg$ condition. Three variants at $S = 16$ were also run: $K = 10$ integration steps (at two values of $\gamma$), $\lambda_p = 0$, and an additional bow-crossing cost term; the last was excluded from selection in advance. \emph{MPPI:} $N \in \{16, 64\}$, one or two iterations, $\tau \in \{0.1, 1\}$, and as output either the highest-weight candidate or the side-wise weighted mean (six settings). Table~\ref{tab:cost_oct} lists the October results. The selected settings are: cost-guided sampling $S = 16$, $\gamma = 2$; MPPI $N = 64$, two iterations, $\tau = 0.1$, weighted mean. \emph{PAR:} goal-rule radius $\in \{200, 400, 800\}$\,m, with the 160\,s horizon and the 10.4\,m progress threshold kept; the default is 400\,m. All three settings satisfy the $\reachg$ condition, and their October collision rates are 0.87--0.91\% (22--23 collisions in 2,520 episodes). The rule selects the 200\,m radius: 22 collisions against 23 for the default, a paired difference of $-0.04$\,pp with 95\% interval $[-0.16, +0.08]$. 

All comparisons are paired on identical episodes within one evaluation job and one GPU type. Confidence intervals are 95\% day-clustered bootstrap intervals (2,000 resamples). FM-BC was re-run inside every evaluation job. Re-runs on different GPU types differ from each other and from Table~\ref{tab:sel_par} by at most 0.2\,pp in either metric, because floating-point differences between GPU types change the outcome of a few episodes. Re-runs on the same GPU type are identical episode by episode. The FM-BC row of Table~\ref{tab:cost_levels} is from the guided-sampling jobs. PAR with the October-selected goal-rule radius was evaluated in separate jobs. Eight of its twelve runs used the GPU type of the FM-BC row. AEW seed 1 and the three AESPA seeds used a different GPU type, on which a re-run of PAR with the default goal-rule radius differs from the same-type run by at most 0.35\,pp in the collision rate of one seed and 0.14\,pp in a regional mean.

\paragraph{Results.} Table~\ref{tab:cost_levels} gives the November--December results.
\begin{itemize}
\item \emph{Both cost-based layers remove a large share of FM-BC's collisions.} Cost-guided sampling removes 68--89\% and MPPI 50--72\%, against 91--94\% for PAR with the October-selected goal-rule radius.
\item \emph{Neither reaches PAR.} The collision rate of cost-guided sampling is 1.6--3.5 times PAR's, and that of MPPI is 4.0--5.5 times PAR's. Both are above PAR's in every region and seed.
\item \emph{Goal reaching.} Paired with FM-BC in the same job, cost-guided sampling changes the regional mean $\reachg$ by $-0.75$ to $+0.91$\,pp and MPPI by $-3.94$ to $+2.26$\,pp. MPPI executes a weighted mean of samples in almost every decision (about 97\% of decisions on AEW in October), so it changes the policy's behavior rather than intervening only when needed.
\item \emph{Cost.} Per episode, the guided layers take 2.0--2.7 times PAR's wall time and the MPPI layers 3.6--5.1 times (same GPU within each comparison).
\end{itemize}

With this cost, tuning procedure and policy, PAR, with its goal-rule radius selected on the same validation data, had lower collision rates than the tested guided-sampling and MPPI variants, under the same forecast, hull check and limits on replacement actions.

\begin{table}[!htbp]\color{blue}
\centering\scriptsize
\setlength{\tabcolsep}{4pt}
\begin{tabular}{lccc}
\toprule
Setting & $\reachg$ (\%) $\uparrow$ & $C$ (\%) $\downarrow$ & $\gcade$ (m) $\downarrow$ \\
\midrule
FM-BC & 89.33 & 19.33 & 23.20 \\
\midrule
\multicolumn{4}{l}{\emph{Cost-guided sampling, $S = 16$}} \\
$\gamma = 0.05$ & 89.64 & 10.28 & 23.17 \\
$\gamma = 0.1$ & 89.56 & 8.02 & 23.14 \\
$\gamma = 0.2$ & 89.68 & 5.75 & 23.11 \\
$\gamma = 0.5$ & 89.56 & 3.25 & 23.08 \\
$\gamma = 1$ & 89.40 & 2.54 & 23.00 \\
{\boldmath$\gamma = 2$} & \textbf{89.05} & \textbf{2.10} & \textbf{23.10} \\
$\gamma = 4$ $^\dagger$ & 88.65 & 1.55 & 23.07 \\
\midrule
\multicolumn{4}{l}{\emph{MPPI}} \\
$N = 16$, 1 iteration, highest weight & 88.85 & 13.69 & 22.87 \\
$N = 64$, 1 iteration, highest weight & 88.85 & 12.06 & 22.80 \\
$N = 64$, 1 iteration, weighted mean, $\tau = 0.1$ & 93.13 & 11.55 & 22.24 \\
$N = 64$, 1 iteration, weighted mean, $\tau = 1$ & 91.87 & 12.98 & 21.79 \\
$N = 64$, 2 iterations, highest weight & 92.38 & 6.35 & 26.22 \\
{\boldmath$N = 64$}\textbf{, 2 iterations, weighted mean} & \textbf{92.30} & \textbf{5.75} & \textbf{22.12} \\
\midrule
\multicolumn{4}{l}{\emph{PAR, goal-rule radius (horizon 160\,s, progress threshold 10.4\,m)}} \\
\textbf{200\,m} & \textbf{89.09} & \textbf{0.87} & \textbf{23.22} \\
400\,m (default) & 89.13 & 0.91 & 23.22 \\
800\,m & 89.21 & 0.91 & 23.23 \\
\bottomrule
\end{tabular}
\caption{\rev{October tuning results (AEW, seed 0; 2,520 episodes). The settings were tuned on these episodes only. \textbf{Bold}: the selected setting of each layer. $\dagger$: fails the condition $\reachg \geq$ FM-BC's $-$ 0.5\,pp (88.83\%). The $S = 4$ settings and the three variants (collision rate 4.4--16.7\%) are omitted.}}
\label{tab:cost_oct}
\end{table}

\begin{table}[!htbp]\color{blue}
\centering\scriptsize
\setlength{\tabcolsep}{4pt}
\resizebox{\linewidth}{!}{%
\begin{tabular}{lcccc}
\toprule
Layer & AEW & AESPA & AEPBC & AWW \\
\midrule
FM-BC & 89.74 / 20.39 / 22.99 & 92.97 / 45.93 / 26.86 & 90.06 / 27.68 / 24.58 & 78.22 / 27.62 / 27.29 \\
+ cost-guided sampling & 89.49 / 2.30 / 22.87 & 92.22 / 5.27 / 26.76 & 89.79 / 3.44 / 24.45 & 79.13 / 8.86 / 27.31 \\
+ MPPI & 92.02 / 5.79 / 22.04 & 88.92 / 13.04 / 27.26 & 89.33 / 8.02 / 23.91 & 80.25 / 13.71 / 24.90 \\
\midrule
+ PAR  & 89.64 / 1.15 / 22.93 & 92.53 / 3.23 / 27.07 & 90.86 / 2.01 / 24.60 & 81.29 / 2.50 / 27.57 \\
\bottomrule
\end{tabular}}
\caption{\rev{November--December results at the October-selected FM-BC checkpoints, mean over three seeds: $\reachg$ (\%) / $C$ (\%) / $\gcade$ (m). Episodes per seed: AEW 5,080, AESPA 4,791, AEPBC 2,856, AWW 1,546. The cost-based layers use settings selected on AEW seed 0 in October. ``+ PAR'' uses the goal-rule radius selected on the same October episodes (200\,m), with the 160\,s horizon, the 10.4\,m progress threshold and the C3 motion limits.}}
\label{tab:cost_levels}
\end{table}

}